\documentclass[journal]{IEEEtran}

\usepackage[numbers,sort&compress]{natbib}% our
\usepackage{footnote}
\usepackage{url}
\usepackage[colorlinks,linkcolor=red]{hyperref}
\usepackage{soul}% Added by us
\soulregister\cite7
\soulregister\citep7
\soulregister\citet7
\soulregister\ref7
\soulregister\pageref7

\usepackage{color}

\usepackage[pdftex]{graphicx}
\DeclareGraphicsExtensions{.pdf,.jpeg,.png,.jpg}

\usepackage{amsmath}

\DeclareMathOperator*{\argmin}{arg\,min}

\usepackage{amsfonts}
\usepackage{algpseudocode}%Added by us
\usepackage{algorithm}%Added by us
  \usepackage[caption=false,font=footnotesize]{subfig}
\usepackage{fixltx2e}

\usepackage{stfloats}

\begin{document}
\title{Multi-modal Core Tensor Factorization based Low-Rankness and Its Applications to Tensor Completion}

\author{Haijin~Zeng,
	    Jize Xue,
	    Hiêp Q. Luong
	    and Wilfried Philips % <-this % stops a space
\thanks{This work was supported by **.}
%\thanks{Haijin~Zeng  and Sheng Liu are with the College of Science, Northwest A\&F University, Yangling 712100, China e-mail: zeng\_navy@163.com, hzeng@nwafu.edu.cn.}
\thanks{Haijin Zeng, Jize Xue, Hiep Luong and W. Philips are with the Image Processing and Interpretation imec research group at Ghent University, 9000 Ghent, Belgium (e-mail: zeng\_navy@163.com, Jize.Xue@UGent.be, Hiep.Luong@UGent.be, Wilfried.Philips@UGent.be)}% <-this % stops a space
\thanks{Manuscript received April 19, 2005; revised August 26, 2015.}
}

\markboth{Preprint}%
{Shell \MakeLowercase{\textit{et al.}}: Bare Demo of IEEEtran.cls for IEEE Journals}



\maketitle

\begin{abstract}

Low-rank tensor completion has been widely used in computer vision and machine learning.
This paper develops a novel multi-modal core tensor factorization (MCTF) method combined with a tensor low-rankness measure and a better nonconvex relaxation form of this measure (NC-MCTF).
The proposed models encode low-rank insights for general tensors provided by Tucker and T-SVD, and thus are
expected to simultaneously model spectral low-rankness in multiple orientations and accurately restore the data of intrinsic low-rank structure based on few observed entries.
Furthermore, we study the MCTF and NC-MCTF regularization minimization problem, and design an effective block successive upper-bound minimization (BSUM) algorithm to solve them.
%Theoretically, we establish convergence of the proposed algorithm, without using diminishing stepsizes.
This efficient solver can extend MCTF to various tasks, such as tensor completion.
A series of experiments, including hyperspectral image (HSI), video and MRI completion, confirm the superior performance of the proposed method.
%Code is available at: \url{https://github.com/NavyZeng/MCTF}.

\end{abstract}


% \fi
\IEEEpeerreviewmaketitle

\section{Introduction}
%\IEEEPARstart{T}{his}

%begin test

Low-rankness is a common attribute of many data sources.
To date,
methods based on low-rankness have reported empirical and theoretical success on a large variety of scientific and engineering applications:
face modeling \cite{lowrank_face_model},
gene categorization \cite{lowrank_gene},
camera image processing \cite{lowrank_camera},
compressive imaging \cite{lowrank_imaging},
user interest prediction \cite{lowrank_user_recom}, etc.

%***Low rank is much more general than the SVD. The text is therefore a
%bit narrow-sighted***
A promising method that measures the low-rankness of a matrix is to account for the number of non-zero singular values \cite{KBR}.
%The low-rankness of the vector/matrix can be reasonably measured by
%the number of non-zero terms/non-zero
%singular values ($L_0$ norm/rank).
This low-rankness metric and its relaxations (for example, the $L_1$
norm and nuclear norm)
have been proven useful as regularisation terms in applications,
and have inspired various low-rank models and algorithms to cope with different tasks.
%However, data from many practical applications is usually generated by the interaction of multi-factors.
%The traditional 1-D or 2-D methods based on vector or matrix can only solve the single factor/binary factor variability of the data,
%which is obviously not the best way to maintain the multi-factor structure of the underlying data \cite{KBR}.
On the other hand, a large amount of data generated by modern sensors
is naturally represented by high-order tensors, whereas the SVD is
restricted to 2D data.

Early high-dimensional data analysis methods reformatted
high-dimensional data tensors
artificially into 2D matrices and resorted to methods developed for classic
two-dimensional analysis methods.
However, this flattening strategy and the strict assumptions inherent in two-dimensional analysis do not always match the high-dimensional data well.
For example, hyperspectral image (HSI) is the
imaging result of different spectral bands from the same spatial scene,
which indicates that there is a high correlation in the spectral
dimensions \cite {tensor_HSI}; a video contains multiple frames, which contain a high correlation in the temporal dimension, especially for adjacent frames \cite {wu2017robust}.
Therefore, converting these high-dimensional tensors data artificially to 2-D matrices spreads this ``local correlation'' (e.g., between adjacent frequencies) over large strides in the 2D matrix, complicating analysis.
%***It will NOT lose useful structure information because the
%conversion is lossless. In fact, such correlation can still be
%exploited by the SVD***
%Recently, many studies \cite{liu_tensor_2012_lrtdtv_18,yuan_tensor_2016_lrtdtv_19,cao_total_2016_lrtdtv_20,anandkumar_tensor_2016_lrtdtv_21,lu_tensor_2016_lrtdtv_22_trpca}
%have proven that completion methods directly modeling tensors can better preserve the multi-order structure information than the ones modeling the matriczation of tensor and help to enhance performance of various practical tasks, e.g., multispectral image (MSI)/hyperspectral image (HSI) denoising \cite{ZENG_HSI_tensor, zengTGRS, zengSP} and video inpainting/denoising et al. \cite{Matrix_inpainting_1, Matrix_denoising_1, high_order_web, 3Dimage_reconstruction}.
Thanks to the inherent high-dimensional structure of the data,
high-order tensor decomposition allows capturing correlation in a more local fashion along each dimension
%***I am personally not convinced that these tensor approaches are
%really appropriate for dimensionality reduction: all of these methods
%favor the coordinate axes and are poorly suited for any type of
%spatial data analysis. For video they are even less suitable (time
%delay in analysis, too high complexity). And for hyperspectral,
%probably better than tensor processing is to collapse the spatial dimensions
%into one and then apply SVD. However some sort of clustering approach
%would be even better.
%
%I would advise you to focus on other
%representations in your long term PhD research (different methods).
%See below why tensors are not a very good choice for (hyperspectral) image processing.
%We can discuss sometime.
%***
%
%***What does this mean: ``multiple interactions and couplings''. Avoid
%vaguie and meaningless statements***
\cite{cichocki2015tensor}.
In other words, only when analyzing the existing
inherent multi-dimensional patterns, 
%***which is exatly my point:
%tensors cannot do this well. E.g. consider an image with an edge under
%45 degrees.
%\begin{verbatim}
%Try the following in python (I advise not using matlab; python is much
%better):
%x=np.hstack([np.ones(50).reshape([-1,5]),
%np.zeros(50).reshape([-1,5])]);
%[u,s,vt]=np.linalg.svd(x); s
%Out[25]:
%array([7.07106781e+00, 1.19161639e-15,
%8.05127790e-32, 4.52574127e-49,
%       2.61495950e-65, 0.00000000e+00,
%0.00000000e+00, 0.00000000e+00,
%       0.00000000e+00, 0.00000000e+00])
%so this image structure is indeed low rank:
%1 nonzero singular valyue
%
%
%Now try:
%y=np.tril(np.ones(100).reshape([-1,10]));
%[u,s,vt]=np.linalg.svd(y); s
%array([6.690745  , 2.2469796 , 1.36858433,
%1.        , 0.80193774, 0.68207997,
%0.60515194, 0.55495813, 0.52324637,
%0.50564767])
%So suddenly all singilar values are
%non negligible. So SVD only represent
%structures which are aligned vertically
%and horizontally. The same applies to
%all tensors. So they are quite poorly
%suited for exploiting low rank (because x
%and y have exactly the same content
%(and edge) up to a rotation, so they
%are in reality equally low rank)
%
%The solution is to go to curvelets, shearlets,
%wavelet dictionaries... or to impose an
%indirect measure of low rankness.
%\end{verbatim}
we are able to discover the hidden
components in the high-dimensional data, so as to model the data more
accurately.

The tensor is the generalization of the matrix and vector concept:
a vector is a first-order or one-way tensor,
and a matrix is a second-order tensor.
To measure the low-rankness of tensors, a lot of the current work is to
decompose the tensor into a combination of several factors to explore
its low-rank structure through preliminary tensor decomposition, or to
unfold the target tensor into matrices according to the modal, and then
directly applies the rank of matrix or the sparsity of vector to the
resulting matrixes.
Popular decompositions include Tucker \cite{tucker1966some},
Canonical Polyadic (CP) \cite{CP} and tensor SVD (t-SVD) \cite{tensor_kilmer2013third}.
%However, there is little knowledge about the attributes of Tucker and CP levels.
In addition,
there are also some models that further improve the above decomposition or approximation methods.
For example,
some works impose specific prior constraints on the factors obtained by these decompositions \cite{KBR, Xue_Enhanced_Tucker}.
%But there is little work that directly explores ***WHat does
%``directly'' explore mean? and what is a ``modality'' The
%techniquesare indeed well known and not widely used in image
%processing, except in remote sensing (but remote sensing image
%processing is often a bit behind)*** new modalities for
%tensor decomposition.

\begin{figure*}[!t]
	\centering
	\includegraphics[width=0.9\linewidth]{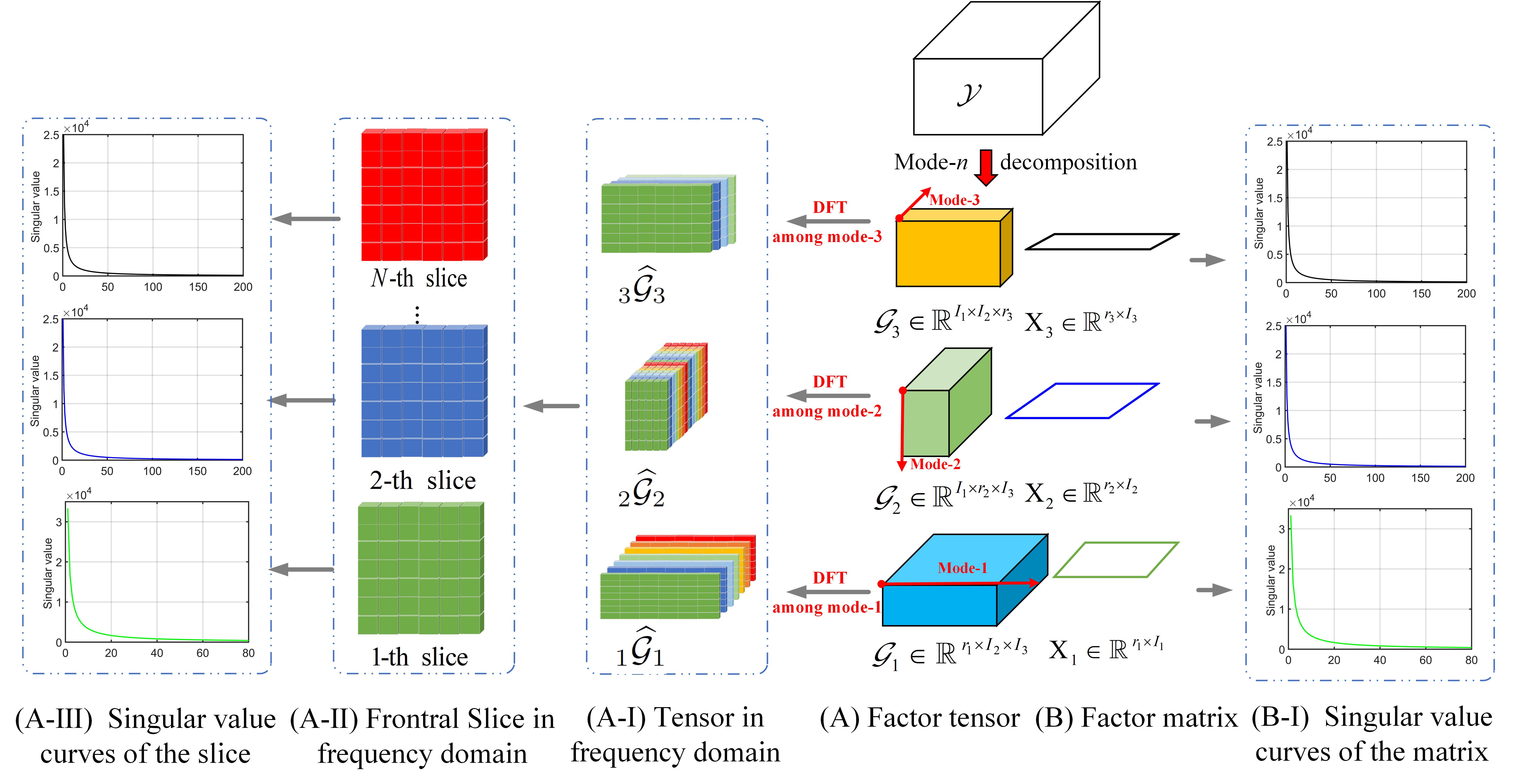}
	\caption{ (A)-(B) Original tensor $\mathcal{Y}$ and the factor tensor $\mathcal{G}_n$ and factor matrix $\mathbf{X}_n$ obtained by MCTF decomposition along its three modes, (B-\uppercase\expandafter{\romannumeral1}) the singular value curve of the factor matrix $\mathbf{X}_n$, (A-\uppercase\expandafter{\romannumeral1}) the tensor $_n\widehat{\mathcal{G}}_n$ is generated by performing DFT along the mode-$n$ fiber of the factor tensor $\mathcal{G}_n$. (A-\uppercase\expandafter{\romannumeral3}) shows the singular value curve from the first modal slice of $_n\widehat{\mathcal{G}}_n$ to the final $N$ slices in (A-\uppercase\expandafter{\romannumeral2}).}
	\label{fig:tensor_matrix_factor_LR}
\end{figure*}

According to the well-established theory of rank function in matrix case, it seems natural to directly extend matrix completion
methods to the tensor completion problem. However, it has been proven that calculating such a tensor rank
(whether it is based on Tucker, CP or T-SVD decomposition) is an
NP-hard problem \cite{tensor_rank_NPhard}.
It is difficult to determine or even limit  
%***??? This is not
%possible unless you mean ``approximate a tensor by a low rank one''
%(but that is not what you say** 
the rank of arbitrary tensors compared to matrix rank,
due to tensor low-rankness insight 
%***I don;t understand this entire
%sentens***
should be explained beyond the low-rank properties of all its expanded subspaces, and more importantly, 
how these subspace low-rank properties are related to the entire tensor structure should also be considered \cite{cichocki2015tensor}.
Some of current works directly extend the rank of a matrix to higher-order by simply summing up ranks 
%***I doubt this is correct; probably it is
%``max'' or something**** 
(or its relaxations) along all tensor modes \cite{cao2015folded, HaLRTC, romera2013new},
or directly consider the global low-rankness of underlying tensors.
Different from the matrix scenarios,
the simple rank summation term is generally short of a clear physical meaning for tensors \cite{KBR}. 
Furthermore, when the sampling rate is very low, it is also not sufficient to explore only global low-rankness.
As shown in Fig. \ref{fig:tensor_matrix_factor_LR}, after exploring the global low-rank prior through tensor decomposition, for the factor obtained by the decomposition, instead of being identical global low-rankness, the low-rankness in different modes or orientations are evidently exsisted and different. Actually,
from Fig. \ref{fig:tensor_matrix_factor_LR}-(B-\uppercase\expandafter{\romannumeral 1}), one can see that most singular values of the factor matrices are very close to zero, and much smaller than the first several larger singular values. 
Moreover,
instead of being independent, as shown in Fig. \ref{fig:tensor_matrix_factor_LR}-(B-\uppercase\expandafter{\romannumeral 3}), there are apparent correlations across
different slices of the each mode of the factor tensor and matrix. 
%***It is not at all clear how the figure shows this. You should explain***

%In \cite{tensor_ill_posedness} the ill-posedness of the best low-rank approximation of tensors is studied,
%while in \cite{tensor_rank_bounds} the upper and lower limits of the tensor rank are studied.

In this paper, for the tensor completion problem, we propose a novel tensor low-rankness measure to effectively model
multi-modal low-rankness of high-order tensors.
Similar to Tucker, our method is also based upon the
tensor and matrix decomposition definitions. 
But instead of using the tensor Tucker decomposition directly, and require the components to be orthogonal, firstly,
a novel tensor decomposition is proposed, in which a high-order
low-rank decomposition is introduced into each mode of the underlying
tensor, but the factors is not required to be orthogonal, as shown in the Fig. \ref{fig:tensor_matrix_factor_LR}(A)-(B). 
%***But this is not at all
%clear from the figure itself. So you just confuse tehreader by
%referring to the figure***
Instead of utilizing only one mode low-rankness of the underlying tensor as tensor nuclear norm based on T-SVD, this decomposition utilize all mode low-ranknesses of the tensor to give much better performance. Compared with Tucker decomposition, our method do not require the components to be orthogonal, thus, no need to use the SVD in our decomposing algorithms, which is computationally much cheaper than Tucker and T-SVD.

Secondly, a tensor low-rankness measure based on the proposed
decomposition is proposed, which combines both the low-rank prior of
the global tensor and the local factors obtained by the proposed tensor decomposition method. 
Its insight can be easily interpreted as a regularization for the factor tensor and matrix derived from the non-orthogonal Tucker decomposition.
Furthermore, an alternative convex relaxation of the proposed low-rankness measure is presented.
Such measure not only unifies the traditional understanding of low-rankness from matrix
to tensor, but also encodes both sparsity
insights delivered by common Tucker, SVD and T-SVD low-rank decompositions for a general tensor.

Thirdly,
we applied the proposed low-rank measures to high-dimensional
tensor completion tasks, e.g., video, hyperspectral image, MRI completion, and designed a block successive upper-bound
minimization (BSUM) method to efficiently solve the resulting
models.
%, and prove that our numerical scheme converges to the coordinatewise minimizers.
The validity of the proposed models are evaluated on a series of experiments, including  video, MRI and hyperspectral image completion.

%Hence, the key contributions of this work are:
%
%\begin{enumerate}[]
%  \item .
%
%  \item .
%
%  \item .
%\end{enumerate}

\section{Notions and Preliminaries}\label{notation}

In this section, we summarized some notations, tensor operations and operators used in this paper.

%***The notation $\langle\mathcal{X}, \mathcal{X}\rangle$ is undefined***
\begin{table}
	\centering
	\caption{Notational convention in this paper}
	\label{table:notations}
	\begin{tabular}{|c|c|}
		\hline
		Notation & Definition\\
		\hline
		$\mathbf{x}, \mathbf{y}$ &Vectors\\
		$\mathbf{X}, \mathbf{Y}$&Matrices\\
		$\mathcal{X}, \mathcal{Y}$&Tensors\\
		$x_{i_{1}, \cdots, i_{N}}$&$(i_{1}, \cdots, i_{N})$-th entry of $\mathcal{X}$\\
		$\operatorname{unfold}_{n}(\mathcal{X})=\mathbf{X}_{(n)}$&Mode-$n$ unfolding of $\mathcal{X}$\\
		$\operatorname{fold}_{n}\left(\operatorname{unfold}_{n}(\mathcal{X})\right)=\mathcal{X}$&The inverse of $\operatorname{unfold}_{n}$\\
		$\operatorname{rank}_{n}(\mathcal{X}) = \operatorname{rank}(\mathbf{X}_{(n)})$&The $n$-rank of $\mathcal{X}$ \\
		\hline
	\end{tabular}
\end{table}

Table \ref{table:notations} summarizes the common notations throughout this
paper.
Following \cite{Tmac, zeng2020tensor}, a \textbf{fiber} of
tensor $\mathcal{X} \in \mathbb{R}^{n_{1} \times n_{2} \times n_{3}}$ is defined as a vector obtained by fixing all
indices of $\mathcal{X}$ except one, and a \textbf{slice}
of $\mathcal{X}$ is defined as a matrix 
%***
%THis is a very
%unconventional definition. The rank is officially defined as the number of
%rank-1 tensor in a decomposition. ``vector ranks'' should be called
%``multi-rank.'' Rank is a number, not a
%matrix. Also, you say it is a matrix but then you write down a  vector
%instead. This is very confusing
%*** 
by fixing all indices
of $\mathcal{X}$ except two.
$\hat{\mathcal{X}}$ denotes  the result of discrete Fourier transformation (FFT) of $\mathcal{X}$ along the 3-rd dimension, and $\mathcal{X}$ can be computed from $\hat{\mathcal{X}}$ via the inverse FFT.
Then the multi-rank of $\mathcal{X}$ is defined as the array $\operatorname{rank}(\mathcal{X})=(\operatorname{rank}(\hat{\mathbf{X}}_{(1)}), \ldots, \operatorname{rank}(\hat{\mathbf{X}}_{(N)}))$, where $\hat{\mathbf{X}}_{(N)}$ denotes the rank of the $N$-th frontal slice of $\hat{\mathcal{X}}$.
Specially, for a 3-way tensor $\mathcal{A} \in \mathbb{C}^{n_{1} \times n_{2} \times n_{3}}$, its $(i, j, k)$-th entry is denoted as $a_{i j k}$ and $\mathcal{A}(i,:,:)$, $\mathcal{A}(:, i,:)$ and $\mathcal{A}(:,:, i)$ represent the $i$-th horizontal, lateral and frontal slice, respectively.

Based on these common notations of tensor, one can define \emph{inner product, t-product} and \emph{n-mode product}. 

\emph{Definition 1 (inner product} \cite{kolda2009tensor}): For $\mathcal{X}, \mathcal{Y} \in \mathbb{R}^{I_{1} \times \cdots \times I_{N}}$, their inner product is defined as
\begin{equation}
	\langle\mathcal{X}, \mathcal{Y}\rangle=\sum_{i_{1}=1}^{I_{1}} \cdots \sum_{i_{N}=1}^{I_{N}} x_{i_{1}, \cdots, i_{N}} y_{i_{1}, \cdots, i_{N}},
\end{equation}
and $\|\mathcal{X}\|_{\text{F}}=\sqrt{\langle\mathcal{X}, \mathcal{X}\rangle}$ denotes the Frobenius norm of $\mathcal{X}$.

%***
%It is not at all clear why the dft is suddenly introduced in a
%paper on tensors. You have to explain why you introduce it? In general
%image processing in the fourier domain is not a very good approach. It
%has been abandoned 30 years ago
%*** 

%\emph {Definition * (T-Product):} The t-product between  $\mathcal{A} \in
%	\mathbb{R}^{n_{1} \times n_{2} \times n_{3}} \text { and } \mathcal{B} \in \mathbb{R}^{n_{2} \times n_{4} \times n_{3}} \text { is defined as } \mathcal{A} * \mathcal{B}=
%	\operatorname{fold}(\operatorname{bcirc}(\mathcal{A}) \cdot \operatorname{unfold}(\mathcal{B})) \in \mathbb{R}^{n_{1} \times n_{4} \times n_{3}}.$

\emph{Definition 2 (T-product} \cite{kilmer2013third}): Given $\mathcal{X} \in \mathbb{R}^{d_{1} \times d_{2} \times d_{3}}$ and $\mathcal{Y}\in \mathbb{R}^{d_{2} \times d_{4} \times d_{3}}$, their t-product $\mathcal{T}=\mathcal{X} * \mathcal{Y} \in$ $\mathbb{R}^{d_{1} \times d_{4} \times d_{3}}$ is a tensor whose $(i, j)^{\text {th }}$ fiber $\mathcal{T}(i, j,:)=$ $\sum_{k=1}^{d_{2}} \mathcal{X}(i, k,:) \bullet \mathcal{Y}(k, j,:)$, where $\bullet$ is the circular convolution.

%
%\textcolor{red}{***You should define this concept first!***}

%***what is t-prod?***

%\textcolor{red}{The definitions of $\overline{\mathcal{L}}$ and bcirc(A) are the basis of tensor rank and nuclear norm that will be introduced subsequently.}

%\subsection{Operators} \label{operators}
%The \textbf{Proximal Operator} of a given convex function $f(x)$ is defined as
%\begin{equation}
%\label{PPA}
%	\operatorname{prox}_{f}(x,y):=\arg \min _{x} f(x)+\frac{\rho}{2}\|x-y\|^{2},
%\end{equation}
%where $\rho$ is a positive constant.
%Friendly, the problem $\arg\min _{x}\{f(x)\}$ is equivalent to
%$$\arg\min _{x, y}\left\{f(x)+\frac{\rho}{2}\|x-y\|^{2}\right\}.$$
%Thus, one can obtain the minimization of $f(x)$ by iteratively solving prox$_{f}\left(x, x^{k}\right)$, where $x^{k}$ is the latest update of $x$.
%The highlight of the proximal operator is that it can guarantee the strong convexity of objective function (\ref{PPA}), as long as $f(x)$ is convex.

\emph {Definition 3 ($n$-mode (matrix) product)} \cite{kolda2009tensor}: The $n$-mode product of tensor
$\mathcal{X} \in \mathbb{R}^{I_{1} \times I_{2} \times \cdots \times I_{N}}$
with a matrix $\mathbf{U} \in$ $\mathbb{R}^{J \times I_{n}}$,
denoted by $\mathcal{X} \times{ }_{n} \mathbf{U} \in \mathbb{R}^ {I_{1} \times \cdots \times I_{n-1} \times J \times I_{n+1} \times \cdots \times I_{N}}$,
is defined as
$$
\left(\mathcal{X} \times{ }_{n} \mathbf{U}\right)_{i_{1} \cdots i_{n-1} j i_{n+1} \cdots i_{N}}=\sum_{i_{n}=1}^{I_{n}} x_{i_{1} i_{2} \cdots i_{N}} u_{j i_{n}} .
$$

\emph{Definition 4 (Conjugate transpose} \cite{TRPCA}): The conjugate transpose of tensor $\mathcal{A} \in \mathbb{C}^{n_{1} \times n_{2} \times n_{3}}$ is the tensor $\mathcal{A}^{*} \in \mathbb{C}^{n_{2} \times n_{1} \times n_{3}}$ obtained by conjugate transposing each of the frontal slices and then reversing the order of transposed frontal slices 2 through $n_{3}.$

%\begin{figure}
%	\centering
%	\includegraphics[width=0.8\linewidth]{myfigure/Tensor_decomposition}
%	\caption{}
%	\label{fig:tensordecomposition}
%\end{figure}

\section{Related Works on Tensor Low-Rankness}

%In this section,
%we briefly reviewed the highly related topics:
%Tensor decomposition and related tensor low-rankness measure.
%
%\subsection{Tensor Decomposition}

Tensor completion is to fill in the missing values of a tensor $\mathcal{Y} \in \mathbb{R}^{n_{1} \times n_{2} \times n_{3}}$ under a given subset $\Omega$ of its entries $\left\{\mathcal{Y}_{i j k} \mid(i, j, k) \in \Omega\right\}$. Since tensor data of high dimensional are usually underlying low-rank \cite{TCTF}, the formulation of tensor completion can be written as
$$
\min _{\mathcal{C}} \operatorname{rank}_{t}(\mathcal{X}), \quad \text { s.t. } P_{\Omega}(\mathcal{X}-\mathcal{Y})=\mathbf{0}
$$
where $\operatorname{rank}_{t}(\mathcal{X})$ denotes the tubal rank of $\mathcal{C}$ and $P_{\Omega}$ is the linear operator that extracts entries in $\Omega$ and fills the entries not in $\Omega$ with zeros, i.e., 
\begin{equation}
\nonumber
\left(\mathcal{P}_{\Omega}(\mathcal{Y})\right)_{i_{1} \cdots i_{N}}=\left
\{\begin{array}{ll}
{y_{i_{1}, \cdots, i_{N}},} & {\left(i_{1}, \cdots, i_{N}\right) \in \Omega} \\
{0,} & {\text { otherwise }}
\end{array}\right.
\end{equation}

A tensor is a high-dimensional extension of a two-dimensional matrix, and on the other hand, it can also be reordered into a two-dimensional matrix. Therefore a natural tensor filling technique is to unfold the tensor into a matrix and use matrix-based filling methods to achieve its filling and then fold it back to the original high-dimensional tensor. There are two main categories of such methods: low-rank matrix decomposition methods (LRMF) and rank minimization techniques. The principle of low-rank matrix decomposition is to decompose the target matrix into two planar matrices to achieve the inscription of low-rank prior, while rank minimization achieves this by directly imposing an additional rank constraint on the matrix to be estimated \cite{buchanan2005damped, candes2011robust, eriksson2010efficient, ke2005robust}.

Although the method of reordering the tensor into a matrix is computationally efficient, this method of dimensionality reduction inevitably destroys the intrinsic structure of the tensor. For example, unfolding hyperspectral or multispectral images along the spectral dimension, and those unfolding videos along the temporal dimension, will destroy the spatial information of each frequency/time band/frame of these data \cite{KBR}. Therefore, in the past ten years, a lot of work has focused on completing the task of tensor completion by directly imposing low-rank and sparse constraints on the target tensor.  Motivated by the great success of matrix nuclear norms and decomposition, their promotion in the form of tensors has aroused more and more research interests, and there have been many results, e.g.,  tubal nuclear norm (TNN) \cite{TNN} and partial sum of the tubal nuclear norm (PSTNN) \cite{PSTNN}, Tucker rank based on Tucker decomposition, CANDECOMP/PARAFAC (CP) rank based on CP decomposition, framelet based TNN (FTNN) \cite{FTNN}.
Among existing tensor nuclear norms and tensor decomposition, 
%Given a real tensor $\mathcal{Y} \in R^{I_{1} \times I_{2} \times \cdots \times I_{N}}$,
%tensor decomposition attempts to decompose it as a combination of several factor tensors, matrices or vectors.
%Among the popular decompositions, 
Tucker decomposition, TNN and their extensions has shown superior performance in various applications, such as image/video inpainting/denoising \cite{ZENG_HSI_tensor, LRTDTV, xue2019nonlocal}, clustering \cite{xie2018unifying} and WiFi
fingerprint-based indoor localization \cite{Wifi_Tensor}.

%\textcolor{red}
%{***Justify why tucker decomposition is usful in
%(multispectral?) image processing and why it should be discussed
%here.****}

In Tucker decomposition \cite{tucker1966some}, an $N$-order tensor $\mathcal{X} \in \mathbb{R}^{I_{1} \times \cdots \times I_{N}}$ can be written as the following form:
\begin{equation} \label{equation:Tucker}
\mathcal{X}=\mathcal{S} \times_{1} \mathbf{U}_{1} \times_{2} \mathbf{U}_{2} \times_{3} \cdots \times_{N} \mathbf{U}_{N}
\end{equation}
where $\mathcal{S} \in \mathbb{R}^{R_{1} \times \cdots \times R_{N}}\left(r_{i} \leq R_{i} \leq I_{i}\right)$ is called the core tensor,
and $\mathbf{U}_{i} \in \mathbb{R}^{I_{i} \times R_{i}}(1 \leq i \leq N)$ is composed by the $R_{i}$ orthogonal bases along the $i$-th mode of $\mathcal{X}$.
With this Tucker formula, high-order low-rankness can be quantified as a vector $\left(r_{1}, r_{2}, \cdots r_{N}\right)$, i.e., Tucker rank.
%When the subspace bases along each mode is sorted based on their importance for tensor representation,
%the values of elements of the core tensor will show an approximately descending order along each of tensor modes.
%Under such a Tucker formulation, high-order low-rankness can be quantified as the vector $\left(r_{1}, r_{2}, \cdots r_{N}\right)$,
%which is often called Tucker rank.
%Tucker rank considers the low-rank property of the vector subspace unfolded along each of its modes as a reflection of high-order low-rankness\cite{KBR}.
%Such a low-rankness understanding is equivalent to consider the size of nonzero-block in core tensor,
%which represents the coefficients affiliated from all combinations of the used subspace bases.
The degree of freedom of the above-mentioned Tucker decomposition is $\prod_{i=1}^{N} r_{i}$, which uses the volume of the core tensor to evaluate the low-rankness of the underlying tensor.
However, the core tensor obtained by the decomposition of natural data usually has a low-rank structure, which causes this degree of freedom to be further restricted to a smaller number \cite{sparse_tucker, KBR}.
Therefore, it is difficult for Tucker rank to take reasonable measures to fully describe the inherent low-rank priors of tensors.

%\textcolor{red}{***In the next equation what is the meaning of ``*'' . * does ot mean
%multiplication. It means convolution. What is the meaning of the other
%``*'' in $V^*$? It is not possible to understand the text without
%clear notations****}

TNN is induced by the T-SVD \cite{tensor_kilmer2013third} which attempts to decompose an third-order tensor as a tensor product of three factor tensors. Specifically, let $\mathcal{A} \in \mathbb{R}^{n_{1} \times n_{2} \times n_{3}}$, then, by using T-SVD, it can be factorized as
\begin{equation} \label{equation:TSVD}
\mathcal{A}=\mathcal{U} * \mathcal{S} * \mathcal{V}^{*}
\end{equation}
where $\mathcal{U} \in \mathbb{R}^{n_{1} \times n_{1} \times n_{3}}, \mathcal{V} \in \mathbb{R}^{n_{2} \times n_{2} \times n_{3}}$ are orthogonal,
and $\mathcal{S} \in \mathbb{R}^{n_{1} \times n_{2} \times n_{3}}$ is a f-diagonal tensor, which is defined as a tensor whose frontal
slices is a diagonal matrix. 
$\mathcal{U}*\mathcal{S}$ and $\mathcal{V}^*$ are the T-product of $\mathcal{U, S}$, conjugate transpose of $\mathcal{V}$, respectively, which are defined in Section \ref{notation}.
% ***
% what is a frontal slice?
% ***

%\textcolor{red}{
%***this makes no sense: wy does a `` a unique calculation method''
%make something successful? How can a tensor decompisition ``explore
%characteristics of data?'' (only humans can do that). Is this method
%relevant for image processing?***}

%With the help of Fourier transform, TNN has a natural computing advantage for data with strong continuity along a certain dimension \cite{TRPCA}.
TNN induced method reported success on various applications in recent years \cite{lu_tensor_2016_lrtdtv_22_trpca, tensor_tubal_rank, tensor_t_product, twist_TNN}.
%The operation of Fourier transform along the third dimension makes TNN based models have a natural computing advantage for video and other data with strong time continuity along the third dimension \cite{Tensor_Q_Rank}.
%However,
%when considering the smoothness of different data,
%applying Fourier transform to the fixed third dimension may bring some limitations.
%Specifically,
%the experiments in related papers \cite{TNN, tensor_tubal_rank, TCTF} are usually focus on special datasets with smooth changes among the third dimension, such as RGB images and short videos.
%However, it cannot be ignored that these non-smooth data may increase the number of non-zero singular values \cite{factorization_3dTensor, TNN},
%thereby weakening the importance of low-rank structures.
%Due to tensor multi-rank \cite{TNN} is actually the rank of each Fourier projection matrix on a different basis \cite{Tensor_Q_Rank},
%non-smooth changes along the three dimensions may result in larger singular values on high-frequency projection matrix slices.
It utilizes the low rank of the tensor spectrum, which can well
capture the spatial-temporally smooth.
However, by calculating the nuclear norm of the frontal slice
after 1-D DFT on the mode-3 fiber, it is sensitive to the orientation of mode
%***orientation of what?***
and cannot capture the complex intra-mode and inter-mode correlation
of tensors in multiple directions. 

%\textcolor{red}{***yes but this is impossible
%anyway. Tensor decompositions can only capture correlation along the direction of
%coordinate axes.***}

\begin{figure}[H]
	\centering
	\subfloat[]{\includegraphics[width=0.21\linewidth]{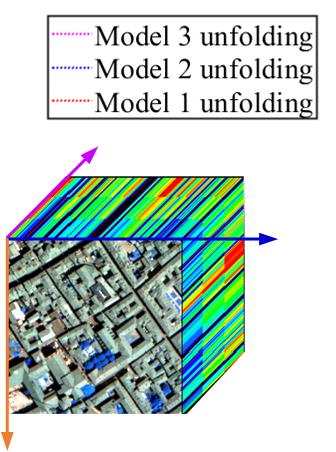}}
	\subfloat[]{\includegraphics[width=0.35\linewidth]{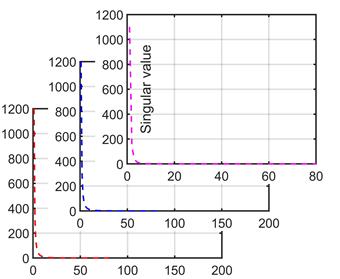}}
	\subfloat[]{\includegraphics[width=0.4\linewidth]{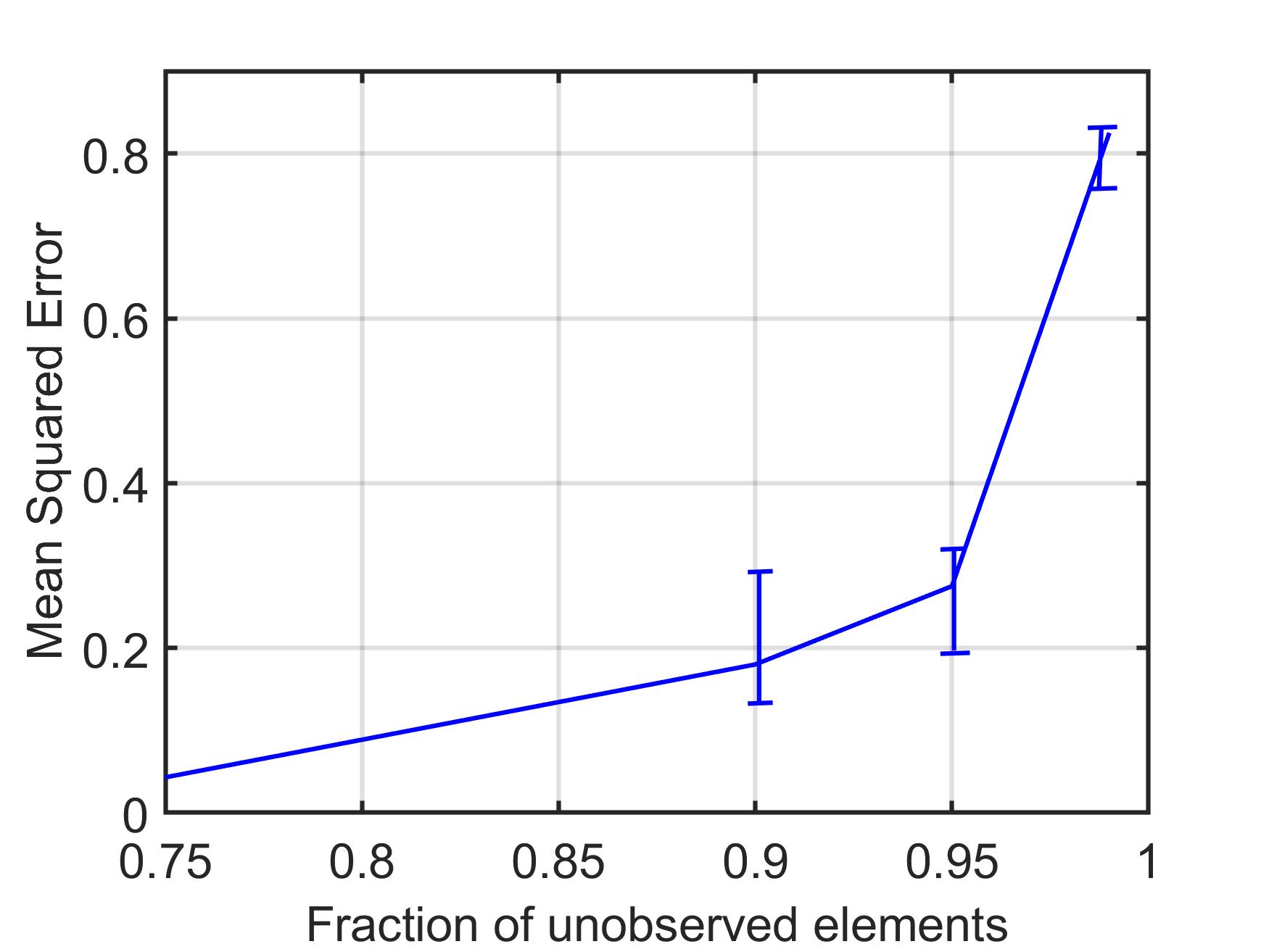}}
	\caption{(a)
		A real HSI of size 200*200*80; (b) Singular value curves of the matrices unfolded along three tensor modes. (c) The tensor completion performance by T-SVD decomposition (TNN) for the video “Suzie” dataset. Prediction accuracy severely degenerates when observations are sparse.}
	\label{fig:tnndisadvantage}
\end{figure}

The tensors collected from real scenes often have obvious
correlations along each of its modes. Taking HSI as an example,
Fig. \ref{fig:tnndisadvantage}-(b)
%***the figure is not very clear: (c) mentions ``unobserved
%elements''. What is an element? And why would an element be
%``unobserved''?
%****
shows the singular value curve diagram of the three modes of Fig. \ref{fig:tnndisadvantage}-(a).
From the figure, it can be quantitatively 
%*** how can you observe this? You should explain***
% and intuitively
%  ***????*** 
observed that only a small fraction of singular values of the three mode unfolding matrices are greater than zero, 
which means that the three modes along its spectrum and spatial are correlated.
This reflects the fact that the tensor along each mode is located on the low-rank subspace, and the entire tensor corresponds to the membership of the subspace along all tensor modes.
%The low-rank property of the subspace along each mode can be easily observed from the dramatic decreasing effect of these curves;
%Fig. \ref{fig:tnndisadvantage}(a) is a HSI cube, 
%Fig. \ref{fig:tnndisadvantage}(b) shows the plot of the singular value curves for its three modes. 
%The significant drop of these singular value curves can be seen in the figure, meaning that one can easily observe the low-rank properties of the subspace along each mode, i.e., the three modes along its spectrum and spatial are correlated.
These facts motivate us to define TNN along different dimension as the
natural intuitive meaning.
%****????? meaning of what????***

On the other hand, in real scenarios that the data representation
along a meaningful factor 
%***when is a factor meaningful?***
 (e.g., $\mathcal{S, U, V}$ in
(\ref{equation:Tucker})
%***That equation does not even contain $\mathcal{V}$ ***
or (\ref{equation:TSVD})) should always has an evident correlation and thus a low-rank structure \cite{KBR},
such a useful knowledge, however, can not be well expressed by Tucker
or T-SVD decomposition.
%***After reading this sentence three times, I still do not understand
%it: ``that the data representation ...and should  have. Part of the
%sentence seems missing.***
To ameliorate this issue, we attempt to propose a measure for more
rationally measuring low-rankness of tensor.

%***The problem is that the ``issue'' you are trying to solve is not at
%all clear. The link with image processing problems is completely unclear****

%\subsection{Tensor Decomposition based Low-Rankness Measure}
%Mathematically, a low-rankness based tensor recovery model can generally be expressed as follows:
%\begin{equation}
%\label{equation:low-rank-measure}
%\min _{x} F_{LR}(\mathcal{X})+\tau F_{DFT}(\mathcal{X}, \mathcal{Y})
%\end{equation}
%where
%$\mathcal{X, Y} \in R^{I_{1} \times I_{2} \times \cdots \times I_{N}}$
%denote the original tensor and observed tensor, respectively.
%$F_{DFT}(\mathcal{X}, \mathcal{Y})$
%is the data fidelity term,
%$F_{LR}(\mathcal{X})$ defines the tensor low-rankness measure of $\mathcal{X}$, $\tau$ is a non-negative regularization parameter used to balance two regular terms.
%The core problem of (\ref{equation:low-rank-measure}) is to design an appropriate low-rank measure for tensor data. There have been many mature research results on low-rank measures in the case of vector/matrix, but it is difficult to extract reasonable high-rank low-rank measures for tensors.

%In addition,
%the single nuclear norm or its relaxation is usually adopted to approximate the rank function, which would lead to suboptimal solution deviated from the original one \cite{double_nuclear_norm_CVPR}.

\section{MCTF Decomposition based Tensor Low-Rankness Measure}

Here we first introduce the details of our multi-modal core tensor
factorization model, and then introduce the low-rankness measure metric based on it and a better non-convex relaxation form of the low-rankness measure.
%\textcolor{red}{Finally, we compare our method with the previous work.}

\subsection{MCTF Decomposition}
\label{Section:proposed model}

Specifically, tensor low-rankness insight should be interpreted beyond the low-rank property of all its unfolded subspaces,
and should more importantly consider how such subspace low-rankness
are affiliated 
%***means: belinging to an organisation. What do you mean?*** 
over the entire tensor structure, especially when the
elements in the tensor to be restored are seriously missing.
%***SOmething is missing or not. ``seriously missing'' is
%meaningless... This paper is about tensor completion, but  the apper
%has not mentioned even a single application of tensor completion in image processing...
%***
For example,
Fig. \ref{fig:tnndisadvantage}-(c) shows the prediction errors by T-SVD decomposition against the fraction of unobserved elements for a particular HSI dataset, i.e., Pavia City Centre\footnote{http://www.ehu.es/ccwintco/index.php/Hyperspectral$\_$Remote$\_$Sensing$\\\_$Scenes}.
It can be seen that when there is less data missing,
the error remains a small state,
but when there is more data missing, the error starts to increase dramatically.

\begin{figure*}
	\centering
	\includegraphics[width=1\linewidth]{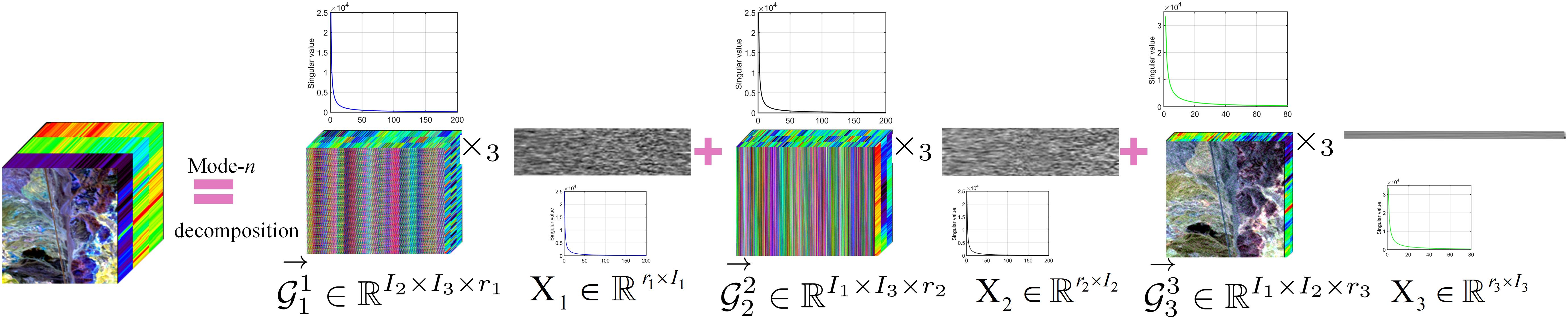}
	\caption{Illustration of the the proposed MCTF tensor decomposition for 3D tensors, which encourages simultaneously low rank structure in all orientations.}
	\label{fig: MCTF_addnorm}
\end{figure*}

%****This sentence is so complicated. I cannot understand it. Split in
%3 or 4 sentences***
To reduce the sensitivity to direction and simultaneously capture the complex intra- and
inter-modal correlations of high-order tensors in multiple directions, 
improve the limited representation ability and flexibility 
of the tensor decomposition model in multi-origin correlation modeling, 
we first propose an omnidirectional tensor decomposition strategy, called the multi-modal core tensor factorization (MCTF) model, by employing multilinear techniques. 
%\textcolor{red}{***What is multilinear? All decompositions in
%this paper are non-linear***}

\noindent \textbf{Definition 2.1 (MCTF)}. Given an $N$-way tensor $\mathcal{Y} \in \mathbb{R}^{I_{1} \times I_{2} \times \cdots \times I_{N}}$, as Tucker our MCTF decomposition
decomposes the input tensor as follows: 
\begin{equation}\label{equation:MMCT}
  \begin{aligned}
\mathcal{Y} &=w_1(\mathcal{G}_1 \times_{1} \mathbf{X}_{1}) + w_2(\mathcal{G}_2 \times_{2} \mathbf{X}_{2}) + \cdots + w_n(\mathcal{G}_N \times_{N} \mathbf{X}_{N})\\
 &=\sum_{n=1}^{N}w_n(\mathcal{G}_n \times_{n} \mathbf{X}_{n}),
\end{aligned}
\end{equation}
however, the main
difference is that we do not require the components $\mathbf{X}_{n}$ to be orthogonal, and where $\mathbf{X}_{n} \in \mathbb{R}^{I_{n} \times r_{n}}$ is the $n$-th ($n=1,2, \cdots, N$) factor matrix
%An illustration of the peoposed MCTF can be found in Fig. \ref{fig:ourtensormatrix}.
%\begin{figure}[H]
%	\centering
%	\includegraphics[width=0.95\linewidth]{myfigure/tensor_matrix_decomposition-1}
%	\caption{Illustration of (a) Vector case representation. (b) Matrix case
%		decomposition.(c) tensor CP decomposition and (d) the proposed tensor MCTF decomposition.}
%	\label{fig:tensordecomposition}
%\end{figure}
which reflecting 
%****???? sentence is also grammatically incorrect*** 
the
connections (or links) between the latent components and factor
matrices,  
%***I was just thinking that this was just the Tucker
%decomposition. So why can't you write: ``As tucker our method
%decomposes the input tensor as follows: .... However the main
%difference is that we do not require the components to be orthogonal.''
%I don't understand if there is any other difference.***
$\mathcal{G}_n \in \mathbb{R}^{I_1 \times \cdots \times I_{n-1} \times r_{n} \times I_{n+1} \times I_{N}}$
is a tensor reflecting 
%***???***  
the joint connections 
%***what is a joint connection?*** 
between the latent components in each mode.
$w_n (n=1,2,\cdots,N$) are positive weights satisfying $\sum_{n=1}^{N}w_{n}=1$.
%***how is this different from non-negative tucker decomposition?***
Tucker decomposition imposes the condition of all-orthogonality, instead of diagonality, on tensor $\mathcal{G}_n$, implies that the Tucker is always defined. 
As a matter of fact, $\mathcal{G}$ cannot be diagonal in general, which means that the Tucker does not necessarily reveal the rank of $\mathcal{Y}:$ in the cases where $\mathcal{G}$ is diagonal, the orthogonality of the matrices of mode- $n$ singular vectors implies that
$
\mathcal{Y}=\sum_{i_{n}}^{R_{n}} g_{i_{n} i_{n} \ldots i_{n}} X_{i_{n}}^{(1)} \circ X_{i_{n}}^{(2)} \circ \ldots \circ X_{i_{n}}^{(N)}
$
is a decomposition in a minimal number of rank-1 terms. On the other hand, the number of non-zero (significant) mode-$n$ singular values corresponds to the mode-$n$ rank (in a numerical sense) of $\mathcal{Y}$ \cite{Non_orthogonal}.

\begin{figure*}
	\centering
	\includegraphics[width=1\linewidth]{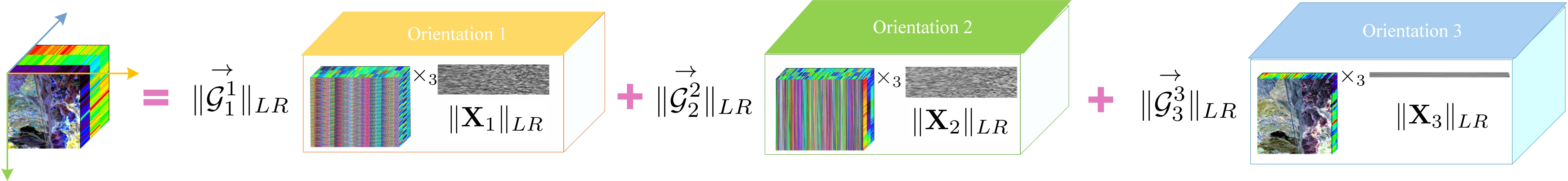}
	\caption{Illustration of the proposed MCTF based low-rankness measure, which models the underlying tensor as a mixture of three low rank combinations of tensor and matrix.}
	\label{fig: MCTF_LR_SVD}
\end{figure*}

MCTF encourages a low-rank structure, which means
low-rankness in spectral domain of all orientations.
It models a data tensor as simultaneously low tubal rank in all orientations (See Fig. \ref{figure:one example} and \ref{fig: MCTF_LR_SVD}).
It differs from TNN which only considers low tubal rank of one spectral orientation.
On the other hand, the proposed MCTF can be regarded as a generalization of matrix
factorization in the form of tensor.
%****???? Do you mean: ``in the form of tensors when ....'' or ``in the
%form of tensor. When....''? In the first case I don't understand.
When the factor tensor $\mathcal{G}_n$ degenerates into a matrix, MCTF degenerates into a classic matrix factorization.
In addition, MCTF is also a generalized form of Tucker decomposition,
if we additionally require the factor matrix $\textbf{X}_n$ to be orthogonal, then MCTF is equivalent to Tucker-1 decomposition.

%\begin{figure}
%	\centering
%	\includegraphics[width=1\linewidth]{"myfigure/tensor_matrix_LR_example_new_small_Noborder"}
%	\caption{Illustration of MCTF and the low-rankness measure based it for 3D tensors (a): MCTF encourages simultaneously low tubal rank structure in all orientations; (b): low-rankness measure based on MCTF.}
%	\label{fig:tensormatrixlrexample---new}
%\end{figure}

For ease of reading, while facilitating the algorithm implementation,
we further simplified the form of MCTF.
%***remove all the matlab references. Mathematics and programming
%should never be mixed and you should not advertise commercial software***
Firstly, we defined a
"tensor permutation" operation $\overrightarrow{\mathcal{X}}^{k}$ to
rearrange the dimensions of an tensor. 
%***Yes, you could define a
%million of things, but the reader is thinking''why does he define all
%of these things?  Are they really needed for something or will the rest of the
%paper be a long list of definitions without ever reaching the point of
%the paper?****
Specifically, for a tensor $\mathcal{X} \in \mathbb{R}^{n_{1} \times n_{2} \times n_{3}}$, $\overrightarrow{\mathcal{X}}^{k}$ is defined as the tensor whose $i$ th mode-3 slice is the $i$-th mode-$k$ slice of $\mathcal{X}$., i.e., $\mathcal{X}(i, j, s)=$ $\overrightarrow{\mathcal{X}}^{1}(j, s, i)=\overrightarrow{\mathcal{X}}^{2}(s, i, j)=\overrightarrow{\mathcal{X}}^{3}(i, j, s) .$
%With Matlab commond, it can be represented as $\overrightarrow{\mathcal{X}}^{k}:=\operatorname{permute}(\mathcal{X}, [n_1  \quad n_2 \quad n_k \quad ... \quad n_{k-1} \quad n_3 \quad n_{k+1} ...])$.
Then, by using this permutation, the modal-$n$ product of the tensor and the matrix can be uniformly transformed  into the modal-3 product, and so the MCTF can be rewritten as follows:
\begin{equation}
\begin{aligned}
\mathcal{Y} &=w_1(\mathcal{G}_1 \times_{1} \mathbf{X}_{1}) + w_2(\mathcal{G}_2 \times_{2} \mathbf{X}_{2}) + \cdots + w_n(\mathcal{G}_N \times_{N} \mathbf{X}_{N})\\
&=\sum_{n=1}^{N}w_n(\overset{\rightarrow}{\mathcal{G}_n^n} \times_{3} \mathbf{X}_{n}).
\end{aligned}
\end{equation}

%***So if you only need this simplified notation, then why did you
%introduce the more general one? It makes the paper difficult to read
%***

%
%$$\|\mathbf{X}_1\|_{LR}$$
%
%$$\|\overset{\rightarrow}{\mathcal{G}_1^1}\|_{LR}$$
%
%
%$$\|\mathbf{X}_2\|_{LR}$$
%
%$$\|\overset{\rightarrow}{\mathcal{G}_2^2}\|_{LR}$$
%
%
%
%
%$$\|\mathbf{X}_3\|_{LR}$$
%
%$$\|\overset{\rightarrow}{\mathcal{G}_3^3}\|_{LR}$$

%$$\overset{\rightarrow}{\mathcal{G}_1^1}\in \mathbb{R}^{I_{2} \times I_{3} \times r_{1}}$$
%
%
%
%$$\overset{\rightarrow}{\mathcal{G}_2^2}\in \mathbb{R}^{I_{1} \times I_{3} \times r_{2}}$$
%
%
%$$\overset{\rightarrow}{\mathcal{G}_3^3}\in \mathbb{R}^{I_{1} \times I_{2} \times r_{3}}$$

\subsection{MCTF based Tensor Low-rankness Measure}

\begin{figure}
	\centering
	\captionsetup[subfloat]{labelsep=none,format=plain,labelformat=empty} %\captionsetup[subfloat]{labelsep=period}
	\subfloat[80\% Masked]{\includegraphics[width=0.22\linewidth]{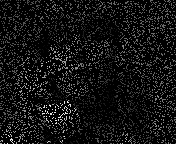}}%
	\hfil
	\subfloat[TMac]{\includegraphics[width=0.22\linewidth]{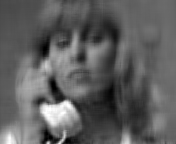}}%
	\hfil
	\subfloat[TNN]{\includegraphics[width=0.22\linewidth]{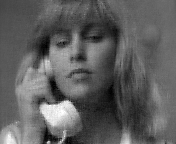}}%
	\hfil
	\subfloat[NC-MCTF]{\includegraphics[width=0.22\linewidth]{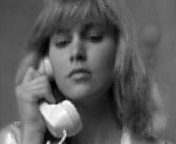}}%
	\hfil
	\subfloat[95\% Masked]{\includegraphics[width=0.22\linewidth]{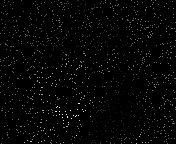}}%
	\hfil
	\subfloat[TMac]{\includegraphics[width=0.22\linewidth]{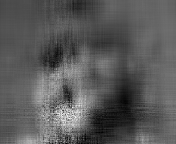}}%
	\hfil
	\subfloat[TNN]{\includegraphics[width=0.22\linewidth]{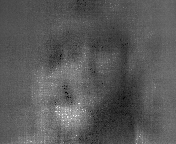}}%
	\hfil
	\subfloat[NC-MCTF]{\includegraphics[width=0.22\linewidth]{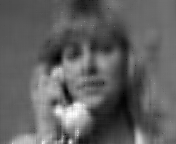}}%
	\caption{Two slices of the recovered video “Suzie” by TMac, TNN and our NC-MCTF. The sampling rate of the first line is 20\%, and the sampling rate of the second line is 5\%.}
	\label{figure:one example}
\end{figure}

Existing methods use either factorization or approximation schemes to recover the missing components.
However, as the number of missing entries increases, factorization schemes may overfit the model because of incorrectly predefined ranks,
while approximation schemes may fail to obtain easy to interpret model
factors.
%***Yes, but this is a completely artificial example. It does not seem
%to be related to any real image processing problem. So why should a
%journal accept this paper? The paper should first explain what
%real-world problems it is trying to solve. Then it should motivate why
%low-rank tensor approximation is a good solution (my opinion: it will
%work, but it is not the best method, it is needlessly complicated
%from a practical point of view. While the theory is nice, it does not
%suit the properties of real image data well.***
Taking the video "Suzie" as an example, as shown in the Fig. \ref{figure:one example},
it can be seen that when the sampling rate is large, all methods can restore a clear image.
But when the sampling rate is very low, that is, when there are few known entries, both the TMac and TNN failed to recover the main information of the image.

Fortunately, there are extra priors that we can utilize, i.e., the
model structure is implicitly included in the low-rank factorization
model,
%***But this sentence contradicts itself: ``extra priors'' versos
%``implicitly included''. If structure is already included, then there
%is no need for extra priors. Plus: what do you mean?***
according to the factor priors,
which are usually known a priori in real-world tensor objects.
Fig. \ref{fig: MCTF_addnorm} is an example of low rank factors.
From the figure, one can observe that the factor tensor obtained by the
proposed MCTF decomposition is close to low rank along all the mode
slices.
%******What does this mean: ``which shows that the spatial and spectral correlation can also
%be clearly observed from the factor tensor.'' How can you observe this
%correlation (spatial or spectral as opposed to general)?
%***
%***It is also a problem that you start discussing results without even
%having explained the method. So far the only ting we know about the
%method is that it is a non-orthogonal tucker decomposition. But what
%criterion is optimised? How is it optimized? ...***
Meanwhile, similar to the
factor tensor, the factor matrix is also low-rank.
%***paragraphs are too long, maing paper difficult to read***
This fact inspires
us to further explore the structure 
%***what is ``characterise the
%structure?''*** 
of the factors obtained by the
proposed decomposition, and then better describe the low-rank nature
of the original tensor.

\begin{figure}
	\centering
	\includegraphics[width=1.0\linewidth]{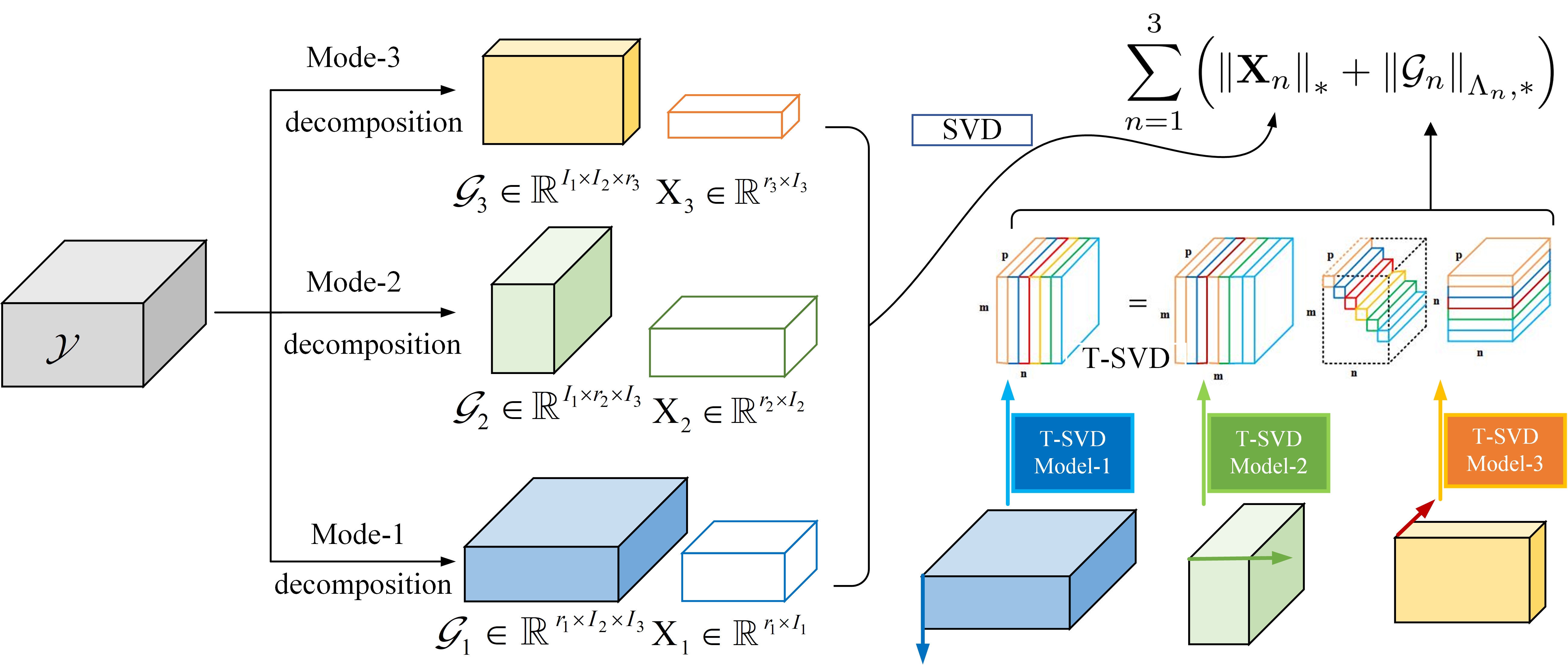}
	\caption{A visual display of the proposed tensor low-rank measure.}
	\label{fig:ourtensormatrix}
\end{figure}

To this end, based on the proposed MCTF, we are motivated to we designed the multiple transform domains based tensor
nuclear norm regularization 
%***so is this another method than what you first proposed?*** 
for the factor tensor obtained by the decomposition, together with classic
matrix nuclear norm for the factor matrix,
to represent the underlying joint-manifold drawn from the model factors, and finally proposed a low-rankness measure for tensor $\mathcal{Y}$ based the proposed MCTF, i.e.,
\begin{equation}\label{equation:convex_MCTF}
\begin{aligned}
\mathcal{S}(\mathcal{Y}) = \sum_{n=1}^{3}&(\tau_n \left\|\mathbf{X}_{n}\right\|_{\text{*}}+\lambda_n \left\|\mathcal{G}_{n}\right\|_{\Lambda_n,\text{*}}),
\end{aligned}
\end{equation}
%\begin{equation}
%\label{equation:convex_MCTF}
%\begin{aligned}
%\min \sum_{n=1}^{3}& \frac{\alpha_{n}}{2}\left\|\mathcal{Y}-\mathcal{G}_{n} \times_{n} \mathbf{X}_{n}\right\|_{\text{F}}^{2}+\tau_n \left\|\mathbf{X}_{n}\right\|_{\text{*}}+\lambda_n \left\|\mathcal{G}_{n}\right\|_{\Lambda_n,\text{*}},
%\end{aligned}
%\end{equation}
where $\mathbf{X}_{n}$ and $\mathcal{G}_n$ are the factor matrix and tensor of $\mathcal{Y}$ with MCTF, respectively;
$\alpha_{n}$ and $\tau_n$ is a parameter to tradeoff the two terms;
%***This equation is incomplete: what is minized over? Are the
%constants $\alpha_n$ changed to achieve the minimum? Or are they
%constants selected by the user? How should they be selected/***
$\left\|\mathbf{X}\right\|_{\text{*}}$ is matrix nuclear
norm, $\left\|\mathcal{X}\right\|_{\Lambda_n,\text{*}}$ denotes the
TNN of $\mathcal{X}$ based on transform domain $\Lambda_n$.
Applying transform domain $\Lambda_n$ to $\mathcal{X}$ is equivalent
to perform the DFT 
%***So it is single transform (only dft) instead of
%multitransform (dft, and some other type of transform)? *** 
along each mode-$n$ fiber of $\mathcal{X}$ (as
shown in the A and A-\uppercase\expandafter{\romannumeral1} column of
Fig. \ref{fig:tensor_matrix_factor_LR}), which can enhance the
flexibility 
%***why would applying a dft enhance ``flexibility.'' What
%is ``flexibility''?
for handling different correlations along different modes
and reduce the sensitivity to direction. 
%**Why would the dft make
%these transforms less sensitive to direction. A scientist should argue
%for choices, not just write all kinds of things without proof or even
%a reason why it would work***
An illustration of the proposed low-rankness measure can be found in Fig. \ref{fig:ourtensormatrix}.
%***How is this figure an example? The measure is the equation on the
%top right, but it seems unrelated to what you just
%explained. Specifically it does not agree with eq (11), because the
%first term is missing.****
As shown in the Fig. \ref{fig:ourtensormatrix}, the first operation in the proposed (\ref{equation:convex_MCTF}) is the MCTF decomposition of the underlying tensor, which serves as a complexity measure in the original domain for all orientations.
It is based on the matrix decomposition theory to extend the Tucker decomposition, and conforms to the internal mechanism of these two decompositions.
As shown in the Fig. \ref{fig: MCTF_LR_SVD},
the second term can be interpreted as the number of non-zero singular values of the factor matrix, and the third term models the TNN of $\mathcal{G}_n$ in all orientations, which measures low-rankness in the Fourier domain.
They tend to normalize the low-rank attributes across the subspace of each tensor mode.
%***What does this sentence mean? :***
This comprehensive consideration in the proposed measures is conducive
to explore the internal low-rank construction of the factor tensor and
the low-rank nature of the tensor quantum space along each
mode.
%Therefore, it is expected to alleviate the limitations of the
%above-mentioned Tucker and T-SVD decomposition.

\subsection{MCTF based Tensor Nonconvex Low-rankness Measure}

Although the proposed multiple transform domains based low-rankness
measure (\ref{equation:convex_MCTF})
can provide an efficient numerical solution
%***You can not just write this without any proof***
and report success on low rank completion experiments
%***Does this mean the results were published earlier and are not novel
%(Then where is the reference?)? Or do you mean that the
%results in this paper are novel (then it should be ``report'' not ``reported'') ***
(the detail performance about MCTF can be found in Section \ref{experiments}), we have to admit that it also has two shortcomings.
Firstly, the TNN in MCTF is essentially the nuclear norm of each spectral slice in the Fourier domain, measuring the $L_1 $ norm of non-zero singular values, which is not an ideal approximation of the tensor tubal rank.
Secondly, both TNN and NN treat each singular value equally, so the main information may not be well preserved.
%***what does this mean: ``so the main information may not be well
%preserved''? What is the ``main information?'' Why is it not
%preserved? And why is this a result of treating all singular values
%equally?''***
%
%***Also: in a non-orthogonal decomposition, the ``singular values''
%cannot be judged independently of each other. For instance, setting
%one of them to 0 means that the optimal values of all the other ones
%can change. Setting two big ones to zero may have no effect if they cancel
%each other, setting a small one to 0 may have a bigger effect than
%setting a large one to zero... None of these things are discussed in
%the paper?****
Due to larger singular values usually correspond to main information, such as contours, sharp edges and smooth areas, while smaller singular values are mainly composed of noise or outliers \cite {PSTNN, WNNM, logDet, IRLS}.
This means that singular values with different numerical values should
be treated differently, 
%***No, this is just semantics. Basically all
%methods set small values to zero and keep the big ones. They are all
%treated equally (same rule applies to all of them) but the outcome is
%different for small and big values.***
that is, the punishment for singular values
with large numerical values should be reduced, 
%***but large values are
%not punished in existing methods, so there is nothing to reduce*** 
and the punishment for
singular values with small numerical values should be
increased. 
%**they are already set to 0, so it is not possible to
%punish them more. Also words like ``punish'' are not appropriate for a
%scientific paper***

To overcome the above two shortcomings, 
%***which shortcomings?***  
we designed a novel tensor log-norm and matrix log-norm to perform non-convex relaxation of TNN
and NN respectively to more accurately describe the low-rank structure
of factor tensor and factor matrix,
%$$\sum_{n=1}^{3}\left(\left\|\mathbf{X}_{n}\right\|_{*}+ \left\|\mathcal{G}_{n}\right\|_{\Lambda_n,*}\right)$$
%
%$$\left\|\mathbf{X}_{2}\right\|_{*}+ \left\|\mathcal{G}_{2}\right\|_{\Lambda_2,*}$$
%
%$$\left\|\mathbf{X}_{3}\right\|_{*}+ \left\|\mathcal{G}_{3}\right\|_{\Lambda_3,*}$$
\begin{equation}
\begin{aligned}
\min \sum_{n=1}^{3}& \frac{\alpha_{n}}{2}\left\|\mathcal{Y}-\mathcal{G}_{n} \times_{n} \mathbf{X}_{n}\right\|_{\text{F}}^{2}+\tau_n \left\|\mathbf{X}_{n}\right\|_{\text{log}}+\lambda_n \left\|\mathcal{G}_{n}\right\|_{\Lambda_n,\text{log}},
\end{aligned}
\end{equation}
where
\begin{equation}
	\begin{aligned}
		\|\mathcal{G}_{n}\|_{\Lambda_n,\text{log}}=\|_n \overline{\mathcal{G}}_{n}\|_{\text{log}}&=\| \text { blockdiag }(_n\widehat{\mathcal{G}}_{n})\|_{\text{log}}\\
		&=\frac{1}{p} \sum_{i=1}^{p}\| _n\widehat{\mathcal{G}}_{n}^{(i)} \|_{\text{log}}
	\end{aligned}
\end{equation}
and
$
	\left\|\mathbf{X}\right\|_{\text{log}}=\sum_{i=1}^{\min \{m, n\}}\left(log\left(\left|\sigma_{i}\left(\mathbf{X}\right)\right|+\epsilon\right)\right),
$
for $\mathbf{X} \in \mathbb{R}^{m \times n}$, $\sigma_{i}(\mathbf{X})$ is the $i$-th singular value
of the matrix $\mathbf{X}$, $\epsilon>0$ is a constant. 
$\alpha_n, n=1, 2, 3$, are positive
weights satisfying $\sum_{n=1}^{3}\alpha_{n}=1$.
%***
%Avoid referring to commercial software in a paper. The paper should
%be about the science, not about implementation details and not about
%advertising.
%Also, if you introduce the
%notation $\widehat{\boldsymbol{v}}=\boldsymbol{F}_{n} \boldsymbol{v}$
%but then you don't use it and replace it by a new notation, which just
%conuses the reader. Replace  $\widehat{\boldsymbol{v}}=$
%fft $(\boldsymbol{v})$ everywhere by  $\widehat{\boldsymbol{v}}=\boldsymbol{F}_{n} \boldsymbol{v}$
%***
For $\mathcal{L} \in \mathbb{R}^{m \times n \times p},$ 
$_n\widehat{\mathcal{L}} \in \mathbb{C}^{m \times n \times p}$
denotes the result of Discrete Fourier Transformation (DFT)
on $\mathcal{L}\in \mathbb{R}^{m \times n \times p}$ along the $n$-th
dimension, i.e., 
$
\label{equation:computefft}
_n\widehat{\mathcal{L}}=\boldsymbol{F}_{n} \mathcal{L},
$
where $\boldsymbol{F}_{n}$ is the DFT matrix defined as
%\begin{equation}
%\boldsymbol{F}_{n}=\left[\begin{array}{ccccc}1 & 1 & 1 & \cdots & 1 \\ 1 & \omega & \omega^{2} & \cdots & \omega^{n-1} \\ \vdots & \vdots & \vdots & \ddots & \vdots \\ 1 & \omega^{n-1} & \omega^{2(n-1)} & \cdots & \omega^{(n-1)(n-1)}\end{array}\right] \in \mathbb{C}^{n \times n},
%\end{equation}
%where $\omega=\mathrm{e}^{-\frac{2 \pi i}{n}}$ is a primitive $n$-th root of unity in which $i=\sqrt{-1} .$
$
\boldsymbol{F}_{n}=\left[\boldsymbol{f}_{1}, \cdots, \boldsymbol{f}_{i}, \cdots, \boldsymbol{f}_{n_{3}}\right] \in \mathbb{R}^{n \times n}.
$
%where $\boldsymbol{f}_{i}=\left[\omega^{0 \times(i-1)} ; \omega^{1 \times(i-1)} ; \cdots ; \omega^{\left(n-1\right) \times(i-1)}\right] \in \mathbb{R}^{n}$
%with
%$ \omega=e^{-\left(2 \pi \sqrt{-1} / n\right)}.$
%***The  notations are unconventional (and it is always better to stick
%	to conventions. Usually $f$ is the frequency, not a column,
%	$\omega=2\pi\, f$ not $\exp(...)$.
%***A paper should use mathematics, not some programming language***}
%
%*** ****
%	***Explain why you need all of these notations. So far in the paper we
%	have not a clue about what the detailed contribution is about
%***
$\overline{\mathcal{L}}$ denotes the block-diagonal matrix of the tensor $\mathcal{L}$ in the Fourier domain, i.e.
%\begin{equation}
%\begin{array}{l}
%\overline{\mathcal{L}} = \text { blockdiag }(\widehat{\mathcal{L}}) \\
%=\left[\begin{array}{cccc}
%\widehat{\mathcal{L}}^{(1)} & & & \\
%& \widehat{\mathcal{L}}^{(2)} & & \\
%& & \ddots & \\
%& & & \widehat{\mathcal{L}}^{\left(p\right)}
%\end{array}\right] \in \mathbb{C}^{m p \times n p},
%\end{array}
%\end{equation}
$\overline{\mathcal{L}} = \text { blockdiag }(\widehat{\mathcal{L}}) 
=\text{diag}(\widehat{\mathcal{L}}^{(1)}, \widehat{\mathcal{L}}^{(2)}, \ddots, \widehat{\mathcal{L}}^{\left(p\right)})\in \mathbb{C}^{m p \times n p},
$
where $\widehat{\mathcal{L}}^{(i)}$ denotes the $i$-th frontal slices of $\widehat{\mathcal{L}}$, $i=1,2,\cdots,p$.
%***
%The log norm seems to have the opposite effect of what you claim:
%large singular values are punished more ($\log x$ becomes smaller
%than $x$ when $x$ becomes very large)
%***

\section{MCTF low-rankness measure-based model \& its solving scheme}
%***I did not read this in detail, but corrected some notational
%errors.
%
%Also, it seems that the optimisation methods are pretty standard
%(relaxation type of approach). They remind me of a pper by one of our
%former phd students. If this is not new, or an adaptation of existing
%work, then remove it from the paper
%(or shorten it) and refer to existing papers.
%
%It would have been better to replace this section by a better and more
%detailed explanation of the method itself. What are the principles
%behind it, i.e., why does it work (this remains mostly unclear. The
%text uses all kinds of meaningless words like ``exploit structures''or
%``punish this or that,'' but it provides very little insight in the ideas.
%***

Here we introduce the optimization of the proposed two models and analyze their convergence.

\subsection{The MCTF and NC-MCTF minimization models}

Before giving the optimization of MCTF and NC-MCTF, we first
introduce two lemmas as follows:

\emph {Lemma 1 (singular value shrinkage operator, SVT):}
For $\mathbf{M} \in \mathbb{R}^ {n_1 \times n_2}$,
$
\mathbf{M}=\mathbf{P}\mathbf{E}_{r} \mathbf{Q}^{\dag}
$
denotes the singular value decomposition (SVD) of matrix $\mathbf{M}$ with rank $r$, where $ \mathbf{E}_{r}=\operatorname{diag}(\left\{\sigma_{i}\right\}_{1 \leq i \leq r})$, $\sigma_{i}$ is the $i$-th largest singular value of $\mathbf{M}$. 
Then,
the following properties hold,
\begin{equation*}
\operatorname{D}_{\delta}(\mathbf{M})=\arg \min_{\operatorname{rank}(\mathbf{X}) \leq r} \delta\|\mathbf{X}\|_{*}+\frac{1}{2}\|\mathbf{X}-\mathbf{W}\|_{\text{F}}^{2},
\end{equation*}
where
$
\label{SVT}
\operatorname{D}_{\delta}(\mathbf{W})=\mathbf{P} \operatorname{diag}\left\{\max \left(\left(\sigma_{i}-\delta\right), 0\right)\right\} \mathbf{Q}^{\dag},
$ and $\|.\|_*$ is the matrix nuclear norm.
%****Multiple notations are undefined: what is $\delta$? What
%is $\|.\|_*$?****

%\textcolor{red}{***What is WNNM. All abbreviations should be written in full. More
%importantly: why do you introduce all of these concepts? You shuld
%first explain what the paper is menat to achieve. Only then present
%some concepts, and only present those that are actually essential.
%The current paper will probably only be read by ther people writing
%papers on tensor definitions because people will give up after seeing
%all of these complicated concepts, without knowing how they relate to
%actual image processing. The number of citations will therefore
%be limited.***}

\emph {Lemma 2 (weighted nuclear norm minimization, WNNM) }\cite{WNNM}: For any $\gamma>0, \mathbf{Y} \in \mathbb{R}^{m \times n}$ and $0 \leq d_{1} \leq d_{2} \leq \cdots \leq d_{r}(r=\min (m, n)),$ a global optimal solution to the following problem
\begin{equation}
\min _{\mathbf{X}} \sum_{j=1}^{r} \gamma d_{j} \sigma_{j}(\mathbf{X})+\frac{1}{2}\|\mathbf{Y}-\mathbf{X}\|_{F}^{2}
\end{equation}
is given by the following singular value thresholding
\begin{equation}
\label{equation:WNNM_operator}
\mathbf{X}^{*}=\mathbf{W}_{\gamma, d}(\mathbf{Y})=\mathbf{U} \mathbf{S}_{\gamma, d}(\Sigma) \mathbf{V}^{T}
\end{equation}
where $\mathbf{Y}=\mathbf{U} \Sigma \mathbf{V}^{T}$ is the SVD of $\mathbf{Y}$, $\sigma_{j}(\mathbf{X})$ denotes the $j$-th singular value of $\mathbf{X}$ and $\mathbf{S}_{\gamma, d}(\Sigma)_{jj}=\max \left(\Sigma_{j j}-\gamma d, 0\right)$.

%***The notation $\sigma_{j}(\mathbf{X})$ was never
%defined.  $\sigma_{j}$ ws defined, but that is not the same***

We then analyzing the optimization of the proposed models, 
the objective function of the proposed MCTF and NC-MCTF are listed as follows:
\begin{equation}
\begin{aligned}
f(\mathbf{X}, \mathcal{G}, \mathcal{Y})= \sum_{n=1}^{3}& \frac{\alpha_{n}}{2}\left\|\mathcal{Y}-\mathcal{G}_{n} \times_{n} \mathbf{X}_{n}\right\|_{\text{F}}^{2}+\tau_n \left\|\mathbf{X}_{n}\right\|_{\text{*}}\\
&+\lambda_n \left\|\mathcal{G}_{n}\right\|_{\Lambda_n,\text{*}},
\end{aligned}
\end{equation}
\begin{equation}
\begin{aligned}
f(\mathbf{X}, \mathcal{G}, \mathcal{Y})= \sum_{n=1}^{3}& \frac{\alpha_{n}}{2}\left\|\mathcal{Y}-\mathcal{G}_{n} \times_{n} \mathbf{X}_{n}\right\|_{\text{F}}^{2}+\tau_n \left\|\mathbf{X}_{n}\right\|_{\text{log}}\\
&+\lambda_n \left\|\mathcal{G}_{n}\right\|_{\Lambda_n,\text{log}}.
\end{aligned}
\end{equation}

%\begin{figure*}
%	\centering
%	\includegraphics[width=1.0\linewidth]{myfigure/factor_LR}
%	\caption{(a) An MSI $\mathcal{Y}$ and it’s Tucker decomposition. (b) Core tensor $\mathcal{G}$ of $\mathcal{Y}$ . Note that the size of the nonzeroblock is ***, and 78.4\% of its elements are zeroes. (c) Typical slices of S, where deeper color of the element represents a larger value of
%it. (d) 6 Kronecker bases of $\mathcal{Y}$ , which relate to the largest 6 elements in core tensor $\mathcal{G}$. (e) Noised MSI (lack of tensor sparsty) and it’s core tensor
%(the size of the nonzero-block is *** and most of its elements are nonzero).}
%	\label{fig:factorlr}
%\end{figure*}

The minimization of the proposed models are two complicated optimization problems, which are difficult to solve directly.
Here, we adopt the block successive upper-bound minimization (BSUM) \cite{BSUM} to solve them.

According to the proximal operator \cite{MFTV}, for the $k$-th iteration, the update can be written as follows:
\begin{equation}
\label{equation_original_PPA}
\operatorname{Prox}_{f}(\mathcal{S}, \mathcal{S}^k)= \argmin_{\mathcal{S}} f\left(\mathcal{S}\right)+\frac{\rho}{2}\left\|\mathcal{S}-\mathcal{S}^k\right\|_{\text{F}}^{2},
\end{equation}
where $\rho>0$ is the proximal parameter, $\mathcal{S}=(\mathbf{X}, \mathcal{G}, \mathcal{Y})$ and $\mathcal{S}^{k}=\left(\mathbf{X}^{k}, \mathcal{G}^{k}, \mathcal{Y}^{k}\right)$.

Let $S_{1}^{k}=\left(\mathbf{X}^{k}, \mathcal{G}^{k}, \mathcal{Y}^{k}\right)$,
$S_{2}^{k}=\left(\mathbf{X}^{k+1}, \mathcal{G}^{k}, \mathcal{Y}^{k}\right)$,
$S_{3}^{k}=\left(\mathbf{X}^{k+1}, \mathcal{G}^{k+1}, \mathcal{Y}^{k}\right)$.
By BSUM, (\ref{equation_original_PPA}) can be rewritten as follows:
\begin{equation}
\label{equation:XAY}
\begin{aligned}
\displaystyle \mathbf{X}^{k+1}=  \operatorname{Prox}_{f}\left(\mathbf{X}, \mathcal{S}_{1}^{k}\right)= \argmin_{\mathbf{X}} f\left(\mathbf{X}, \mathcal{G}^{k}, \mathcal{Y}^{k}\right)\\
+\frac{\rho}{2}\left\|\mathbf{X}-\mathbf{X}^{k}\right\|_{\text{F}}^{2}, \\
\displaystyle\mathcal{G}^{k+1}=  \operatorname{Prox}_{f}\left(\mathcal{G}, \mathcal{S}_{2}^{k}\right)= \argmin_{\mathcal{G}} f\left(\mathbf{X}^{k+1}, \mathcal{G}, \mathcal{Y}^{k}\right)\\
+\frac{\rho}{2}\left\|\mathcal{G}-\mathcal{G}^{k}\right\|_{\text{F}}^{2}, \\
\displaystyle\mathcal{Y}^{k+1}= \operatorname{Prox}_{f}\left(\mathcal{Y}, \mathcal{S}_{3}^{k}\right)= \argmin_{\mathcal{Y}} f\left(\mathbf{X}^{k+1}, \mathcal{G}^{k+1}, \mathcal{Y}\right)\\
+\frac{\rho}{2}\left\|\mathcal{Y}-\mathcal{Y}^{k}\right\|_{\text{F}}^{2}.
\end{aligned}
\end{equation}

%\begin{figure}
%	\centering
%	\includegraphics[width=1.0\linewidth]{myfigure/The_proposed_LR}
%	\caption{}
%	\label{fig:theproposedlr}
%\end{figure}

\subsubsection{Update $\mathbf{X}_n$ with fixing others}

By introducing one auxiliary variable $\mathbf{Z}_n$,
the $\mathbf{X}_n$-subproblem in (\ref{equation:XAY}) can be rewritten as
\begin{equation}
\label{equation_X_aux}
\begin{aligned}
\argmin_{\mathbf{X}_n, \mathbf{Z}_n}& \sum_{n=1}^{3} (\frac{\alpha_{n}}{2}\left\|\mathcal{Y}-\mathcal{G}_{n} \times_{n} \mathbf{X}_{n}\right\|_{\text{F}}^{2}+\tau_n \left\|\mathbf{Z}_{n}\right\|_{\text{*  or log}}\\
&+\frac{\rho_n}{2}\left\|\mathbf{X}_n-\mathbf{X}_n^{k}\right\|_{\text{F}}^{2}), s.t., \mathbf{X}_{n}=\mathbf{Z}_{n}.
\end{aligned}
\end{equation}
Based on the augmented Lagrange multiplier (ALM) method, the above minimization problem (\ref{equation_X_aux}) can be transformed into no-contrined problem, and be solved by SVT (\ref{SVT}) and $\mathrm{WNNM}$ operator (\ref{equation:WNNM_operator}):
\begin{equation}
\label{equation_solution_Zn}
\begin{aligned}
\mathbf{Z}_n^{k+1}=\operatorname{D}_{\frac{\tau_{n}}{\rho_n}}(\mathbf{X}_{n}^{k}+\Gamma_{n}^{\mathbf{X}}/\rho_n), n=1,2, \cdots, N;
\end{aligned}
\end{equation}
\begin{equation}
\label{equation_solution_Zn_log}
\mathbf{Z}_n^{k+1} = W_{\frac{\tau_{n}}{\rho_n}, \epsilon}\left(\mathbf{X}_{n}^{k}+\Gamma_{n}^{\mathbf{X}}/\rho_n\right), n=1,2, \cdots, N.
\end{equation} 
\begin{equation}
\label{equation:solution_Xn}
\begin{aligned}
\mathbf{X}_n^{k+1}&=(\alpha_{n}\mathbf{G}_n^T\mathbf{G}_n+2\rho \mathbf{I}_n)^{-1}[\alpha_{n}\mathbf{G}_n^T\mathbf{Y}_{(n)}\\
&+\mu_n (\frac{\mathbf{Z}_{n}^{k+1}-\Gamma_{n}^k/\mu_n+\mathbf{X}_n^{k}}{2}).
\end{aligned}
\end{equation}
where (\ref{equation_solution_Zn}) for MCTF, (\ref{equation_solution_Zn_log}) for NC-MCTF.
Based on the ALM method, the multipliers are updated by the following equations:
\begin{equation}
\label{equation:Lambda_0}
	\Gamma_{n}^{\mathbf{X}} = \Gamma_{n}^{\mathbf{X}} + \mathbf{X}_n-\mathbf{Z}_n.
\end{equation}

\subsubsection{Update $\mathcal{G}_n$ with fixing others}

By introducing an auxiliary variable, the $\mathcal{G}_n$-subproblem can be rewritten as
\begin{equation}
\label{equation_A_alm}
\begin{aligned}
\argmin_{\mathcal{G}_n} \sum_{n=1}^{3}& (\frac{\alpha_{n}}{2}\left\|\mathcal{Y}-\mathcal{G}_{n} \times_{n} \mathbf{X}_{n}\right\|_{\text{F}}^{2}+\lambda_n \left\|\mathcal{J}_{n}\right\|_{\Lambda_n,\text{*}}\\
&+\frac{\rho_n}{2}\left\|\mathcal{G}_n-\mathcal{G}_n^{k}\right\|_{\text{F}}^{2}),
s.t., \mathcal{G}_{n}=\mathcal{J}_{n}.
\end{aligned}
\end{equation}

By using the ALM and SVT operator (\ref{SVT}), one can also obtain the solutions: 
\begin{equation}
_n\widehat{J}_n^{k+1,(q)} = D_{\frac{1}{\rho_n}}\left(_n\widehat{\mathcal{U}}_n^{(q)}\right), ~q=1,2,\cdots,p.
\end{equation}
Then, the $(k+1)$-th updating of $\mathcal{J}^{k+1}_n$ can be obtained via inverse Fourier transform
\begin{equation}
\label{eq:Jmodel-solution}
\mathcal{J}_n^{k+1} =\operatorname{ifft}\left(_n\widehat{\mathcal{J}}_n^{k+1},[~], n\right).
\end{equation}

Similarly, the $\mathcal{J}_n$ related subproblem can be solved by the $\mathrm{WNNM}$ operator (\ref{equation:WNNM_operator}), i.e.,
\begin{equation}
_n\widehat{J}_n^{k+1,(q)} = W_{\frac{1}{\rho_n}, \epsilon}\left(_n\widehat{\mathcal{U}}_n^{(q)}\right), ~q=1,2,\cdots,p.
\end{equation}
Then, the $(k+1)$-th updating of $\mathcal{J}^{k+1}_n$ can be obtained via inverse Fourier transform
\begin{equation}
\label{eq:Jmodel-solution_1}
\mathcal{J}_n^{k+1} =\operatorname{ifft}\left(_n\widehat{\mathcal{J}}_n^{k+1},[~], n\right).
\end{equation}

With other variables fixed, the minimization subproblem for $\mathcal{G}_n$ is also convex and has the following closed-form solution
\begin{equation}
\label{equation_solution_An}
\begin{array}{r}
\mathcal{G}_n^{k+1}=\operatorname{fold}\left(\left(\mathbf{Y}_{(n)}^{k}\left(\mathbf{X}_{n}^{k+1}\right)^{T}+2\rho_n (\frac{\mathbf{J}_{n}^{k+1}-\Gamma_{n}^\mathcal{G}/\rho_n+\mathbf{G}_n^{k}}{2})\right)\right.\\
\left. \left(\mathbf{X}_{n}^{k+1}\left(\mathbf{X}_{n}^{k+1}\right)^{T}+2\rho_{n} \mathbf{I}_{n}\right)^{\dagger}\right), \\
n=1,2,\cdots, N.
\end{array}
\end{equation}

Finally, the Lagrangian multiplier can be updated by the following equations
\begin{equation}
\label{equation:Lambda_1}
\Gamma_{n}^{\mathcal{G}} = \Gamma_{n}^{\mathcal{G}} + \mathcal{G}_n-\mathcal{J}_n.
\end{equation}

\subsubsection{Update $\mathcal{Y}$ with fixing others}

The update of $\mathcal{Y}_{k+1}$ can be written explicitly as
\begin{equation}
\label{equation_solution_Y}
\begin{array}{l}
\displaystyle \mathcal{Y}^{k+1}=P_{{\Omega}^c}\left(\sum_{n=1}^{3} \alpha_{n} \text { fold }_{n}\left(\frac{\mathbf{G}_{n}^{k+1} \mathbf{X}_{n}^{k+1}+\rho_n \mathbf{Y}_{(n)}^{k}}{1+\rho_n}\right)\right)+\mathcal{F},
\end{array}
\end{equation}
where $\mathcal{F}$ is the observed data;
$P_{{\Omega}}$ is an operator defined in subsection \ref{notation}.

\begin{algorithm}
	\caption{:Algorithm for the proposed MCTF and NC-MCTF based tensor low-rankness measure.} \label{algorithm:A1}
	\begin{algorithmic}[1]
		\State \textbf{Input:} 
		The observed tensor $\mathcal{F}$;
		The set of index of observed entries $\Omega$;
		The given $n$-rank, $r = (r_1, r_2, r_3)$;
		stopping criterion $\varepsilon.$
		%		regularization parameters.
		%		$\lambda, \tau$, and $\mu$.
		\State \textbf{Outpot:}
		the completed tensor.
		\State Initialize:  $\mathbf{X}_n^0=\mathbf{Z}_n^0=\mathbf{0}, \mathcal{G}_n^0=\mathcal{J}_n^0=\mathbf{0}, \Gamma_{n}^\mathbf{X}=\mathbf{0},
		\Gamma_{n}^\mathcal{G}=\mathbf{0},  n=1,2,\cdots, N$; $ \mu_{\max }=10^{6}, \rho=1.5,$ $\mathcal{Y}=\mathcal{P}_{\Omega}(\mathcal{F})$, and $k=0$.
		\State Repeat until convergence:
		\State Update $\mathbf{X}, \mathbf{Z}, \mathcal{G}, \mathcal{J}, \mathcal{Y},  \Gamma^{\mathbf{X}}, \Gamma^{\mathcal{G}}$ via

		1st step: Update $\mathbf{Z}_n$ of MCTF via	(\ref{equation_solution_Zn}) or $\mathbf{Z}_n$ of
		
		 NC-MCTF via (\ref{equation_solution_Zn_log})

		2nd step: Update $\mathbf{X}_n$ via	(\ref{equation:solution_Xn})

		3rd step: Update $\mathcal{G}_n$ via (\ref{equation_solution_An})

		4th step: Update $\mathcal{J}_n$ of MCTF via	(\ref{eq:Jmodel-solution}) or $\mathcal{J}_n$ of 
		
		NC-MCTF via (\ref{eq:Jmodel-solution_1})

		5th step: Update $\mathcal{Y}$ via (\ref{equation_solution_Y})

		6th step: Update the parameter via 	(\ref{equation:Lambda_0}), (\ref{equation:Lambda_1})

		\State Check the convergence condition:
		$ \frac{\left\|\mathcal{Y}^{k+1}-\mathcal{Y}^{k}\right\|_{\text{F}}}{\left\|\mathcal{Y}^{k}\right\|_{\text{F}}}<\varepsilon$.
	\end{algorithmic}
\end{algorithm}

%\subsection{MCTF measure for tensor completion}
%*****************
%
%\subsection{MCTF model for tensor robust PCA}
%*****************
%
%\subsection{Applying MCTF algorithm to MSI Denoising}
%*****************

\subsection{Complexity Analysis}

In this subsection, the proposed algorithm for the proposed MCTF and NC-MCTF are summarized as Algorithm \ref{algorithm:A1}.
%Further, we discuss the complexity and convergence of the proposed algorithms.
Further, we discuss the complexity of the proposed algorithms.
Complexity Analysis:
The cost of computing $\mathbf{X}_{n}$ is $O\left(I_{n} r_{n}^{2}+I_{n} r_{n} s_{n}+r_{n}^{2} s_{n}\right)$;
calculating $\mathbf{Z_n}$ has a complexity of $O\left( \Pi_{j \neq n} I_{j} \times r_{n}^2 \right)$;
the complexity of updating $\mathcal{J}_n$ is $O\left(I_{n} r_{n}^2\right)$;
calculating $\mathcal{G}_{n}, n=1, 2, 3$, in both
MCTF-based solver and NC-MCTF-based solver, have a complexity of $O(I_{1} I_{2} I_{3}(\log (I_{1} I_{2} I_{3})+\sum_{n=1}^{3} \min (I_{n}, I_{n+1})))$, where we define $I_{4}=I_{1}$;
calculating $\mathcal{Y}$ has a complexity of $O(\sum_{n=1}^{N}r_{n} I_{n} s_{n})$.
Then, the total complexity of the proposed algorithms can be obtained by counting the complexity of the above variables.
For easily viewing, we list the total complexity of the proposed
models as follows:
\begin{equation}
	\begin{aligned}
\label{equation:complexity_model1}
	O(3I_{n} r_{n}^2+3 I_{n} r_{n} s_{n}+3 r_{n}^{2} s_{n}+I_{1} I_{2} I_{3}(\log (I_{1} I_{2} I_{3})\\
	+\sum_{n=1}^{3} \min (I_{n}, I_{n+1})))
	\end{aligned}
\end{equation}

\section{Numerical experiments} \label{experiments}

%\begin{table*} [!t]
%	\centering
%	\caption{Quantitative Picture Quality Indices and What are they evaluating}
%	\label{equation:PQIs}
%		\begin{tabular}{|c|c|}
%			\hline
%			&\\
%			Quantitative Picture Quality Indices (PQI)      &What are they evaluating	          \\
%			& \\
%			\hline
%			PSNR \cite{PSNR}	&The similarity between the target image		\\
%			SSIM \cite{SSIM}	&The reference image based on MSE and structural consistency		\\
%			FSIM \cite{FSIM}	        &The perceptual consistency with the reference image		\\
%			ERGAS \cite{EGRAS}	&Fidelity of the restored image based on the weighted sum of MSE in each band		\\
%			SAM \cite{SAM}	    &The similarity between spectral/temporal slice		\\
%			\hline
%	\end{tabular}
%\end{table*}

Three types of public tensor data-sets, i.e., video, MRI and hyperspectral
image, are selected for verification experiments to evaluate the
performance of the proposed model.
%***You have not even explained the details of the image processing
%problem you want to solve. Is it really just about some artificial
%experiment in which you set some values to zero and try to recover
%them? Then the method is of little interest. Or is there are real
%problem to solve? Where does the need for completion come from? Why is
%this problem important. Answers to these questions are essential for a
%good engineering paper***
Five state-of-the-art techniques proposed between 2013-2020, i.e., five tensor completion models related to the proposed models:
TMac(2013) \cite{Tmac}, MF-TV method(2016) \cite{MFTV},TNN(2016) \cite{TNN}, PSTNN(2020) \cite{PSTNN}  and FTNN(2020) \cite{FTNN}, were chosen for comparison.

Two types of standards for evaluation: qualitative visual evaluation of the restored data,
five widely used quantitative picture quality indices (PQIs): PSNR \cite{PSNR}, SSIM \cite{SSIM}, FSIM \cite{FSIM}, ERGAS \cite{EGRAS} and SAM, were utilized to assess the quality of restored tensor.
%The specifics are listed in Table \ref{equation:PQIs}.
All experiments were performed on MATLAB 2018b, Intel core i7@2.2GHz and 64.0 GB RAM.
For a tensor $\mathcal{Y} \in \mathbb{R}^{I_1 \times \ldots \times I_N}$,
The sampling ratio (SR) can be defined as:
$
\mathrm{SR}=\frac{S_{\text{number}}}{\prod_{n=1}^{N} I_{n}},
$
where $S_{\text{number}}$ denotes the number of sampled entries, $\Omega$ represents index set. The sampled entries are chosen randomly from a tensor $\mathcal{Y}$ by a uniform distribution.

\subsection{MRI}

In this subsection, to further verify the versatility of the proposed models for different datasets,
we conduct experiments on cubical MRI data\footnote{http://brainweb.bic.mni.mcgill.ca/brainweb/selection$\_$normal.html}
with size 150 $\times$ 150 $\times$ 181.
SRs are set as follows: 5\%, 10\%, 20\% and 30\%.
Here, we set the rank to $(T_1, T_2, T_3)$, where $T_1, T_2, T_3$
denote the number of the largest 0.5\% singular values of model-1,
model-2 and model-3, respectively.

%***In MRI there applications with ``missing samples'' but in that case
%the missing samples are not randomly selected. So from a practical
%point of view, this experiment is completely artificial***

\begin{figure}[!t]
	\centering
	\includegraphics[width=0.7\linewidth]{myfigure/tuli}
	\hfil
	\includegraphics[width=0.3\linewidth]{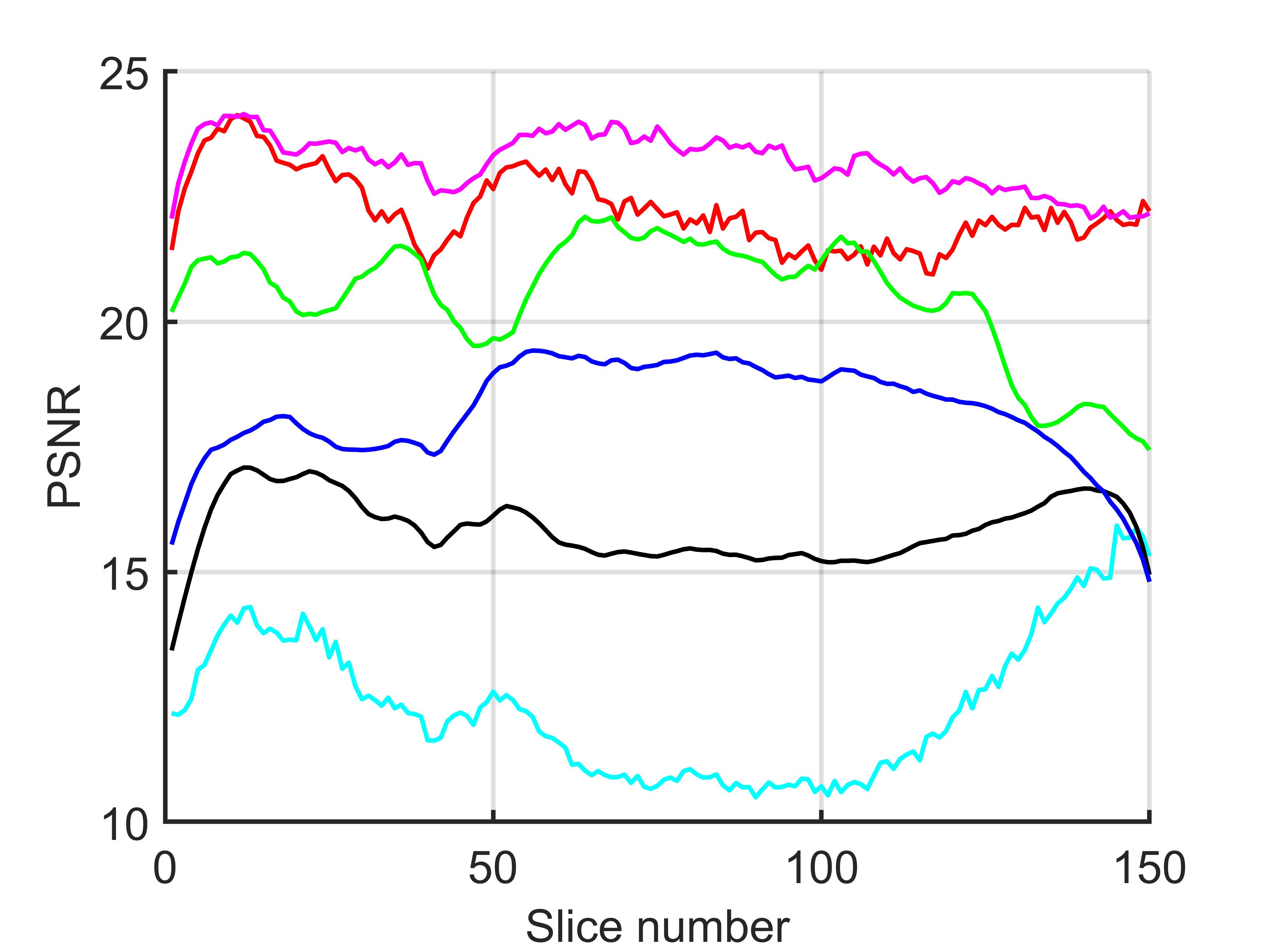}
	\hfil
	\includegraphics[width=0.3\linewidth]{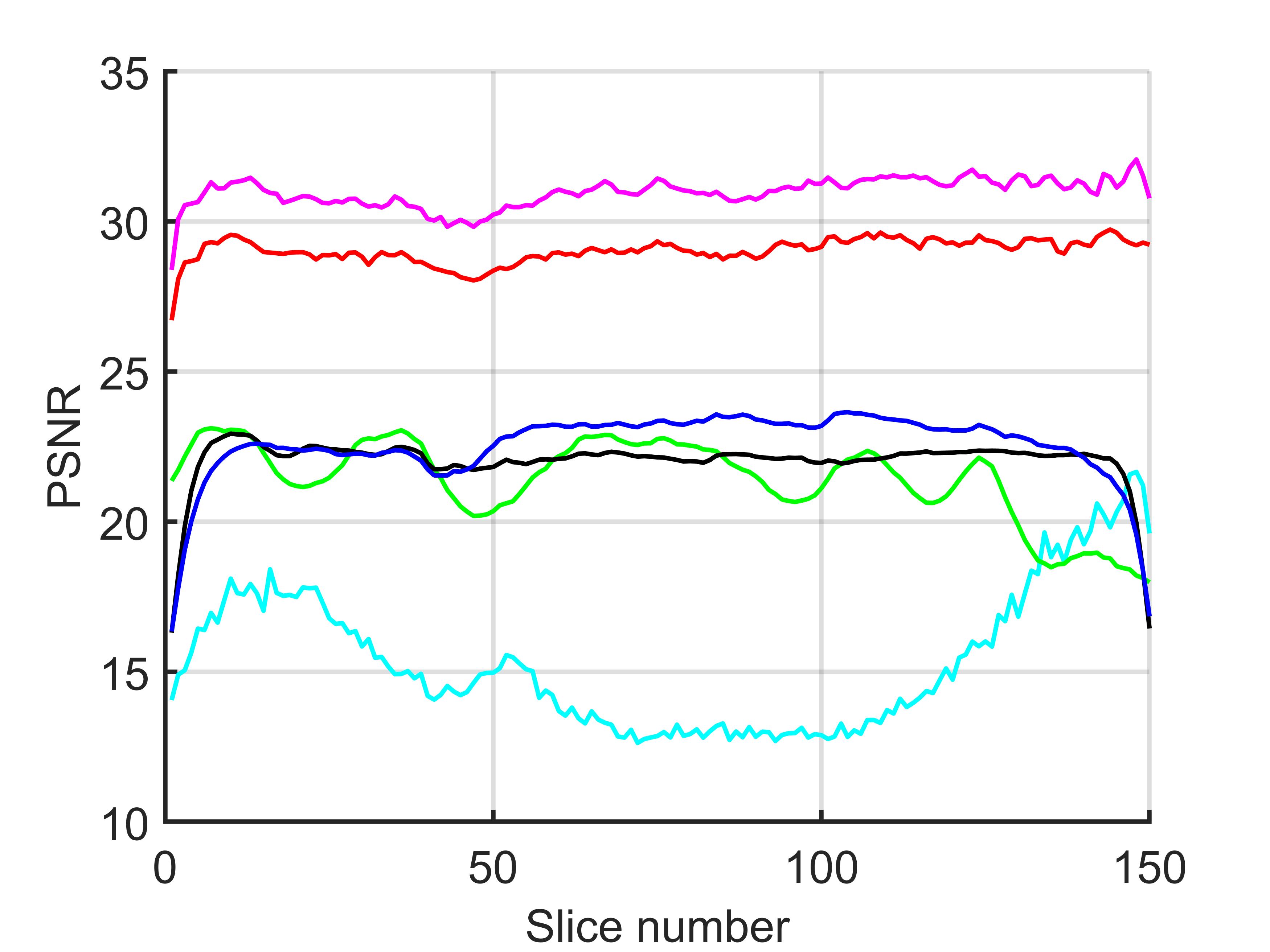}
	\hfil
	\includegraphics[width=0.3\linewidth]{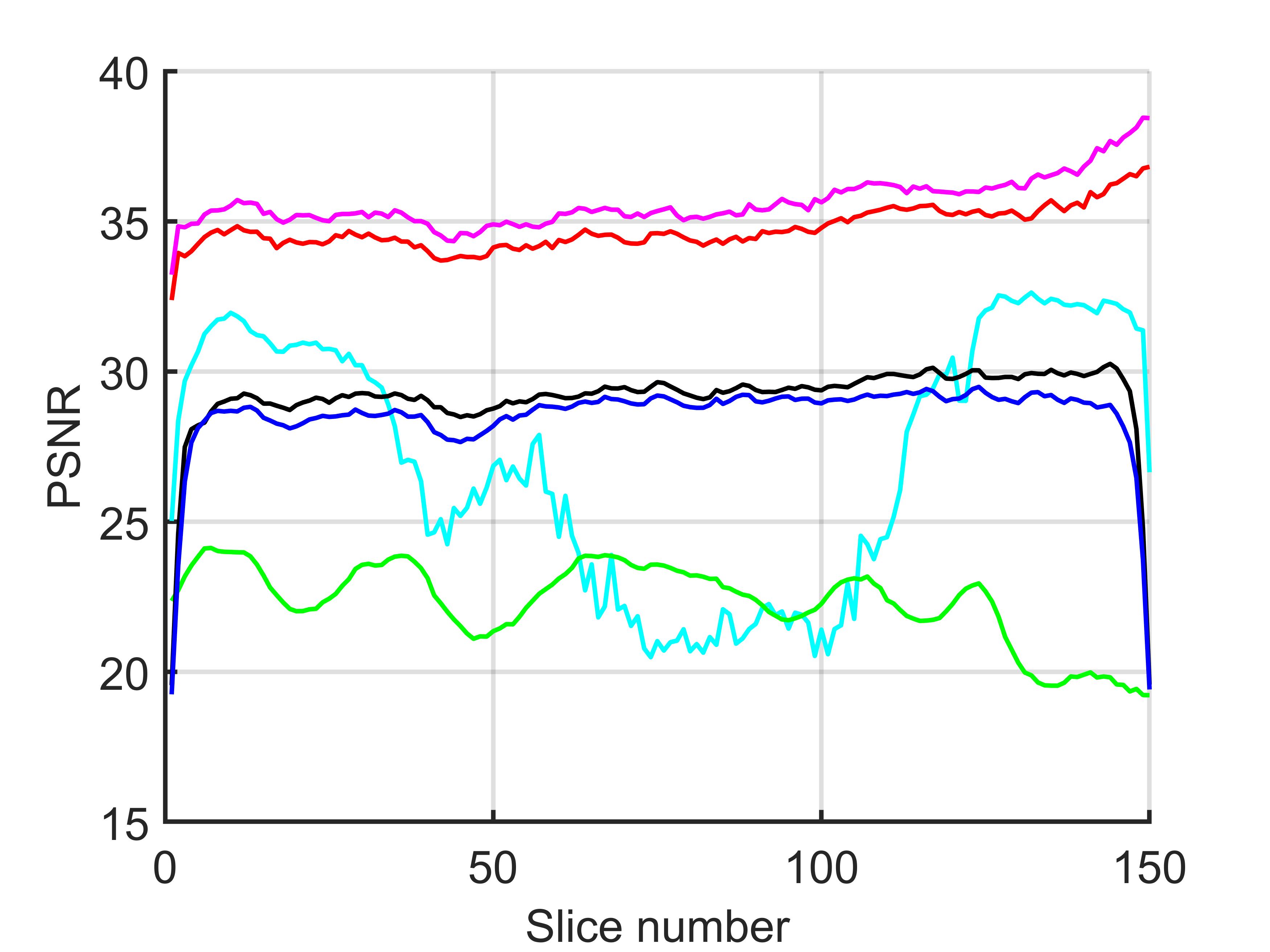}
	\hfil
	\includegraphics[width=0.3\linewidth]{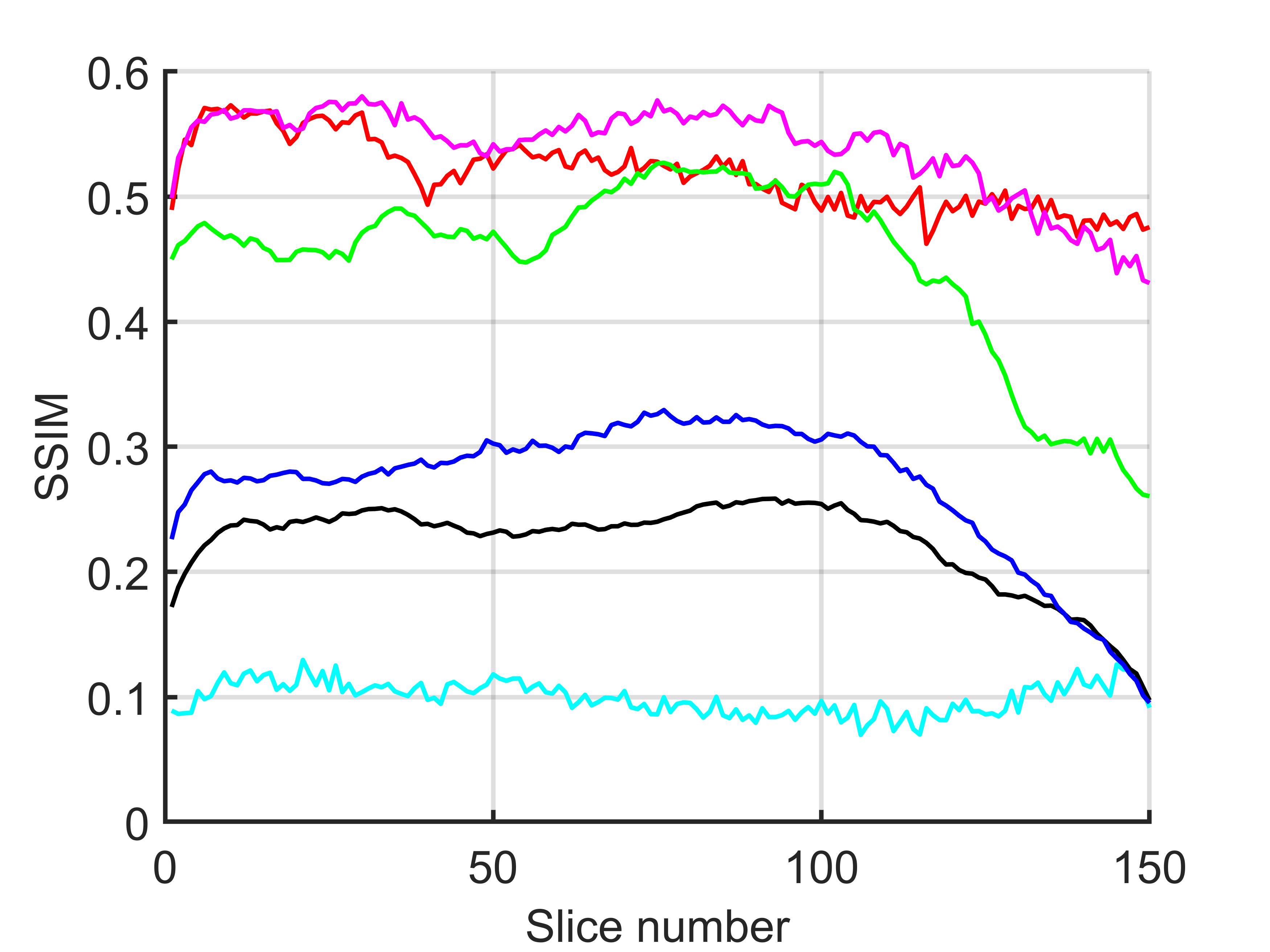}
	\hfil
	\includegraphics[width=0.3\linewidth]{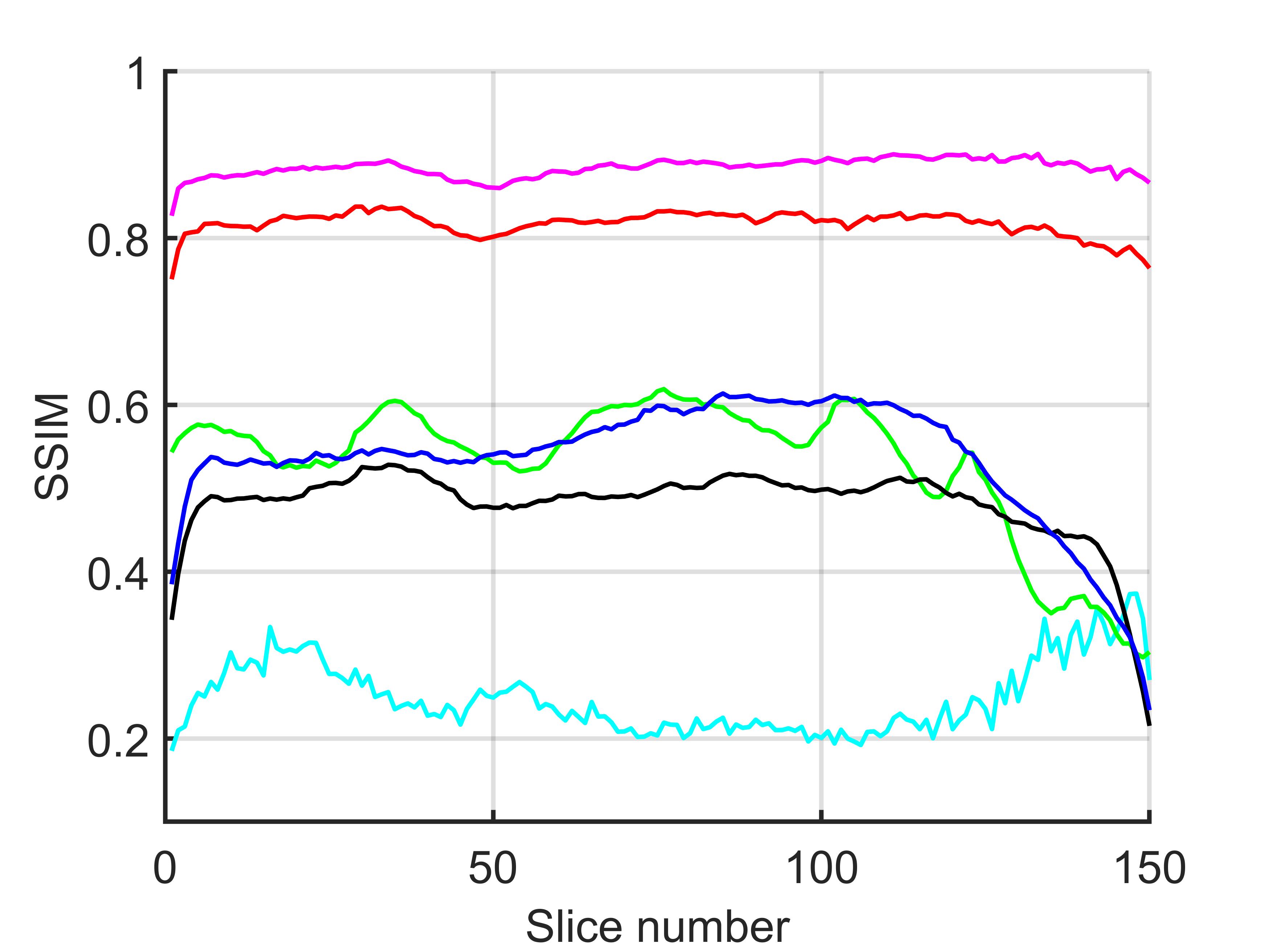}
	\hfil
	\includegraphics[width=0.3\linewidth]{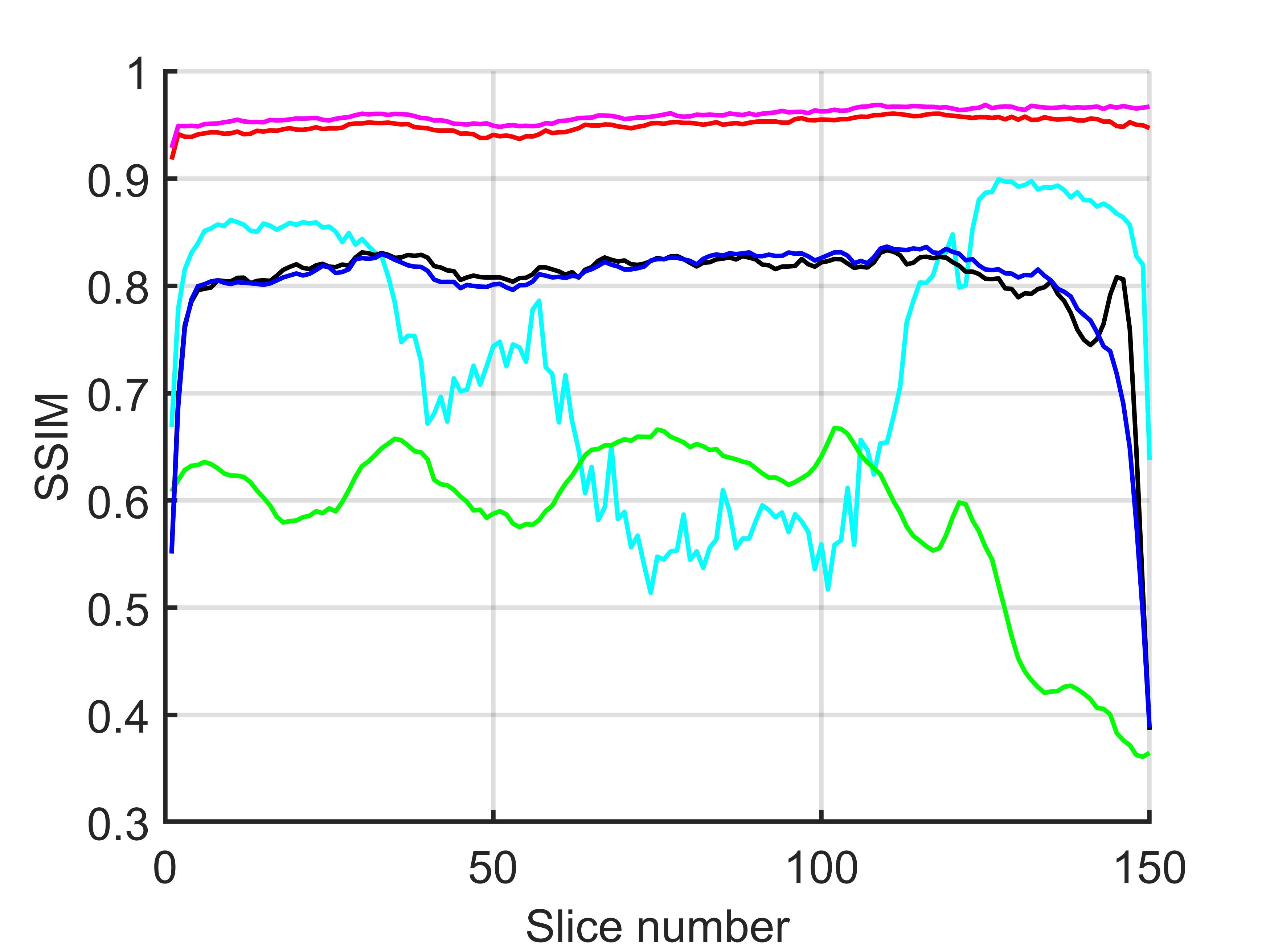}
	\hfil
	\subfloat[SR = 0.05]{\includegraphics[width=0.3\linewidth]{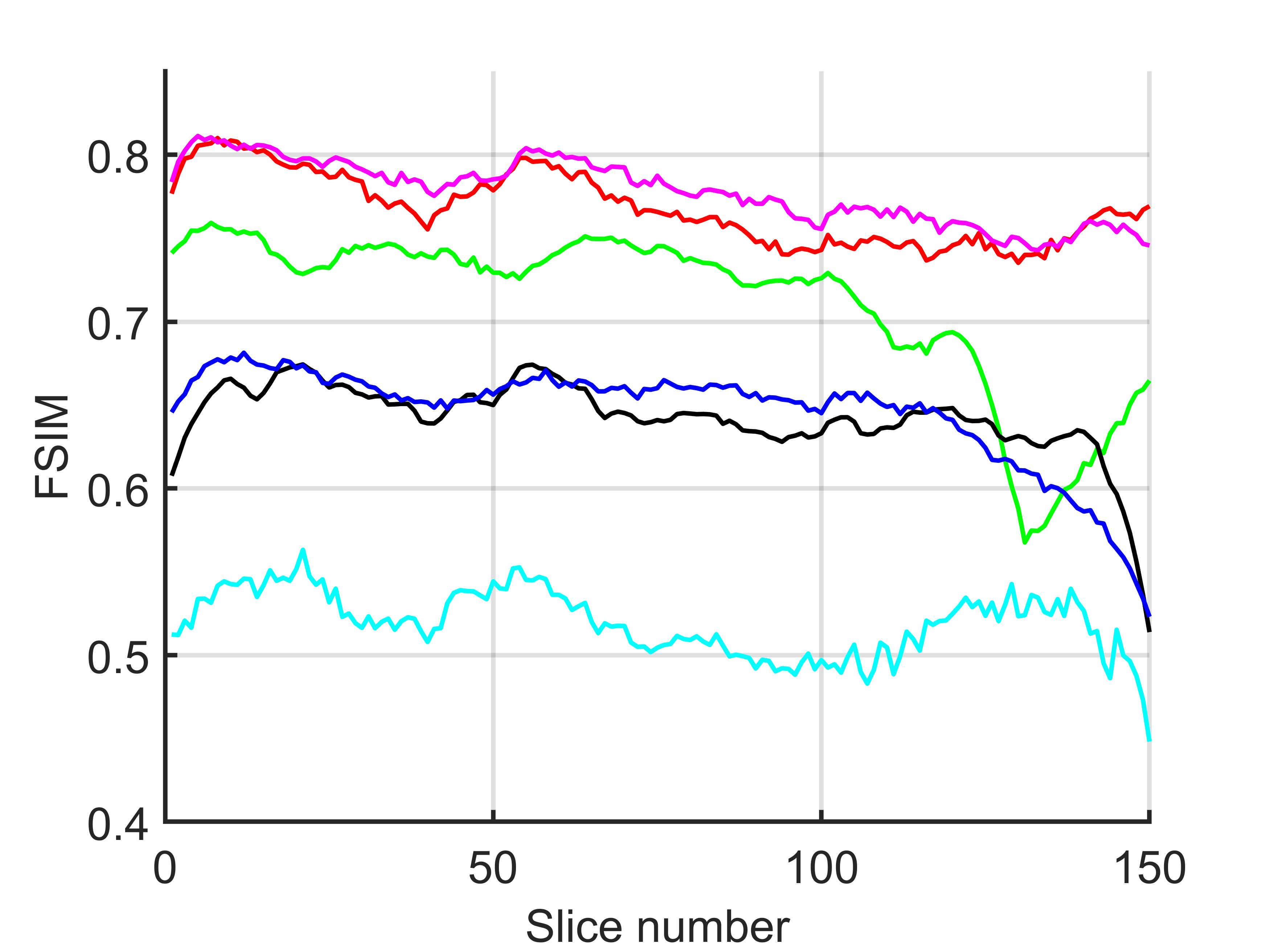}}%
	\hfil
	\subfloat[SR = 0.1]{\includegraphics[width=0.3\linewidth]{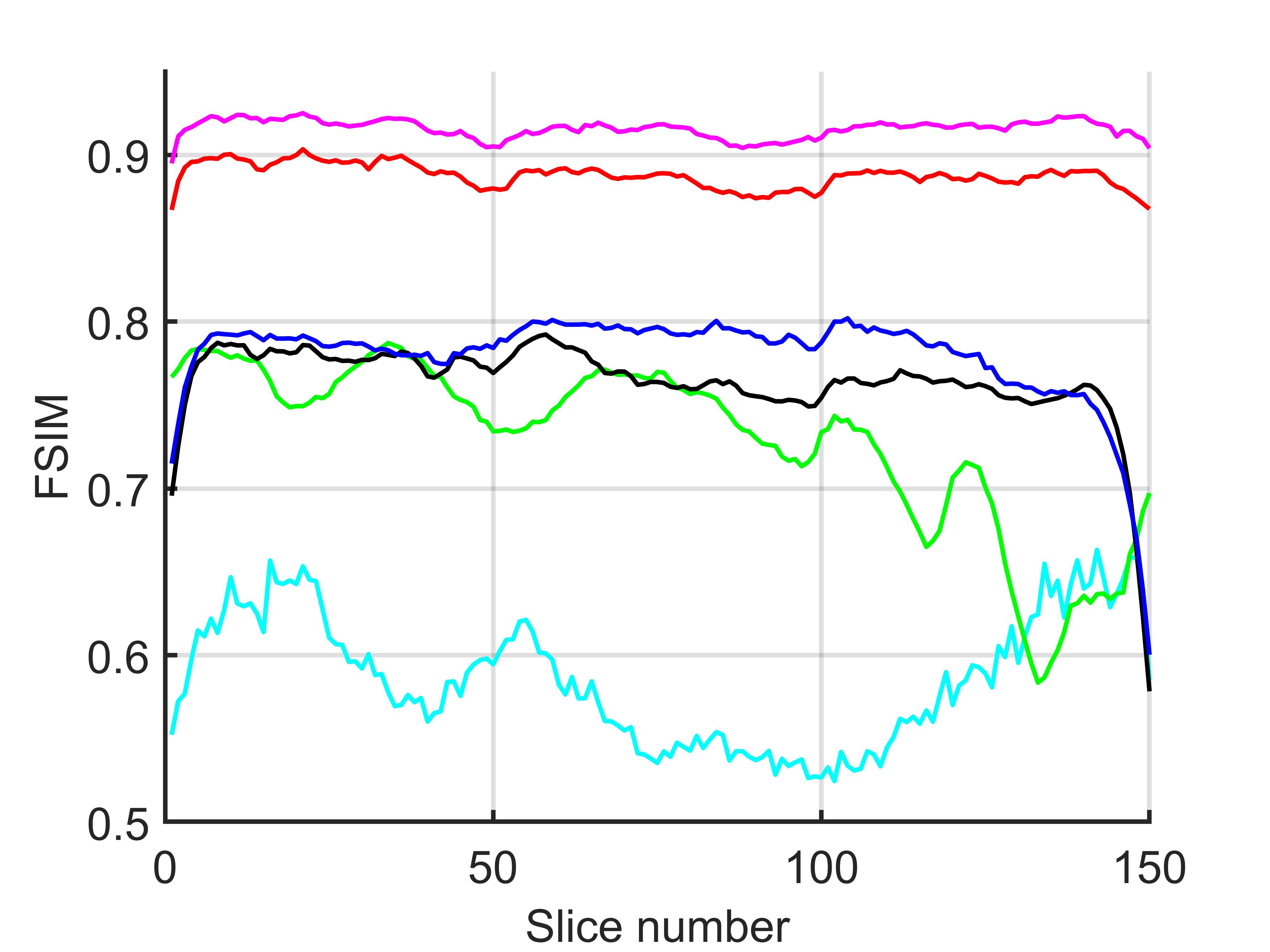}}%
	\hfil
	\subfloat[SR = 0.2]{\includegraphics[width=0.3\linewidth]{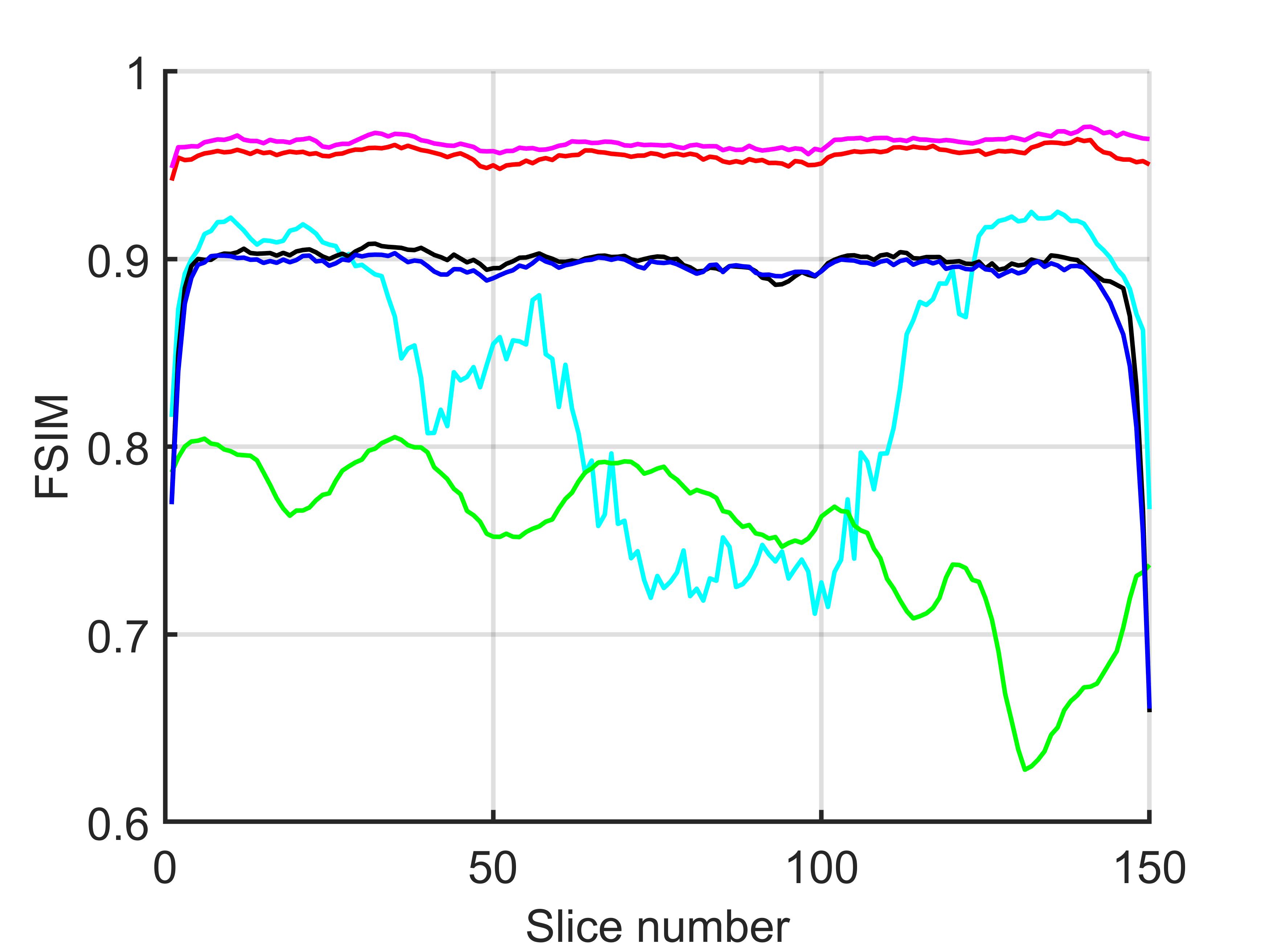}}%
	\caption{The PSNR, SSIM and FSIM of the recovered MRI by MF-TV, TMac, PSTNN, TNN, our MCTF and NC-MCTF for all slices, respectively.}
	\label{PSNR and SSIM of MRI}
\end{figure}

For quantitative evaluation, the table \ref {table_MRI} lists the PQI of all recovery results in detail, the best results are marked in bold.
It can be seen from the table that the proposed NC-MCTF obtains the best PQI, and the second best is the proposed MCTF, both of which are superior to competing methods of the same type.
Fig. \ref {PSNR and SSIM of MRI} shows the detailed PSNR, SSIM and FSIM of all slices of the restored data.
The same advantages of our model can also be seen from it.

\begin{figure*}[!t]
	\centering
	\captionsetup[subfloat]{labelsep=none,format=plain,labelformat=empty} %\captionsetup[subfloat]{labelsep=period}
	\subfloat{\includegraphics[width=0.107\linewidth]{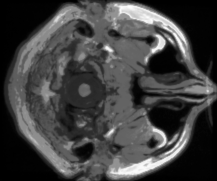}}%
	\hfil
	\subfloat{\includegraphics[width=0.107\linewidth]{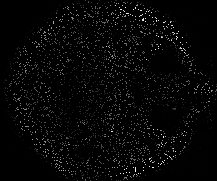}}%
	\hfil
	\subfloat{\includegraphics[width=0.107\linewidth]{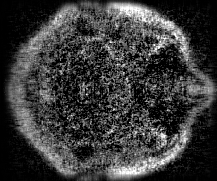}}%
	\hfil
	\subfloat{\includegraphics[width=0.107\linewidth]{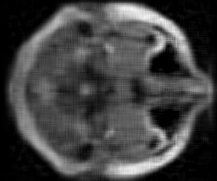}}%
	\hfil
	\subfloat{\includegraphics[width=0.107\linewidth]{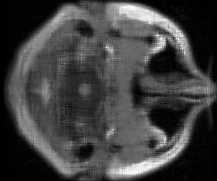}}%
	\hfil
	\subfloat{\includegraphics[width=0.107\linewidth]{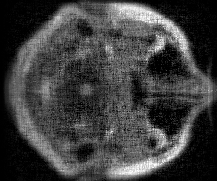}}%
	\hfil
	\subfloat{\includegraphics[width=0.107\linewidth]{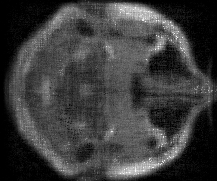}}%
	\hfil
	\subfloat{\includegraphics[width=0.107\linewidth]{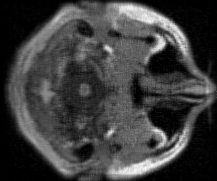}}%
	\hfil
	\subfloat{\includegraphics[width=0.107\linewidth]{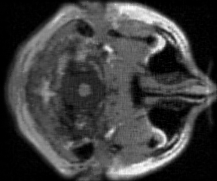}}%
	\hfil
	\subfloat[Original]{\includegraphics[width=0.107\linewidth]{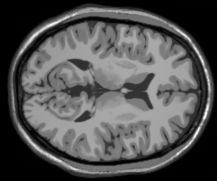}}%
	\hfil
	\subfloat[Masked]{\includegraphics[width=0.107\linewidth]{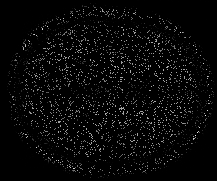}}%
	\hfil
	\subfloat[MF-TV]{\includegraphics[width=0.107\linewidth]{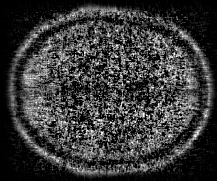}}%
	\hfil
	\subfloat[TMac]{\includegraphics[width=0.107\linewidth]{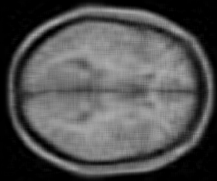}}%
	\hfil
	\subfloat[FTNN]{\includegraphics[width=0.107\linewidth]{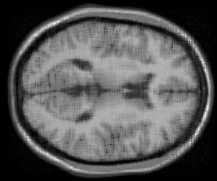}}%
	\hfil
	\subfloat[PSTNN]{\includegraphics[width=0.107\linewidth]{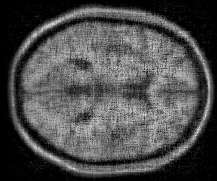}}%
	\hfil
	\subfloat[TNN]{\includegraphics[width=0.107\linewidth]{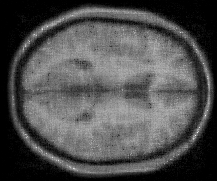}}%
	\hfil
	\subfloat[MCTF]{\includegraphics[width=0.107\linewidth]{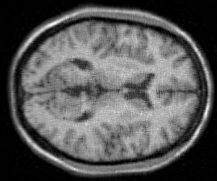}}%
	\hfil
	\subfloat[NC-MCTF]{\includegraphics[width=0.107\linewidth]{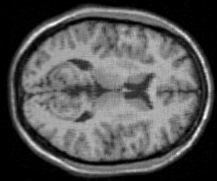}}%
	\caption{Slices of the recovered MRI by MF-TV, TMac, FTNN, PSTNN, TNN, our MCTF and NC-MCTF.  The sampling rate is 10\%.}
	\label{figure_MR_sr0.1_1}
\end{figure*}

Furthermore, the proposed model is evaluated in terms of visual evaluation.
We choose the restoration result of 0.1 sampling rate as an example.
Fig. \ref {figure_MR_sr0.1_1} shows the original MRI data, sampled data and recovery results of different methods.
Compared with competing methods, the image recovered by MCTF and NC-MCTF show richer details and clearer structures.
In addition, one can also see from the figure that the non-convex metric, as shown in NC-MCTF, produces impressive improvement over MCTF.

\begin{figure*}[!t]
	\centering
	\captionsetup[subfloat]{labelsep=none,format=plain,labelformat=empty} %\captionsetup[subfloat]{labelsep=period}
	\subfloat{\includegraphics[width=0.107\linewidth]{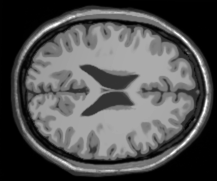}}%
	\hfil
	\subfloat{\includegraphics[width=0.107\linewidth]{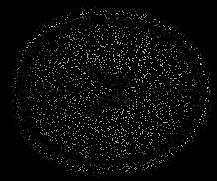}}%
	\hfil
	\subfloat{\includegraphics[width=0.107\linewidth]{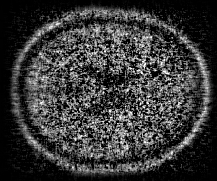}}%
	\hfil
	\subfloat{\includegraphics[width=0.107\linewidth]{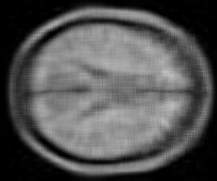}}%
	\hfil
	\subfloat{\includegraphics[width=0.107\linewidth]{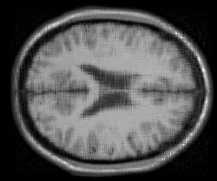}}%
	\hfil
	\subfloat{\includegraphics[width=0.107\linewidth]{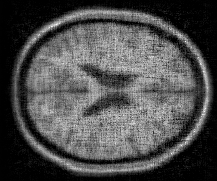}}%
	\hfil
	\subfloat{\includegraphics[width=0.107\linewidth]{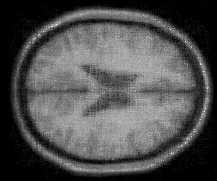}}%
	\hfil
	\subfloat{\includegraphics[width=0.107\linewidth]{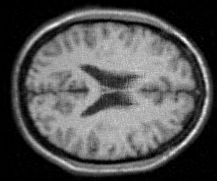}}%
	\hfil
	\subfloat{\includegraphics[width=0.107\linewidth]{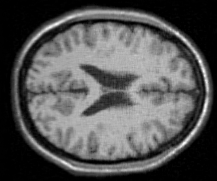}}%
	\hfil
	\subfloat[Original]{\includegraphics[width=0.107\linewidth]{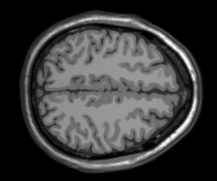}}%
	\hfil
	\subfloat[Masked image]{\includegraphics[width=0.107\linewidth]{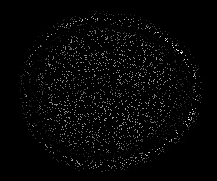}}%
	\hfil
	\subfloat[MF-TV]{\includegraphics[width=0.107\linewidth]{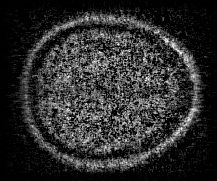}}%
	\hfil
	\subfloat[TMac]{\includegraphics[width=0.107\linewidth]{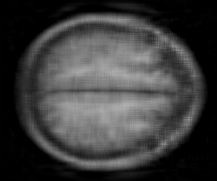}}%
	\hfil
	\subfloat[FTNN]{\includegraphics[width=0.107\linewidth]{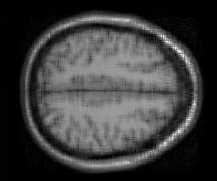}}%
	\hfil
	\subfloat[PSTNN]{\includegraphics[width=0.107\linewidth]{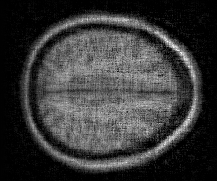}}%
	\hfil
	\subfloat[TNN]{\includegraphics[width=0.107\linewidth]{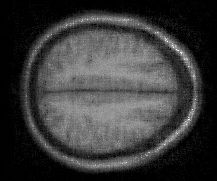}}%
	\hfil
	\subfloat[MCTF]{\includegraphics[width=0.107\linewidth]{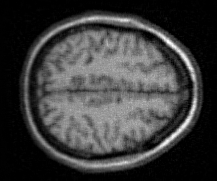}}%
	\hfil
	\subfloat[NC-MCTF]{\includegraphics[width=0.107\linewidth]{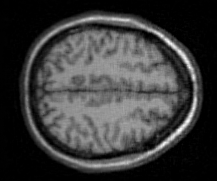}}%
	\caption{Slices of the recovered MRI by MF-TV, TMac, FTNN, PSTNN, TNN, our MCTF and NC-MCTF.  The sampling rate is 10\%.}
	\label{figure_MR_sr0.1_2}
\end{figure*}

%\begin{figure}[!t]
%	\centering
%
%	\caption{One slice of the recovered MRI by MF-TV, TMac, PSTNN, TNN, our MCTF and NC-MCTF.  The sampling rate is 10\%.}
%	\label{figure_MR_sr0.1_2}
%\end{figure}

\begin{table}
	\centering
	\caption{The averaged PSNR, SSIM, FSIM, ERGA and SAM of the recovered results on MRI by MF-TV, Tmac, FTNN, PSTNN, TNN, our MCTF and NC-MCTF with different sampling rates. The best value is highlighted in bolder fonts.}
	\label{table_MRI}
	\resizebox{85mm}{32mm}{
		\begin{tabular}{ccccccccc}
			\hline \hline
			&&&&SR =0.05 &&&	\\
			method&	noisy       &	MF-TV   &	TMac    &FTNN   &	PSTNN   &	TNN	        &MCTF   &NC-MCTF\\
			PSNR	&	10.258	&	12.332	&	20.51	&22.540 &	15.859	&	18.218	    &\underline{22.951}  &\textbf{23.698}  \\
			SSIM	&	0.228	&   0.099	&	0.45	&0.508  &	0.224	&	0.27	    &\underline{0.528}  &\textbf{0.534}	\\
			FSIM	&	0.473	&   0.52	&	0.711	&0.732  &	0.642	&	0.646	    &\underline{0.771}   &\textbf{0.775}	\\
			ERGAS	&	1030.203&	814.747	&	339.385	&268.839&	545.77	&	434.774	    &\underline{277.105} &\textbf{258.370}	\\
%			MSAM	&	76.54	&	55.603	&	31.367	&&	36.355	&	31.11	    &24.244  &\textbf{23.648}	\\
			\hline
			&&&&SR = 0.1&&&\\
			method	&	noisy	&	MF-TV	&	TMac	&FTNN   &	PSTNN	&	TNN	        &MCTF   &NC-MCTF	\\
			PSNR	&	10.492	&	15.406	&	21.411	&27.641 &	22.061	&	22.535	    &\underline{29.592} &\textbf{31.597}  	\\
			SSIM	&	0.241	&	0.25	&	0.531	&0.805  &	0.482	&	0.536	    &\underline{0.814}  &\textbf{0.884}  	\\
			FSIM	&	0.511	&	0.587	&	0.732	&0.885  &	0.764	&	0.78	    &\underline{0.883}  &\textbf{0.912} 	\\
			ERGAS	&	1002.8	&	584.827	&	308.655	&165.366&	275.473	&	266.753	    &\underline{128.252}&\textbf{101.607}  	\\
%			MSAM	&	70.986	&	41.826	&	29.345	&0.290&	24.585	&	24.6	    &17.078 &\textbf{15.182}  	\\
			\hline
			&&&&SR = 0.2&&&\\
			method	&	noisy	&	MF-TV	&	TMac	&FTNN   &	PSTNN	&	TNN		    &MCTF   &NC-MCTF\\
			PSNR	&	11.003	&	27.062	&	22.33	&31.783 &	29.152	&	28.571	    &\underline{35.550}  &\textbf{36.471}  	\\
			SSIM	&	0.271	&	0.737	&	0.586	&0.907  &	0.804	&	0.802	    &\underline{0.950}   &\textbf{0.960 } 	\\
			FSIM	&	0.564	&	0.84	&	0.754	&0.938  &	0.895	&	0.891	    &\underline{0.953}   &\textbf{0.960}  	\\
			ERGAS	&	945.583	&	173.636	&	276.269	&100.444&	127.133	&	136.182	    &\underline{64.008}  &\textbf{57.003}  	\\
%			MSAM	&	62.887	&	21.792	&	27.267	&0.229&	17.513	&	17.855	    &12.514  &\textbf{11.319}  	\\
			\hline
			&&&&SR = 0.3&&&\\
			method	&	noisy	&	MF-TV	&	TMac	&FTNN  &	PSTNN	&	TNN	        &MCTF   &NC-MCTF	\\
			PSNR	&	11.582	&	36.355	&	23.077	&34.806&	32.608	&	32.481	    &\underline{37.783}  &\textbf{38.531}  	\\
			SSIM	&	0.303	&	0.954	&	0.625	&0.949 &	0.895	&	0.89	    &\underline{0.969}   &\textbf{0.974}  	\\
			FSIM	&	0.597	&	0.962	&	0.773	&0.963 &	0.939	&	0.939	    &\underline{0.969}   &\textbf{0.974}  	\\
			ERGAS	&	884.608	&	52.449	&	252.057	&70.542&	85.845	&	87.312	    &\underline{49.323}  &\textbf{45.028}  	\\
%			MSAM	&	56.216	&	13.849	&	25.722	&0.191&	14.858	&	14.879	    &11.025  &\textbf{10.211}  	\\
			\hline \hline
	\end{tabular}}
\end{table}

\subsection{Video}

%***This example is also about a completely artificial problem****

In this subsection, two public and classic video data-sets, i.e., "Suzie" and "Hall"\footnote{http://trace.eas.asu.edu/yuv/} 
with size 144 $\times$ 176 $\times$ 150 are selected for comparative experiments to test the performance of our model.
SRs are set as 5\%, 10\% and 20\%. Further, the proposed model was comprehensively evaluated from both quantitative and qualitative perspectives.
Here, we set the rank to $(T_1, T_2, T_3)$, where $T_1, T_2, T_3$ denote the number of the largest 0.5\% singular values of model-1, model-2 and model-3, respectively.

Quantitative comparison: Table \ref {table_video_suzie} and table \ref {table_video_hall} give the detailed PQI of all recovered data under three sampling rates. The boldface indicates the best PQI for each sampling rate.
It can be clearly seen from the table \ref {table_video_suzie} and table \ref {table_video_hall} that among all the test methods, the proposed NC-MCTF obtains the best results, and the evaluation index obtained by the proposed MCTF is also superior to competitive methods.

\begin{figure*}[!t]
	\centering
	\captionsetup[subfloat]{labelsep=none,format=plain,labelformat=empty} %\captionsetup[subfloat]{labelsep=period}
	\subfloat[Original]{\includegraphics[width=0.107\linewidth]{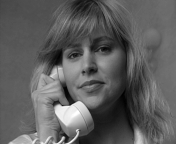}}%
	\hfil
	\subfloat[Masked]{\includegraphics[width=0.107\linewidth]{myfigure/Image_result/Image_video_b94_sr0.05/Nosiy}}%
	\hfil
	\subfloat[MF-TV]{\includegraphics[width=0.107\linewidth]{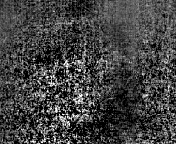}}%
	\hfil
	\subfloat[TMac]{\includegraphics[width=0.107\linewidth]{myfigure/Image_result/Image_video_b94_sr0.05/Tmac}}%
	\hfil
	\subfloat[FTNN]{\includegraphics[width=0.107\linewidth]{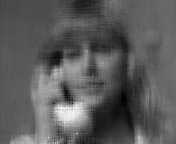}}%
	\hfil
	\subfloat[PSTNN]{\includegraphics[width=0.107\linewidth]{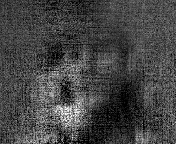}}%
	\hfil
	\subfloat[TNN]{\includegraphics[width=0.107\linewidth]{myfigure/Image_result/Image_video_b94_sr0.05/TNN}}%
	\hfil
	\subfloat[MCTF]{\includegraphics[width=0.107\linewidth]{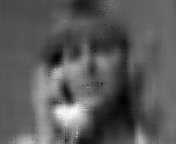}}%
	\hfil
	\subfloat[NC-MCTF]{\includegraphics[width=0.107\linewidth]{myfigure/Band_image_NonMCTF/suzie_sr005_b94}}%
	\caption{One slice of the recovered video for “suzie” by MF-TV, TMac, FTNN, PSTNN, TNN, our MCTF and NC-MCTF.  The sampling rate is 5\%.}
	\label{figure_video_sr0.05}
	\nonumber
\end{figure*}

\begin{figure*}[!t]
	\centering
	\captionsetup[subfloat]{labelsep=none,format=plain,labelformat=empty} %\captionsetup[subfloat]{labelsep=period}
	\subfloat{\includegraphics[width=0.107\linewidth]{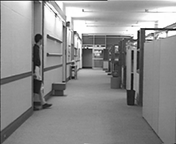}}%
	\hfil
	\subfloat{\includegraphics[width=0.107\linewidth]{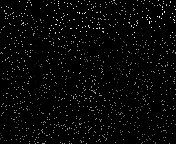}}%
	\hfil
	\subfloat{\includegraphics[width=0.107\linewidth]{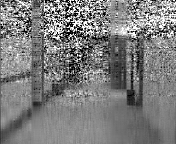}}%
	\hfil
	\subfloat{\includegraphics[width=0.107\linewidth]{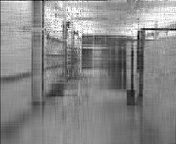}}%
	\hfil
	\subfloat{\includegraphics[width=0.107\linewidth]{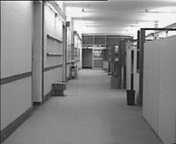}}%
	\hfil
	\subfloat{\includegraphics[width=0.107\linewidth]{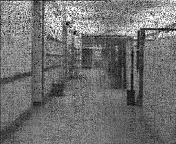}}%
	\hfil
	\subfloat{\includegraphics[width=0.107\linewidth]{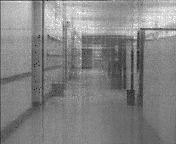}}%
	\hfil
	\subfloat{\includegraphics[width=0.107\linewidth]{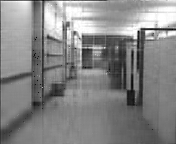}}%
	\hfil
	\subfloat{\includegraphics[width=0.107\linewidth]{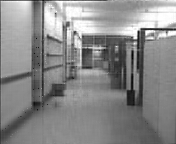}}%
	\hfil
	\setcounter{subfigure}{0}
	\subfloat[Original]{\includegraphics[width=0.107\linewidth]{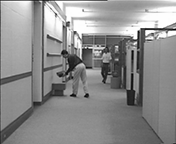}}%
	\hfil
	\subfloat[Masked]{\includegraphics[width=0.107\linewidth]{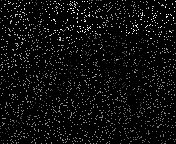}}%
	\hfil
	\subfloat[MF-TV]{\includegraphics[width=0.107\linewidth]{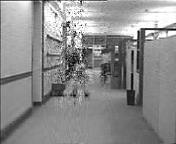}}%
	\hfil
	\subfloat[TMac]{\includegraphics[width=0.107\linewidth]{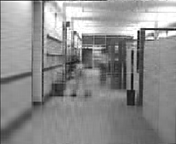}}%
	\hfil
	\subfloat[FTNN]{\includegraphics[width=0.107\linewidth]{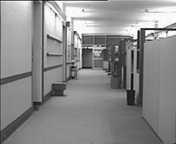}}%
	\hfil
	\subfloat[PSTNN]{\includegraphics[width=0.107\linewidth]{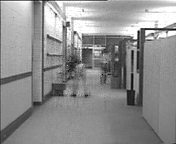}}%
	\hfil
	\subfloat[TNN]{\includegraphics[width=0.107\linewidth]{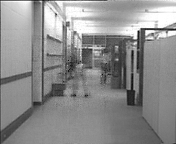}}%
	\hfil
	\subfloat[MCTF]{\includegraphics[width=0.107\linewidth]{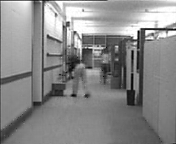}}%
	\hfil
	\subfloat[NC-MCTF]{\includegraphics[width=0.107\linewidth]{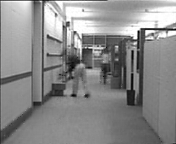}}%
	\caption{One slice of the recovered video for “hall” by MF-TV, TMac, FTNN, PSTNN, TNN, our MCTF and NC-MCTF.  The sampling rates are 5\% and 10\%, respectively.}
	\label{figure_hall_sr0.05}
\end{figure*}

Based on above quantitative comparison, we further conduct a quantitative evaluation of the proposed model in terms of vision. Fig. \ref {figure_video_sr0.05} and Fig. \ref {figure_hall_sr0.05} show the partial slice images of restored data under different sampling conditions.
The closer the restored result is to the original reference image, the better the performance of the corresponding model.
It can be seen from the figure that the proposed model has achieved a significant advantage in restored images, especially at low sampling rates.
Because when the sampling rate is low, the original image information contained in the input observation image is scarce.
If it is desired to restore the data image relatively accurately, as at high sampling rates, it is necessary to impose additional prior constraints on the solution space of the optimization model to increase the accuracy of the obtained solution, as the proposed model does.

\begin{table} [!t]
\centering
	\caption{The averaged PSNR, SSIM, FSIM, ERGA and SAM of the recovered results on video "suzie" by Tmac, MF-TV, TNN, FTNN, PSTNN and our MCTF and NC-MCTF with different sampling rates. The best value is highlighted in bolder fonts.}
	\label{table_video_suzie}
    \resizebox{85mm}{25mm}{
	\begin{tabular}{ccccccccc}
		\hline \hline
		&& & &SR =0.05&&&	\\
		method  &	noisy   &	MF-TV   &	TMac    &FTNN     &	PSTNN   &	TNN     &MCTF   &NC-MCTF          \\
		PSNR	&	7.259	&	13.801	&	23.385	&27.294	  &	17.447	&	22.005  &\underline{27.430}  &\textbf{29.312}	\\
		SSIM	&	0.009	&	0.094	&	0.622	&0.465	  &	0.192	&	0.563   &\underline{0.766}   &\textbf{0.822}	\\
		FSIM	&	0.454	&	0.42	&	0.792	&0.555	  &	0.59	&	0.776   &\underline{0.842}  &\textbf{0.880}	\\
		ERGAS	&	1057.282&	501.117	&	167.927	&129.27	      &	327.678	&	194.844 &\underline{104.955} &\textbf{84.698}	\\
%		MSAM	&	77.324	&	24.095	&	6.927	&	      &	13.775	&	7.797   & 4.560  &\textbf{3.579}	\\
		\hline
		&&&&SR = 0.1&&&\\
		method&	noisy   &	MF-TV   &	TMac   &FTNN  &PSTNN	&TNN     &MCTF   &NC-MCTF\\
		PSNR&   7.493   &	22.356  &	26.189 &\underline{29.484} &26.647 	&26.032  &29.414 &\textbf{30.223}  \\
		SSIM&	0.014   &	0.605   &	0.74   &0.585  &0.68	&0.692   &\underline{0.801}  &\textbf{0.830}  \\
		FSIM&	0.426   &	0.758   &	0.838  &0.670  &0.843	&0.846   &\underline{0.886}  &\textbf{0.897}  \\
		ERGAS&	1029.096&	196.059 &	124.369&95.472	   &117.104	&124.923 &\underline{84.888} &\textbf{77.398}  \\
%		MSAM&	71.725  &	6.99    &	5.423   &	   &5.171	&5.405	 &3.641  &\textbf{3.298}\\
		\hline
		&&&&SR = 0.2&&&\\
		method&	noisy  &	MF-TV &	TMac   &FTNN	&PSTNN 		&	TNN     &MCTF   &NC-MCTF\\
		PSNR&	8.005  &	32.064&	27.274 &32.184	&30.566	&	30.561  &\underline{33.353}  &\textbf{33.992} \\
		SSIM&	0.02   &	0.872 &	0.782  &0.721	&0.829	&	0.831   &\underline{0.906}   &\textbf{0.917} \\
		FSIM&	0.391  &	0.916 &	0.853  &0.788	&0.91	&	0.911   &\underline{0.938}   &\textbf{0.945} \\
		ERGAS&	970.285&	66.692&	109.627&65.322		&75.472		    &	75.598  &\underline{53.121}  &\textbf{49.395} \\
%		MSAM&	63.522 &	2.81  &	4.812  &	 	&	3.399	    &	3.395   &2.390   &\textbf{2.229} \\
		\hline \hline
	\end{tabular}}
\end{table}

\begin{table} [!t]
	\centering
	\caption{The averaged PSNR, SSIM, FSIM, ERGA and SAM of the recovered results on video "hall" by MF-TV, TMac, PSTNN, TNN, FTNN, our MCTF and NC-MCTF with different sampling rates. The best value is highlighted in bolder fonts.}
	\label{table_video_hall}
    \resizebox{85mm}{25mm}{
	\begin{tabular}{ccccccccc}
		\hline \hline
		&&&&SR =0.05 &&&	\\
		method  &	noisy   &   MF-TV   &	TMac    &FTNN&	PSTNN   &	TNN     &MCTF   &NC-MCTF\\
		PSNR	&	4.82	&	13.539	&	22.101	&\textbf{30.022}&	16.075	&	20.78	&26.215 &\underline{27.415}       \\
		SSIM	&	0.007	&	0.412	&	0.675	&0.792&	0.36	&	0.636	&\underline{0.856}  &\textbf{0.882}       	\\
		FSIM	&	0.387	&   0.612	&	0.789	&0.835&	0.672	&	0.792	&\underline{0.890}  &\textbf{0.906}       	\\
		ERGAS	&	1225.779&	452.351	&	168.866	&98.14&	335.52	&	195.315	&\underline{105.199}&\textbf{91.728}       	\\
%		MSAM	&	77.299	&	12.865	&	3.818	&&	8.64	&	4.299	&2.465  &\textbf{2.263}       \\
		\hline
		&&&&SR = 0.1&&&\\
		method	&	noisy	&MF-TV	&	TMac	&FTNN	&	PSTNN	&	TNN	        &MCTF   &NC-MCTF\\
		PSNR	&	5.055	&24.855	&	26.936	&\textbf{32.790}	&	29.014	&	28.433	    &30.731 &\underline{31.481}\\
		SSIM	&	0.013	&0.829	&	0.854	&0.854  &	0.892	&	0.905   	&\underline{0.933} &\textbf{0.942}\\
		FSIM	&	0.393	&0.873	&	0.888	&0.889  &	0.934	&	0.936	    &\underline{0.945} &\textbf{0.952}\\
		ERGAS	&	1193.075&131.422&	97.185	&\underline{59.375}	&	77.395	&	82.259	    &62.923 &\textbf{57.805}\\
%		MSAM	&	71.7	&3.669	&	2.404	&       &	2.417	&	2.46	    &1.821 &\textbf{1.730}\\
		\hline
		&&&&SR = 0.2&&&\\
		method	&	noisy	&MF-TV		&	TMac	&FTNN&	PSTNN	&	TNN	        &MCTF   &NC-MCTF	\\
		PSNR	&	5.567	&	33.006	&	27.648	&\textbf{35.755}&	33.629	&	33.691	&33.052 &\underline{34.097}	\\
		SSIM	&	0.025	&	0.94	&	0.869	&0.902&	0.961	&	0.962	&\underline{0.956}  &\textbf{0.962}	\\
		FSIM	&	0.403	&	0.954	&	0.897	&0.927&	\underline{0.973}	&	\textbf{0.974}	&0.965  &0.970	\\
		ERGAS	&	1124.737&	50.971	&	89.271	&\underline{44.184}&	46.123	&	45.851	&48.414 &\textbf{43.191}	\\
%		MSAM	&	63.507	&	1.779	&	2.226	&&	1.584	&	1.565	&1.573  &\textbf{0.454}	\\
		\hline \hline
	\end{tabular}}
\end{table}

\begin{figure*}[!t]
	\centering
	\captionsetup[subfloat]{labelsep=none,format=plain,labelformat=empty} %\captionsetup[subfloat]{labelsep=period}
	\subfloat{\includegraphics[width=0.107\linewidth]{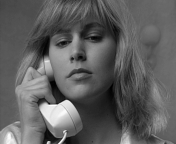}}%
	\hfil
	\subfloat{\includegraphics[width=0.107\linewidth]{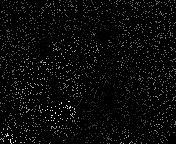}}%
	\hfil
 	\subfloat{\includegraphics[width=0.107\linewidth]{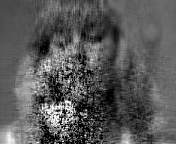}}%
	\hfil
	\subfloat{\includegraphics[width=0.107\linewidth]{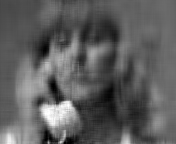}}%
	\hfil
	\subfloat{\includegraphics[width=0.107\linewidth]{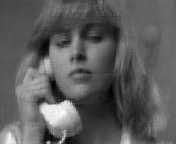}}%
	\hfil
	\subfloat{\includegraphics[width=0.107\linewidth]{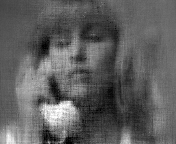}}%
	\hfil
	\subfloat{\includegraphics[width=0.107\linewidth]{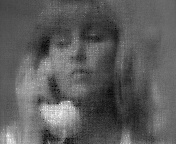}}%
	\hfil
	\subfloat{\includegraphics[width=0.107\linewidth]{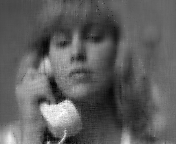}}%
	\hfil
	\subfloat{\includegraphics[width=0.107\linewidth]{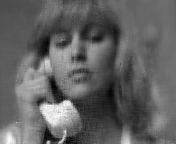}}%
	\hfil
	\setcounter{subfigure}{0}
	\subfloat[Original]{\includegraphics[width=0.107\linewidth]{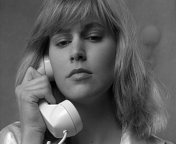}}
	\hfil
	\subfloat[Masked]{\includegraphics[width=0.107\linewidth]{myfigure/Image_result/Image_video_b10_sr0.2/Nosiy}}%
	\hfil
	\subfloat[MF-TV]{\includegraphics[width=0.107\linewidth]{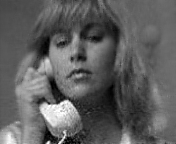}}%
	\hfil
	\subfloat[TMac]{\includegraphics[width=0.107\linewidth]{myfigure/Image_result/Image_video_b10_sr0.2/Tmac}}%
	\hfil
	\subfloat[FTNN]{\includegraphics[width=0.107\linewidth]{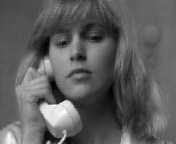}}%
	\hfil
	\subfloat[PSTNN]{\includegraphics[width=0.107\linewidth]{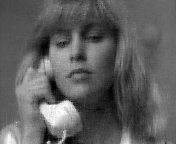}}%
	\hfil
	\subfloat[TNN]{\includegraphics[width=0.107\linewidth]{myfigure/Image_result/Image_video_b10_sr0.2/TNN}}%
	\hfil
	\subfloat[MCTF]{\includegraphics[width=0.107\linewidth]{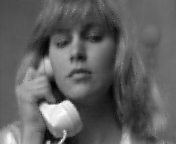}}%
	\hfil
	\subfloat[NC-MCTF]{\includegraphics[width=0.107\linewidth]{myfigure/Band_image_NonMCTF/suzie_sr02_b10}}%
	\caption{One slice of the recovered video for “suzie” by MF-TV, TMac, FTNN, PSTNN, TNN, our MCTF and NC-MCTF.  The sampling rates are 10\%, 20\%, respectively.}
	\label{figure_video_sr0.1and0.2}
		\nonumber
\end{figure*}

%\begin{figure}[!t]
%	\centering
%
%	\caption{One slice of the recovered video for “suzie” by MF-TV, TMac, PSTNN, TNN, our MCTF and NC-MCTF. The sampling rate is 10\%.}
%	\label{figure_video_sr0.1}
%\end{figure}
%
%
%
%
%\begin{figure}[!t]
%	\centering
%
%	\caption{One slice of the recovered video for “suzie” by MF-TV, TMac, PSTNN, TNN, our MCTF and NC-MCTF.  The sampling rate is 20\%.}
%	\label{figure_video_sr0.2}
%\end{figure}

%\begin{figure}[!t]
%	\centering
%
%	\caption{One slice of the recovered video “hall” by MF-TV, TMac, PSTNN, TNN, our MCTF and NC-MCTF.  The sampling rate is 10\%.}
%	\label{figure_hall_sr0.1}
%\end{figure}

\begin{figure}[!t]
	\centering
	\includegraphics[width=0.7\linewidth]{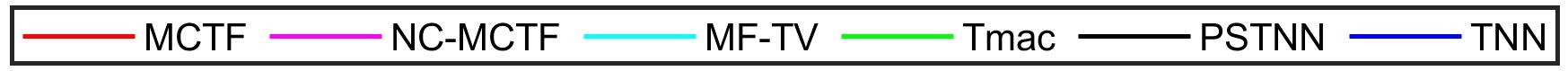}
	\hfil
	\includegraphics[width=0.3\linewidth]{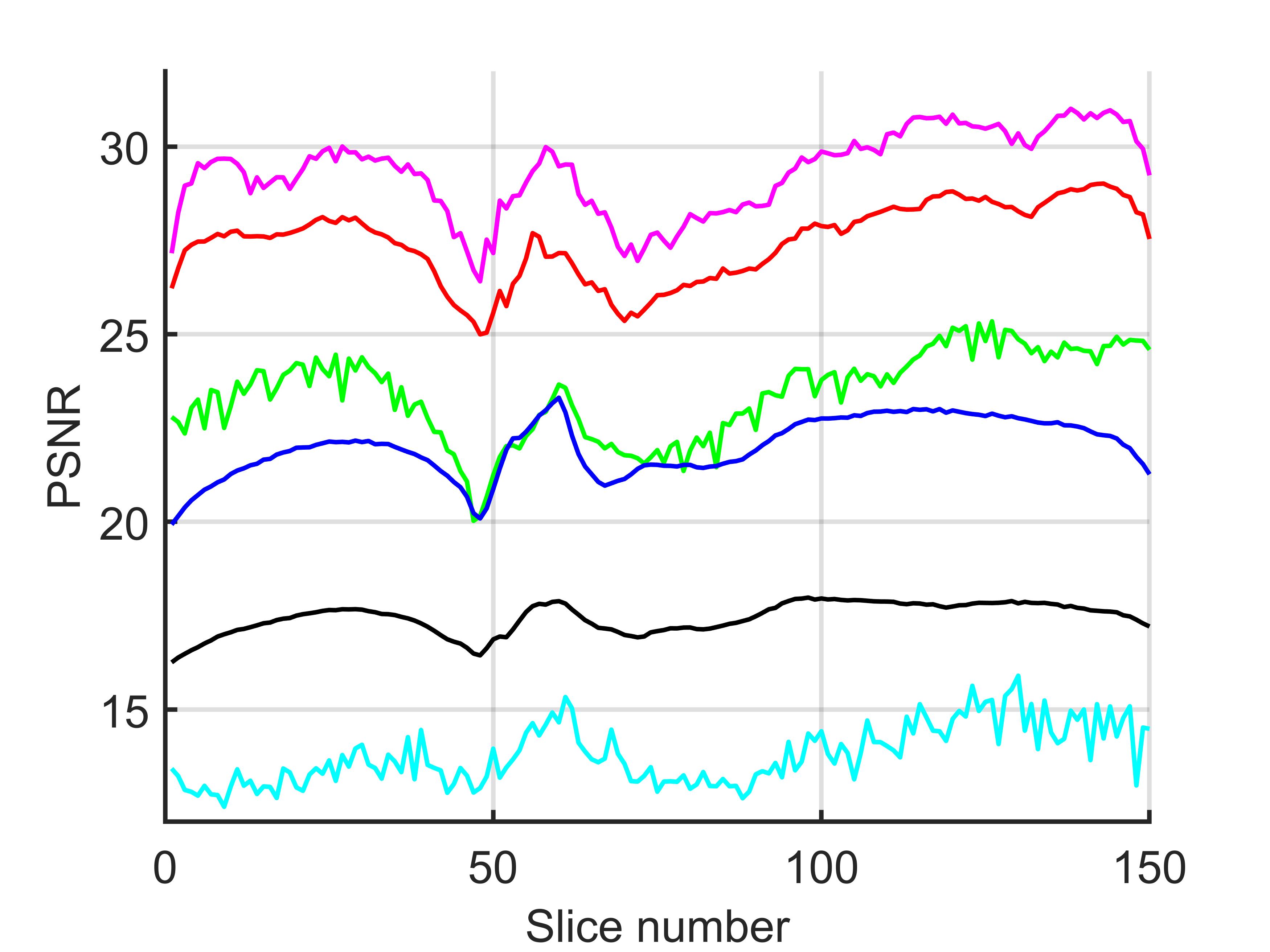}
	\hfil
	\includegraphics[width=0.3\linewidth]{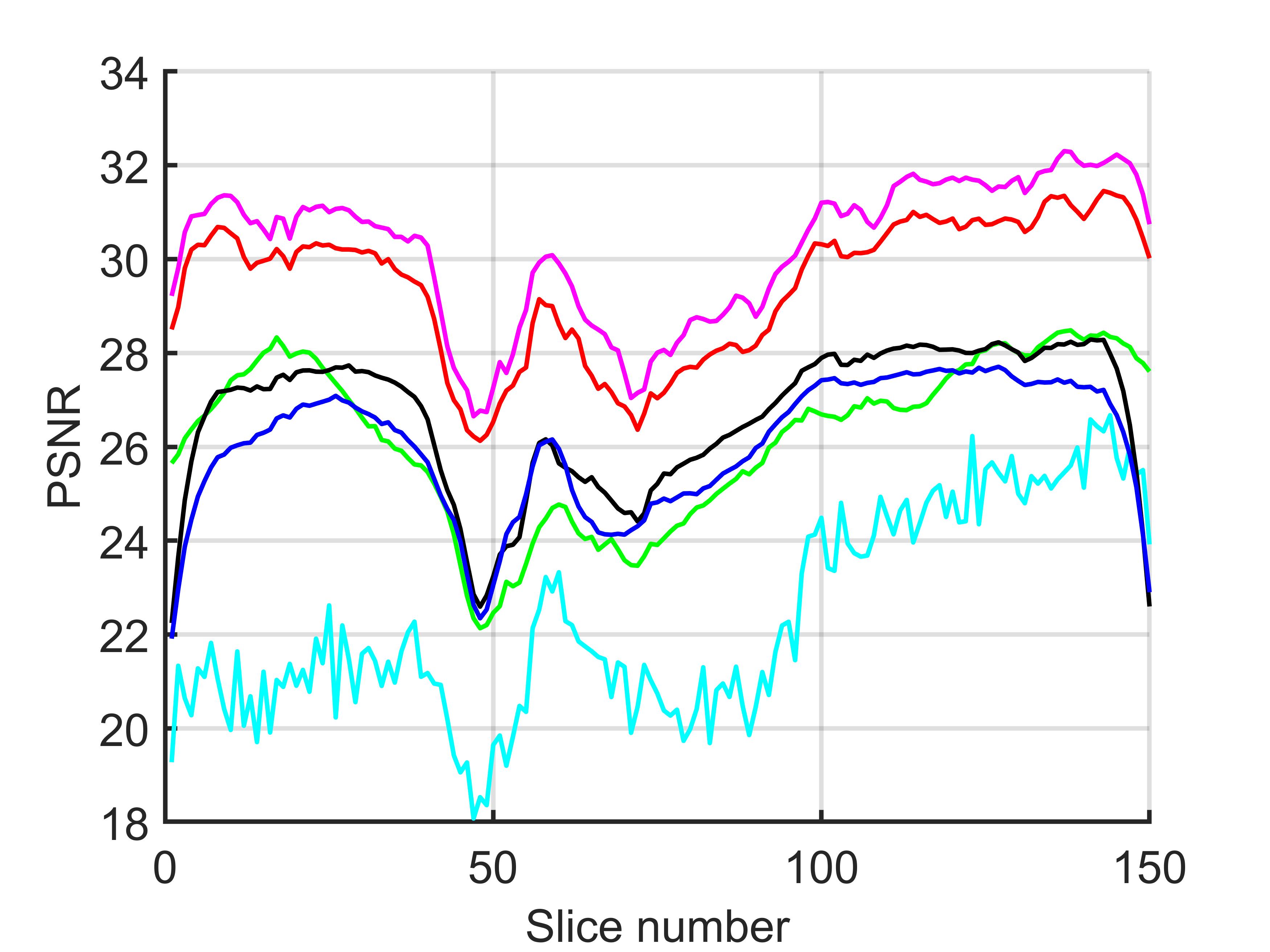}
	\hfil
	\includegraphics[width=0.3\linewidth]{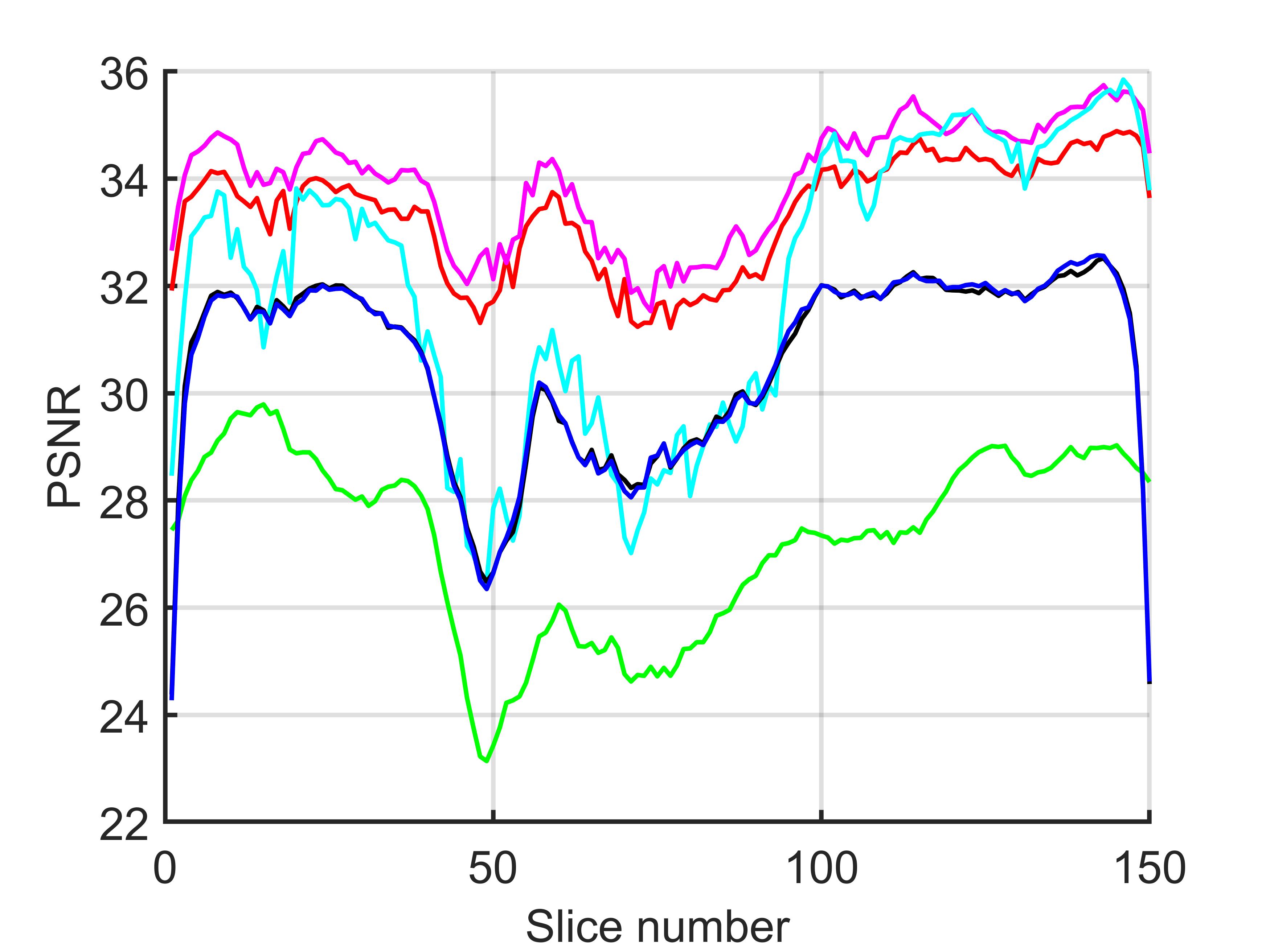}
	\hfil
	\includegraphics[width=0.3\linewidth]{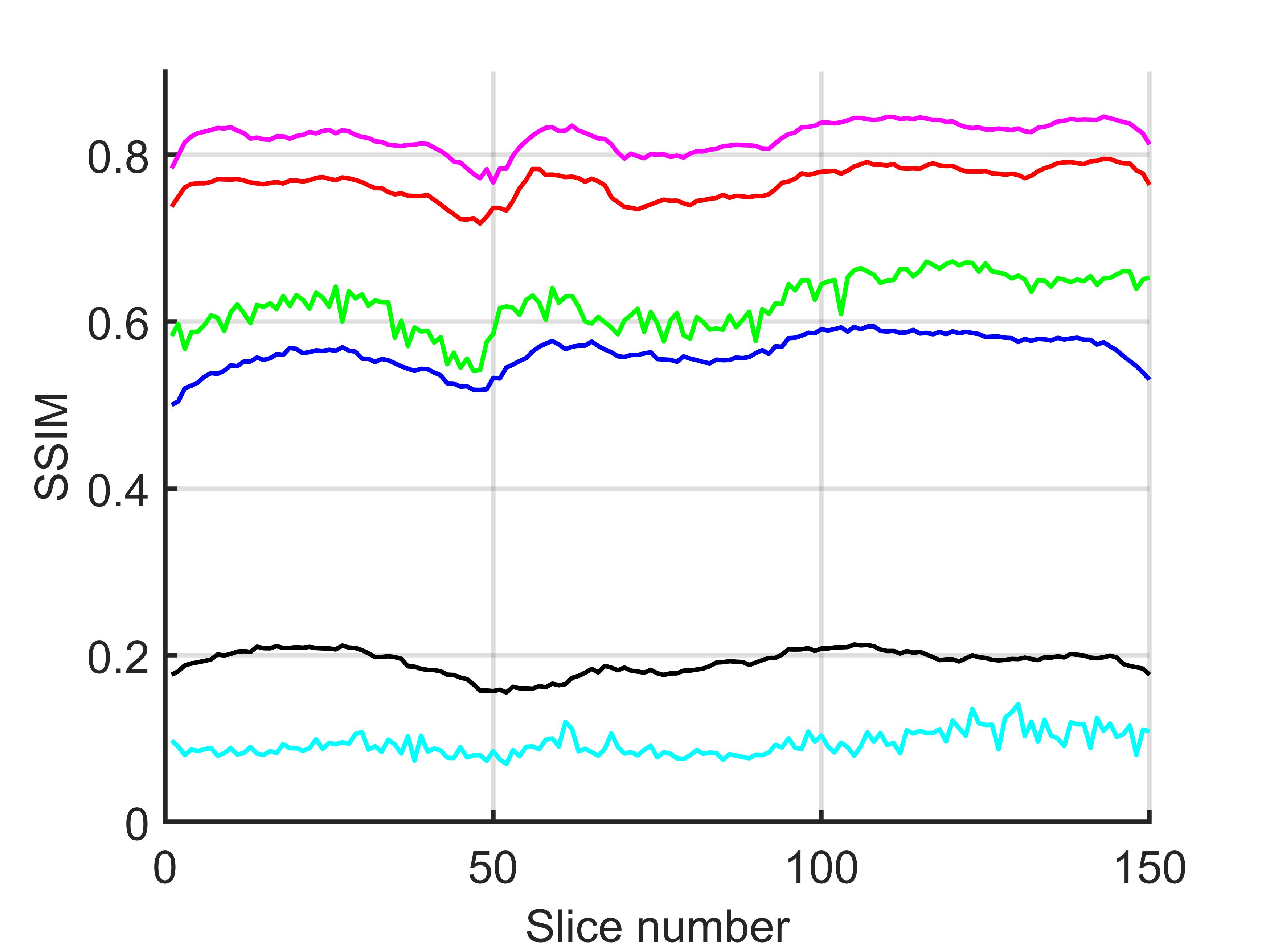}
	\hfil
	\includegraphics[width=0.3\linewidth]{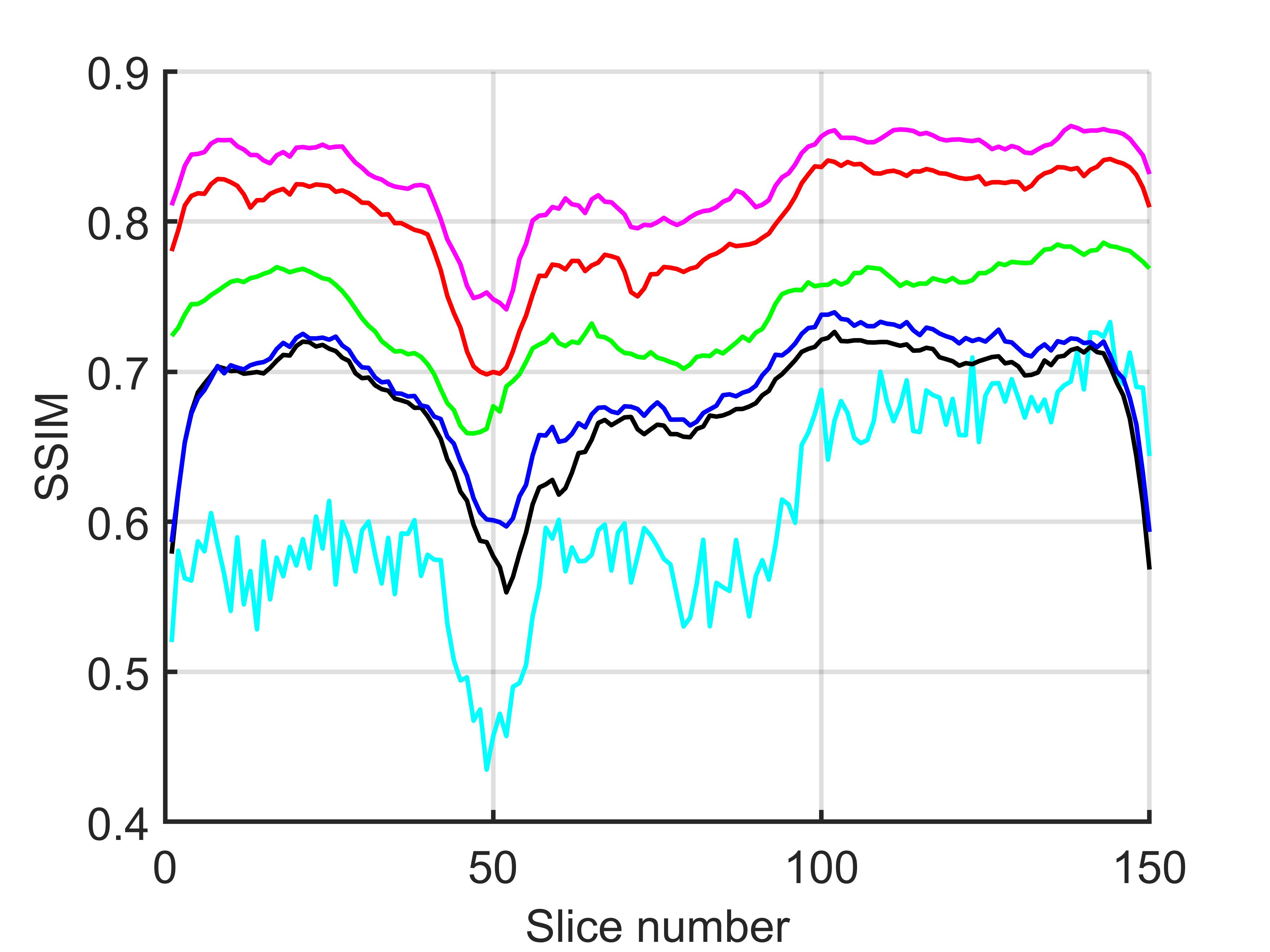}
	\hfil
	\includegraphics[width=0.3\linewidth]{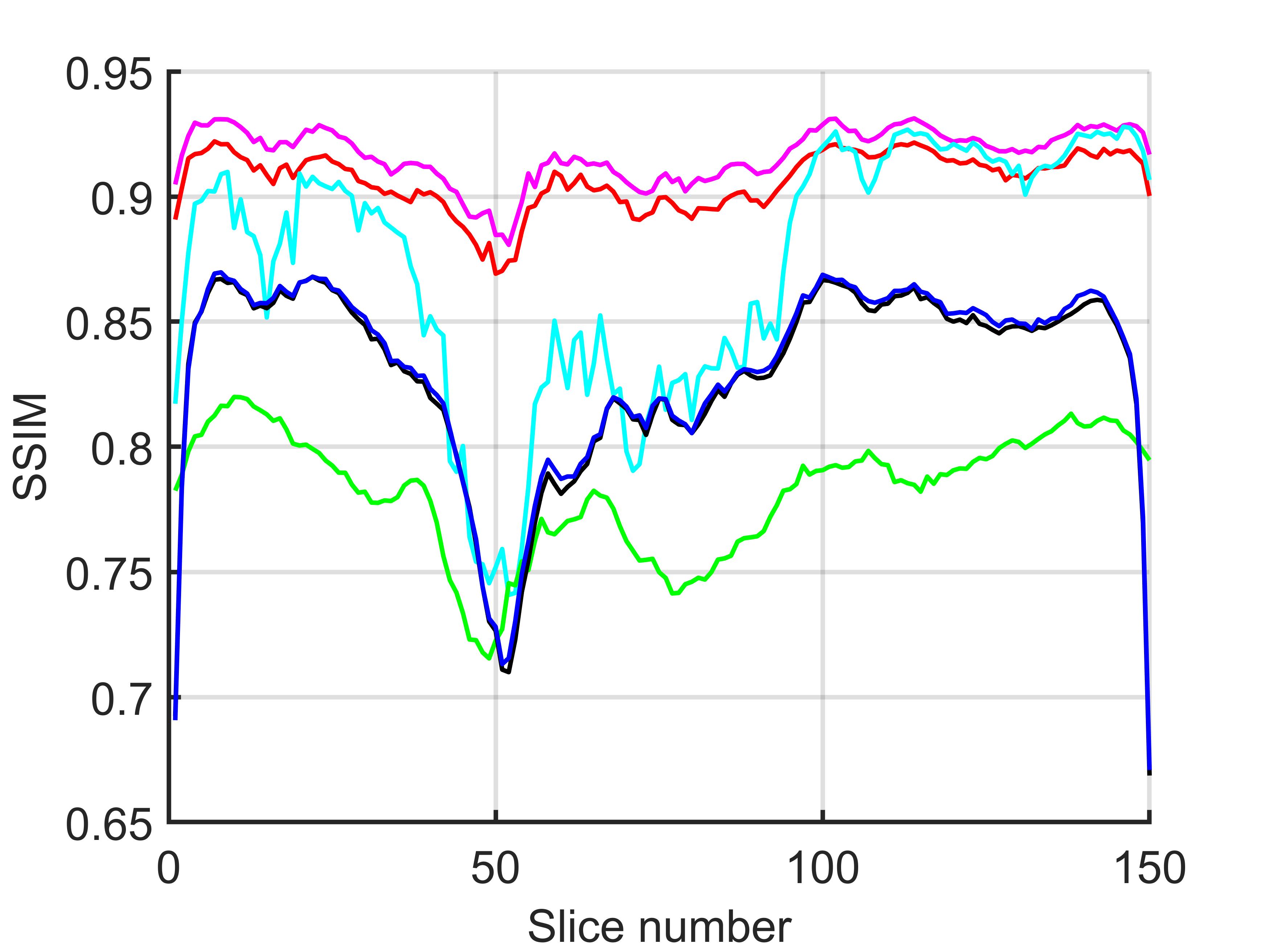}
	\hfil
	\subfloat[SR = 0.05]{\includegraphics[width=0.3\linewidth]{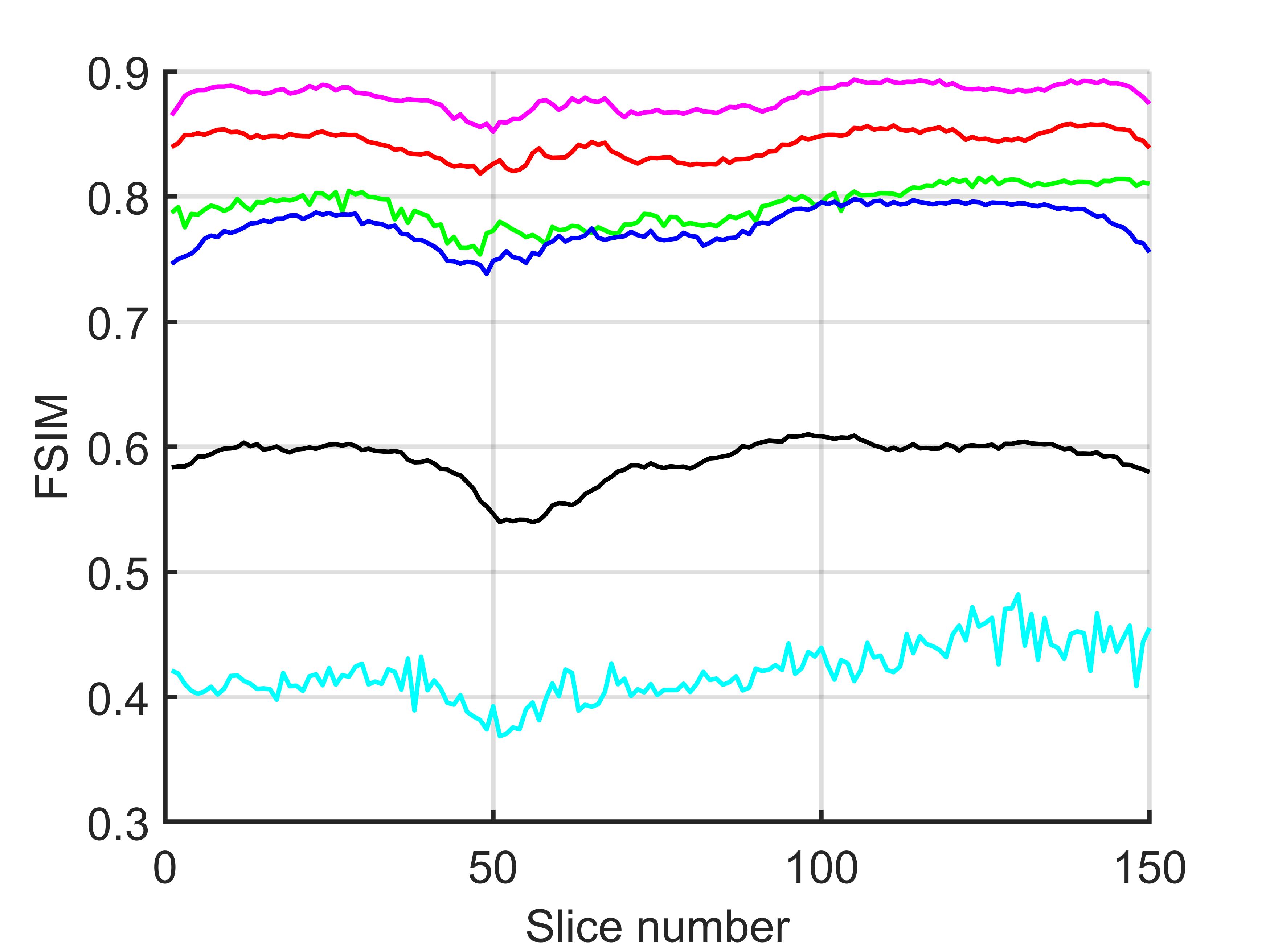}}%
	\hfil
	\subfloat[SR = 0.1]{\includegraphics[width=0.3\linewidth]{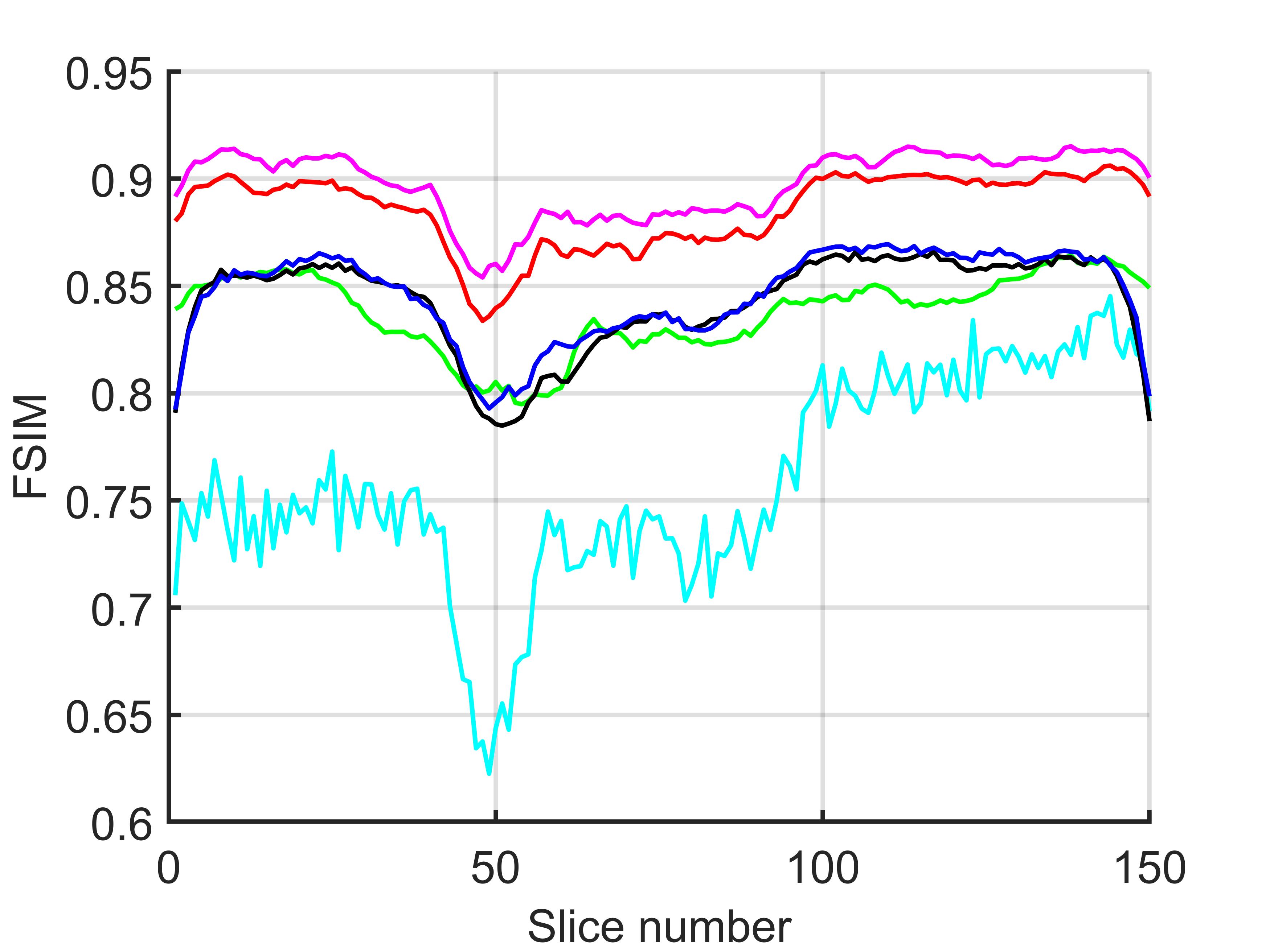}}%
	\hfil
	\subfloat[SR = 0.2]{\includegraphics[width=0.3\linewidth]{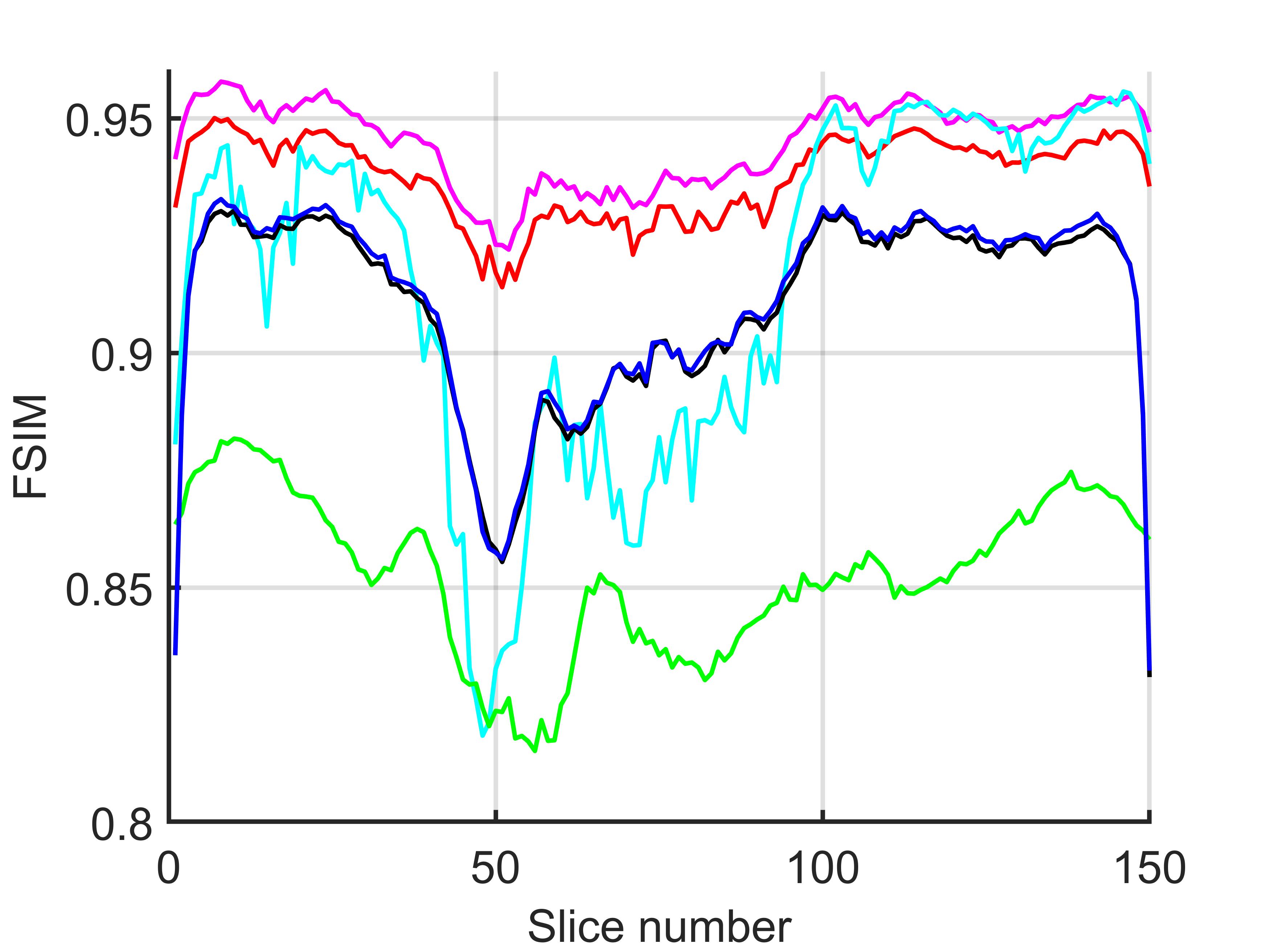}}%
	\caption{The PSNR, SSIM and FSIM of the recovered video "suzie" by MF-TV, TMac, FTNN, PSTNN, TNN, our MCTF and NC-MCTF for all slices, respectively.}
	\label{PSNR and SSIM of video}
\end{figure}

%\begin{figure}[!t]
%	\centering
%
%	\caption{One slice of the recovered MRI by MF-TV, TMac, PSTNN, TNN, our MCTF and NC-MCTF.  The sampling rate is 10\%.}
%	\label{figure_MR_sr0.1_3}
%\end{figure}
%
%
%
%\begin{figure}[!t]
%	\centering
%
%	\caption{One slice of the recovered MRI by MF-TV, TMac, PSTNN, TNN, our MCTF and NC-MCTF.  The sampling rate is 10\%.}
%	\label{figure_MR_sr0.1_4}
%\end{figure}

\begin{table}
	\centering
	\caption{The averaged PSNR, SSIM, FSIM, ERGA and SAM of the recovered results on hyperspectral image "Cuprite" by MF-TV, TMac, FTNN, PSTNN, TNN, our MCTF and NC-MCTF with different sampling rates. The best value is highlighted in bolder fonts.}
	\label{table_HSI}
	\resizebox{82mm}{27mm}{
		\begin{tabular}{ccccccccc}
			\hline \hline
			&&&SR =0.025&&&&	\\
			method  &	noisy	&   MF-TV   &	TMac    &	PSTNN   &	TNN	        &MCTF   &NC-MCTF\\
			PSNR	&	7.666	&   26.115	&	21.25	&	13.387	&	22.783	    &\underline{31.091}  &\textbf{31.208} 	\\
			SSIM	&	0.007	&	0.539	&	0.412	&	0.124	&	0.554	    &\underline{0.771}   &\textbf{0.774} 		\\
			FSIM	&	0.48	&	0.765	&	0.755	&	0.613	&	0.775	    &\underline{0.842}   &\textbf{0.847} 		\\
			ERGAS	&	1043.633&	237.074	&	235.594	&	539.574	&	245.333	    &\underline{77.458}  &\textbf{76.503} 		\\
%			MSAM	&	81.221	&	12.913	&	7.842	&	17.98	&	9.156	    &2.512   &\textbf{2.468} 		\\
			\hline
			&&&SR = 0.05&&&&\\
			method  &	noisy   &	MF-TV   &	TMac    &	PSTNN   &	TNN	        &MCTF   &NC-MCTF\\
			PSNR	&	7.779	&   34.684	&	28.945	&	20.621	&	26.579   	&\underline{34.739}  &\textbf{35.481} 		\\
			SSIM	&	0.01	&	0.845	&	0.712	&	0.31	&	0.663		&\underline{0.860}   &\textbf{0.879} 	\\
			FSIM	&	0.471	&	\underline{0.915}	&	0.846	&	0.735	&	0.836	    &0.907   &\textbf{0.920} 		\\
			ERGAS	&	1030.139&	89.372	&	93.352	&	234.445	&	154.292	    &\underline{51.913}  &\textbf{48.063} 		\\
%			MSAM	&	77.268	&	4.386	&	3.278	&	7.886	&	5.413	    &1.751   &\textbf{1.653} 		\\
			\hline
			&&&SR = 0.1&&&&\\
			method	&	noisy	&	MF-TV	&	TMac	&	PSTNN	&	TNN	        &MCTF   &NC-MCTF	\\
			PSNR	&	8.013	&	\textbf{40.888}	&	35.627	&	35.51	&	35.015	    &37.449  &\underline{37.623} 		\\
			SSIM	&	0.014	&	\textbf{0.957}	&	0.885	&	0.907	&	0.897	    &0.912   &\underline{0.913} 		\\
			FSIM	&	0.451	&	\textbf{0.978}	&	0.931	&	0.951	&	0.943	    &0.943   &\underline{0.943} 		\\
			ERGAS	&	1002.75	&	\textbf{34.263}	&	44.518	&	54.421	&	57.537	    &39.232  &\underline{38.546} 		\\
%			MSAM	&	71.695	&	1.46	&	1.445	&	2.072	&	2.192	    &1.452   &\textbf{1.410} 		\\
			%	\hline
			%	&&&SR = 0.15&&&&\\
			%	method	&	Nosiy	&	our model-2	&	our model-1	&	MF-TV	&	Tmac	&	PSTNN	&	TNN	\\
			%	PSNR	&	8.262	&	\textbf{42.389}	&	41.009	&	42.097	&	36.066	&	39.639	&	40.174	\\
			%	SSIM	&	0.016	&	\textbf{0.968}	&	0.961	&	0.967	&	0.893	&	0.951	&	0.955	\\
			%	FSIM	&	0.433	&	\textbf{0.984}	&	0.979	&	0.984	&	0.935	&	0.975	&	0.976	\\
			%	ERGA	&	974.415	&	\textbf{26.614}	&	29.588	&	31.141	&	42.464	&	36.065	&	34.653	\\
			%	MSAM	&	67.293	&	\textbf{1.055}	&	1.146	&	1.267	&	1.379	&	1.483	&	1.437	\\
			\hline \hline
	\end{tabular}}
\end{table}

\subsection{Hyperspectral Image}

%**This example has some real applications, but in these applications
%very specific measurements are absent (not randomly picked ones). As
%you do not explain how you remove data, it is not possible to be sure,
%but probably the problem is an artificial one as well.
%
%As a reviewer I would therefore reject the paper as it does not show
%that it can be used for any practical real application
%****

\begin{figure*}[t]
	\centering
	\captionsetup[subfloat]{labelsep=none,format=plain,labelformat=empty} %\captionsetup[subfloat]{labelsep=period}
	\subfloat[]{\includegraphics[width=0.14\linewidth]{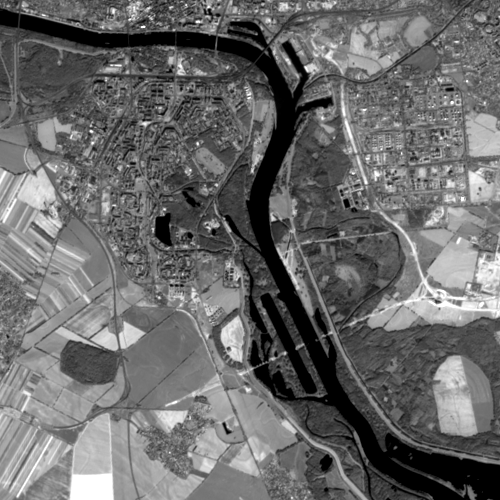}}%
	\hfil
	\subfloat[]{\includegraphics[width=0.14\linewidth]{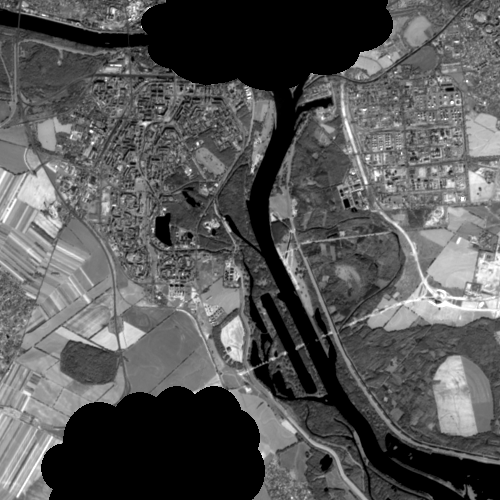}}%
	\hfil
	\subfloat[]{\includegraphics[width=0.14\linewidth]{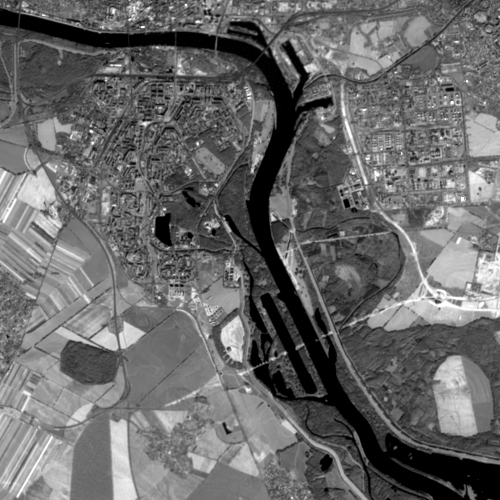}}%
	\hfil
	\subfloat[]{\includegraphics[width=0.14\linewidth]{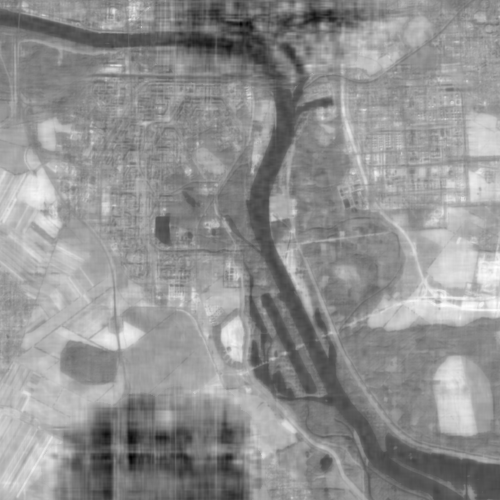}}%
	\hfil
	\subfloat[]{\includegraphics[width=0.14\linewidth]{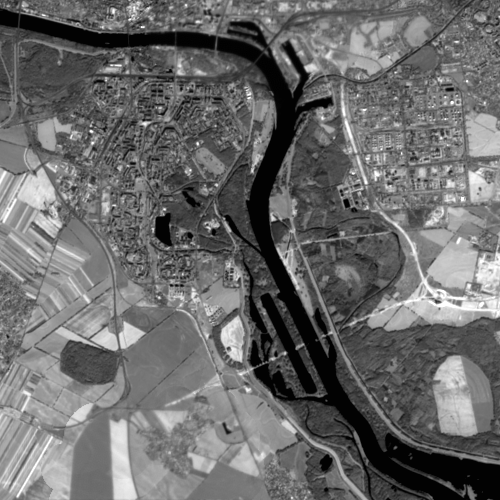}}%	
	\hfil
	\subfloat[]{\includegraphics[width=0.14\linewidth]{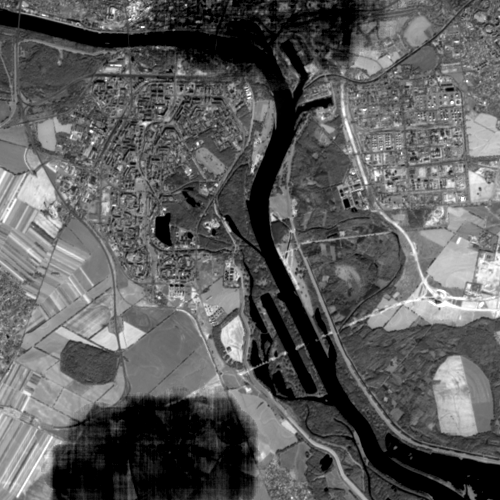}}%
	\hfil
	\subfloat[]{\includegraphics[width=0.14\linewidth]{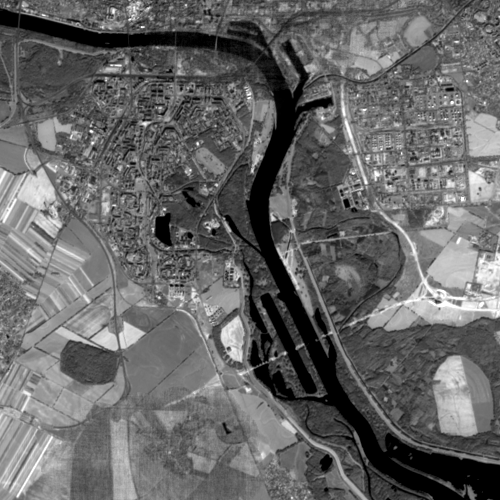}}%%
	\hfil
	\subfloat[]{\includegraphics[width=0.14\linewidth]{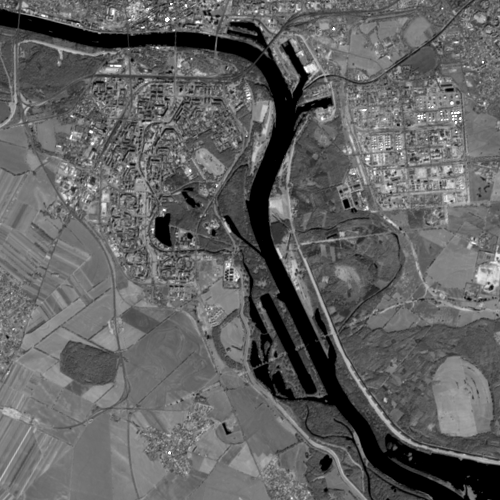}}%
	\hfil
	\subfloat[]{\includegraphics[width=0.14\linewidth]{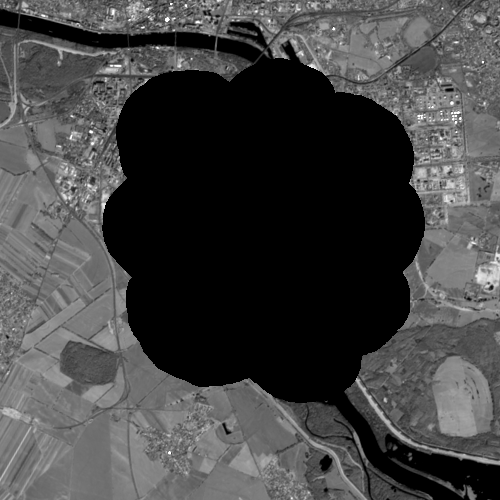}}%
	\hfil
	\subfloat[]{\includegraphics[width=0.14\linewidth]{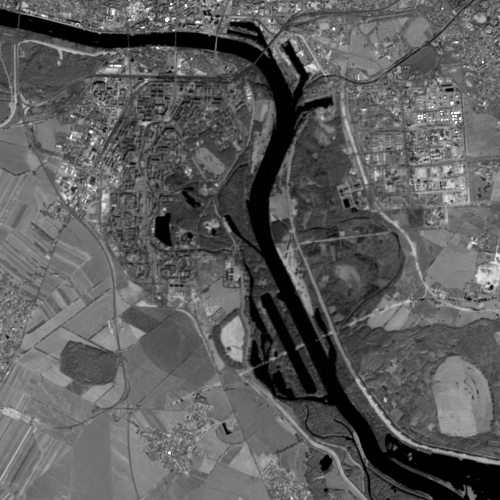}}%
	\hfil
	\subfloat[]{\includegraphics[width=0.14\linewidth]{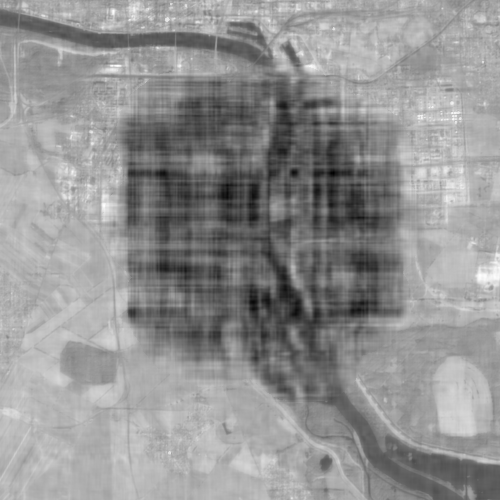}}%
	\hfil
	\subfloat[]{\includegraphics[width=0.14\linewidth]{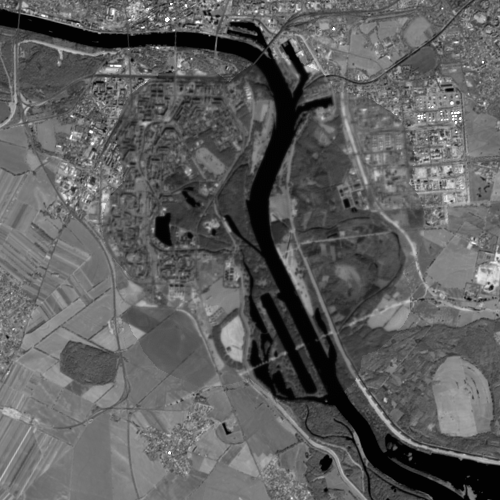}}%	
	\hfil
	\subfloat[]{\includegraphics[width=0.14\linewidth]{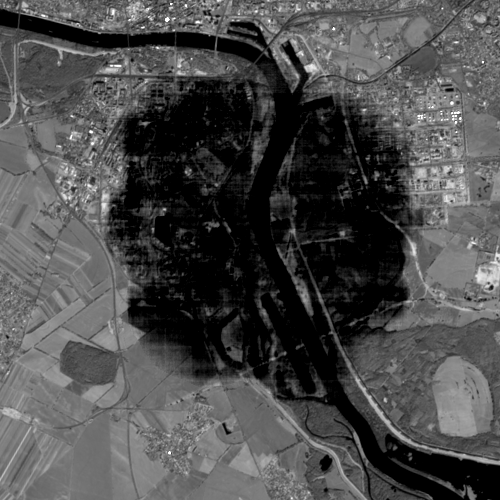}}%
	\hfil
	\subfloat[]{\includegraphics[width=0.14\linewidth]{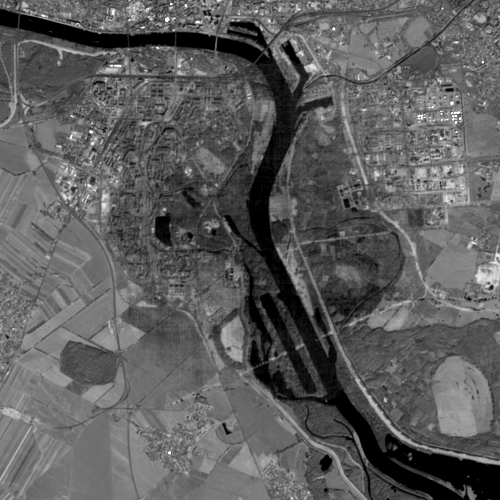}}%%
	\hfil
	\subfloat[Original]{\includegraphics[width=0.14\linewidth]{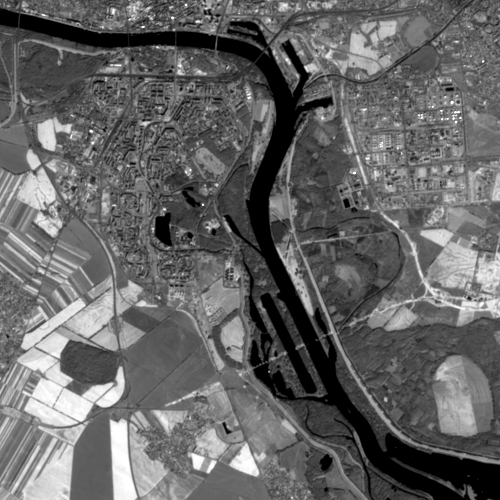}}%
	\hfil
	\subfloat[Cloud-covered]{\includegraphics[width=0.14\linewidth]{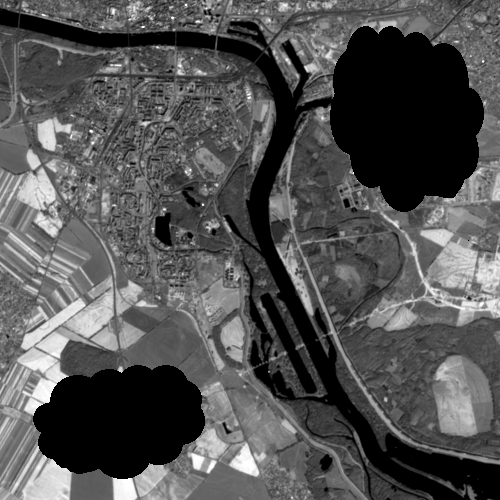}}%
	\hfil
	\subfloat[NC-MCTF]{\includegraphics[width=0.14\linewidth]{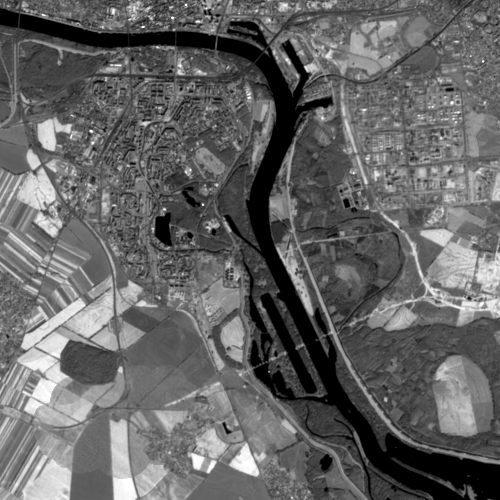}}%
	\hfil
	\subfloat[TMac]{\includegraphics[width=0.14\linewidth]{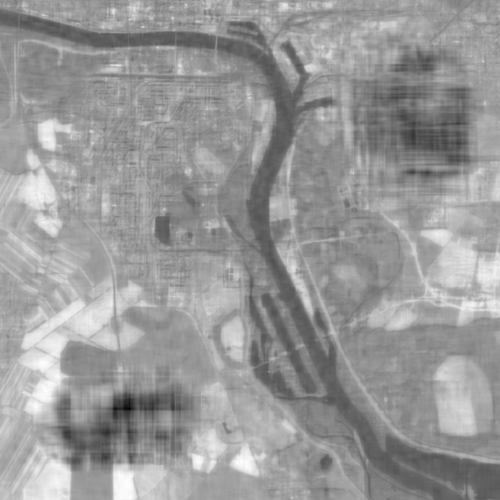}}%
	\hfil
	\subfloat[MF-TV]{\includegraphics[width=0.14\linewidth]{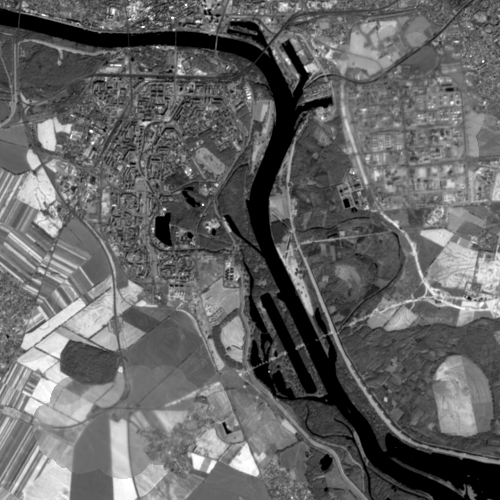}}%	
	\hfil
	\subfloat[PSTNN]{\includegraphics[width=0.14\linewidth]{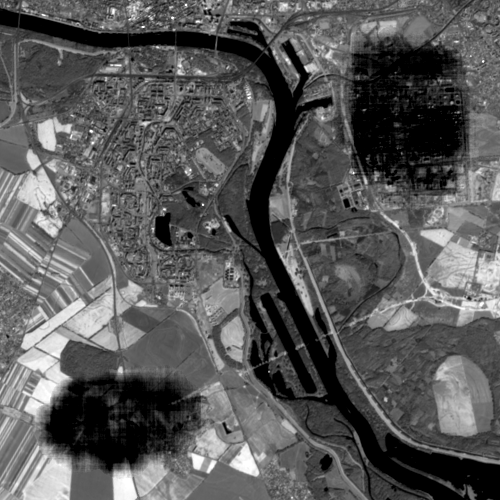}}%
	\hfil
	\subfloat[TNN]{\includegraphics[width=0.14\linewidth]{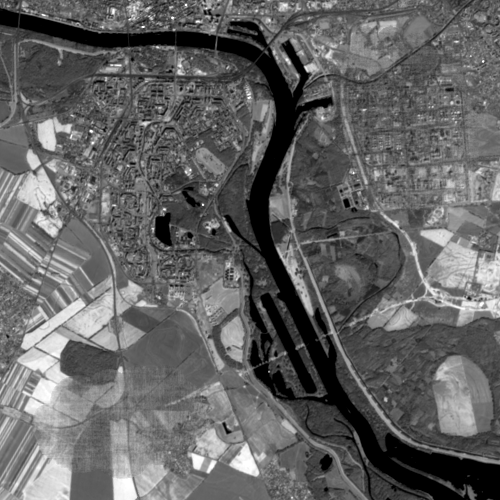}}%%
	\caption{Sentinel-2 MSI multi-temporal data sequence cloud removal experiments (20 m spatial-resolution; 10 $\times$ 10 km; five temporal images) over Mechelen in Belgium.}
	\label{figure_cloudy2}
\end{figure*}

In this subsection, we choose two HSI data to apply simulation experiments.
The first dataset is five sequential images\footnote{https://drive.google.com/file/d/1LlvUKtUWAKoF6R0igbREwvP2Wfja9U\\Bv/view} were acquired
by the Sentinel-2 MSI on 05/09, 15/09, 20/09, 5/10 and 15/10, 2018, in Belgium, with 20 m spatial-resolution, 10 $\times$ 10 km.
For this dataset, the HSIs are corrupted by various types of missing areas with cloud shape (see the second row of Fig. \ref{figure_cloudy2}) .

The second data-set is the airborne visible/infrared imaging spectrometer (AVIRIS) copper salt data\footnote{http://aviris.jpl.nasa.gov/html/aviris.freedata.html} with size 150 $ \times $ 150 $ \times $ 210.
SRs are set as follows: 0.025, 0.05 and 0.1.
Here, the missing values are also random sampled, and we set the rank to $(T_1, T_2, T_3)$, where $T_1, T_2, T_3$ denote the number of the largest 0.5\% singular values of model-1, model-2 and model-3, respectively. Because FTNN did not perform HSI experiments, the original article of FTNN did not describe the parameter settings of HSI dataset, therefore, in this subsection we will not perform comparison experiments on FTNN.

\begin{figure*}[t]
	\centering
	\captionsetup[subfloat]{labelsep=none,format=plain,labelformat=empty} %\captionsetup[subfloat]{labelsep=period}
	\subfloat[Original]{\includegraphics[width=0.12\linewidth]{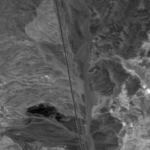}}%
	\hfil
	\subfloat[95\% Masked]{\includegraphics[width=0.12\linewidth]{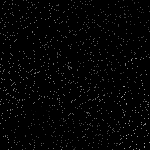}}%
	\hfil
	\subfloat[MF-TV]{\includegraphics[width=0.12\linewidth]{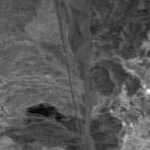}}%
	\hfil
	\subfloat[TMac]{\includegraphics[width=0.12\linewidth]{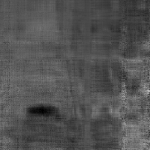}}%
	\hfil
%	\subfloat[FTNN]{\includegraphics[width=0.107\linewidth]{myfigure/Image_result/Image_HSI_b72_sr0.05/FTNN}}%
%	\hfil
	\subfloat[PSTNN]{\includegraphics[width=0.12\linewidth]{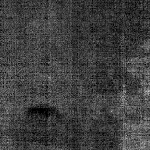}}%
	\hfil
	\subfloat[TNN]{\includegraphics[width=0.12\linewidth]{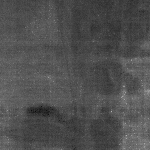}}%
	\hfil
	\subfloat[MCTF]{\includegraphics[width=0.12\linewidth]{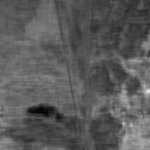}}%%
	\hfil
	\subfloat[NC-MCTF]{\includegraphics[width=0.12\linewidth]{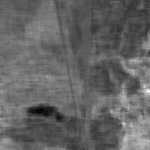}}%%
	\caption{One slice of the recovered HSI "Cuprite" by MF-TV, TMac, PSTNN, TNN, our MCTF and NC-MCTF.  The sampling rate is 5\%.}
	\label{figure_HSI_sr0.05}
\end{figure*}

Table \ref{table_HSI} lists the PQIs of the results restored by all the test models at three different SRs.
It can be clearly seen that the two proposed methods obtain the best PQIs, among all the test methods.
Fig. \ref{figure_cloudy2} and Fig. \ref{figure_HSI_sr0.05} show the visual results of ground truth, simulated cloud-covered/missing area, recovery results of TMac, MF-TV, PSTNN, TNN and the proposed NC-MCTF.

\subsection{Parameter selection}

\begin{figure}[t]
	\centering
	\subfloat[SR=0.05]{\includegraphics[width=0.32\linewidth]{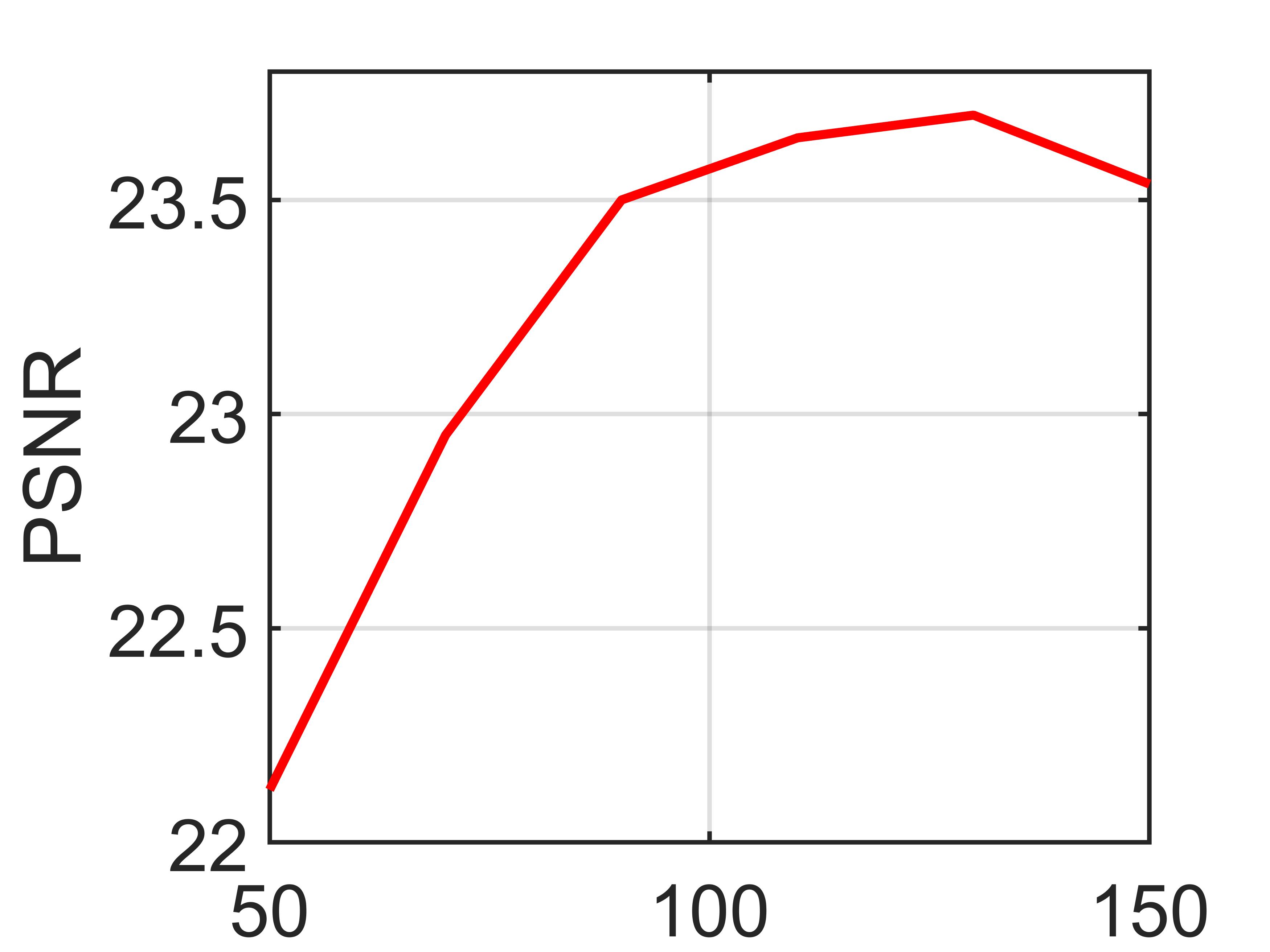}}%
	\hfil
	\subfloat[SR=0.2]{\includegraphics[width=0.32\linewidth]{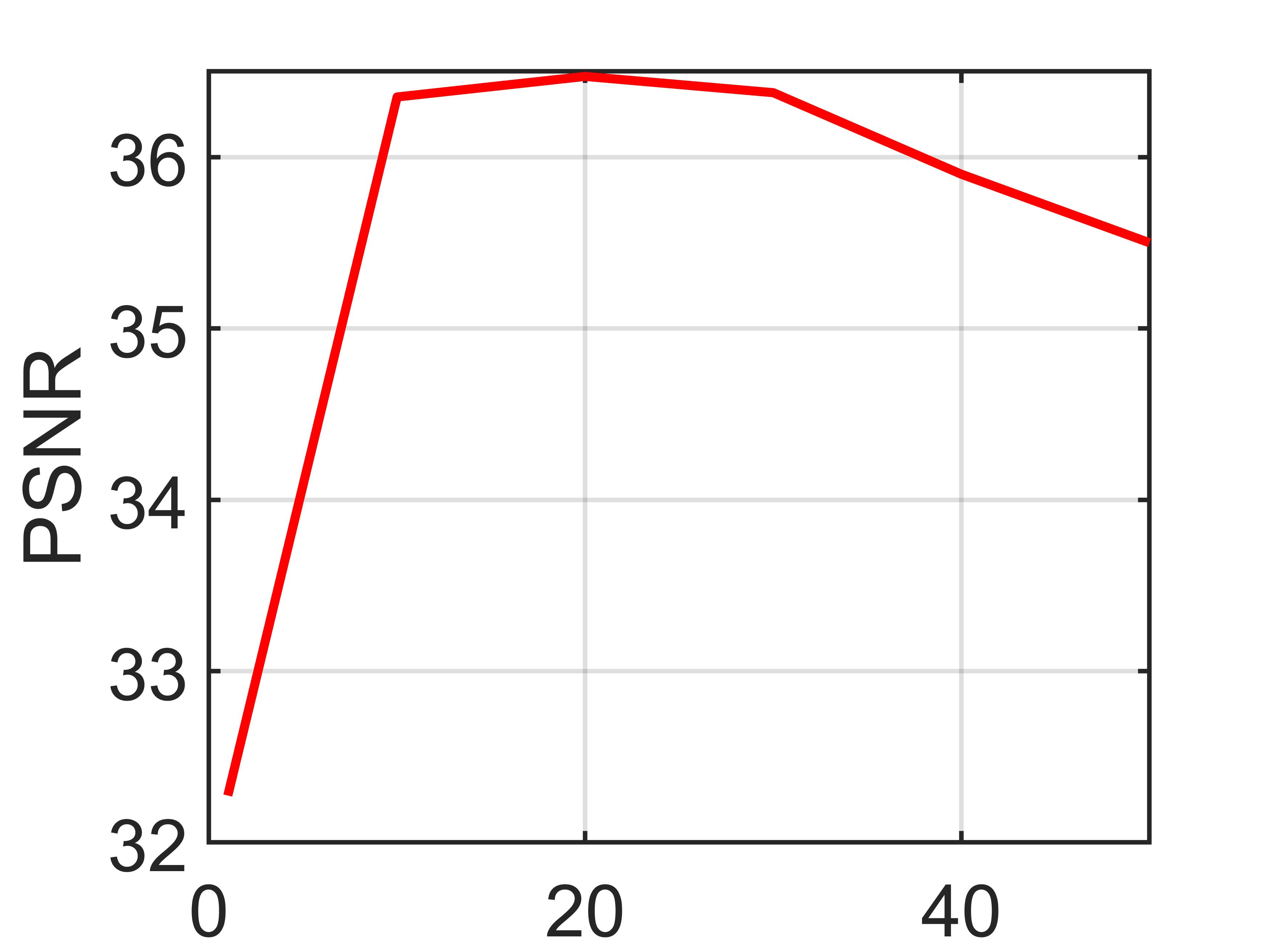}}%
	\hfil
	\subfloat[SR=0.3]{\includegraphics[width=0.32\linewidth]{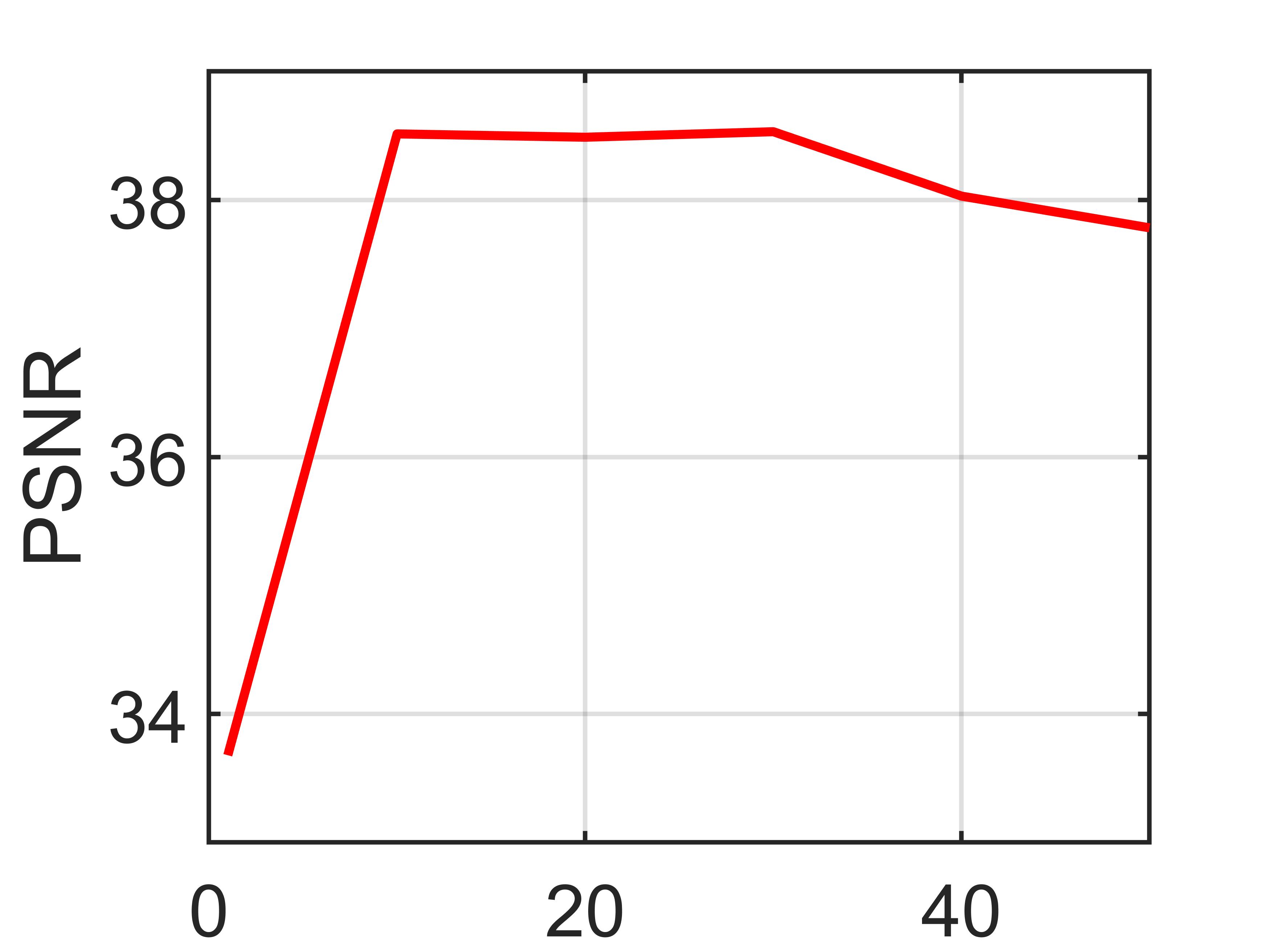}}%%
	\hfil
	\subfloat[SR=0.05]{\includegraphics[width=0.32\linewidth]{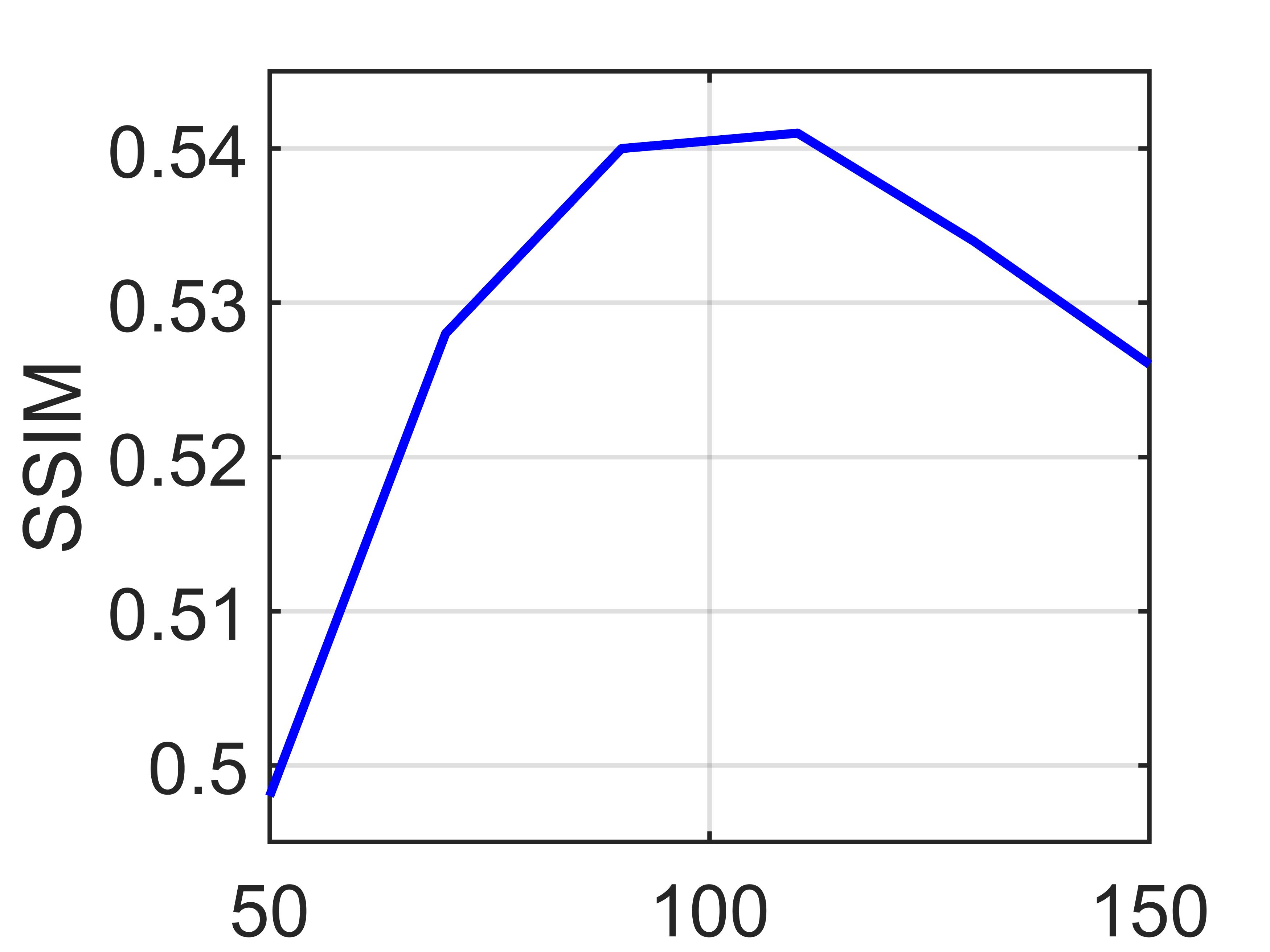}}%
	\hfil
	\subfloat[SR=0.2]{\includegraphics[width=0.32\linewidth]{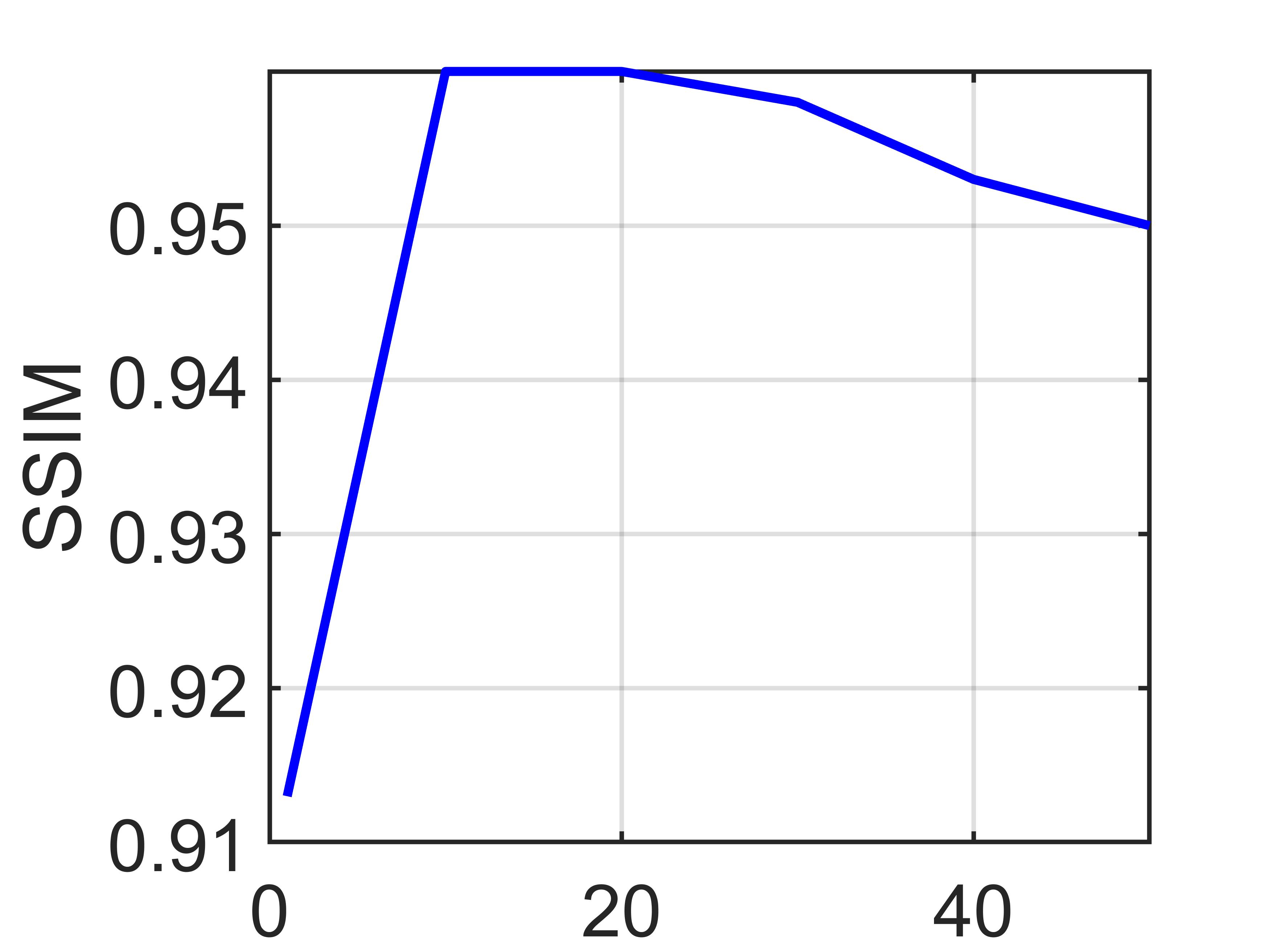}}%
	\hfil
	\subfloat[SR=0.3]{\includegraphics[width=0.32\linewidth]{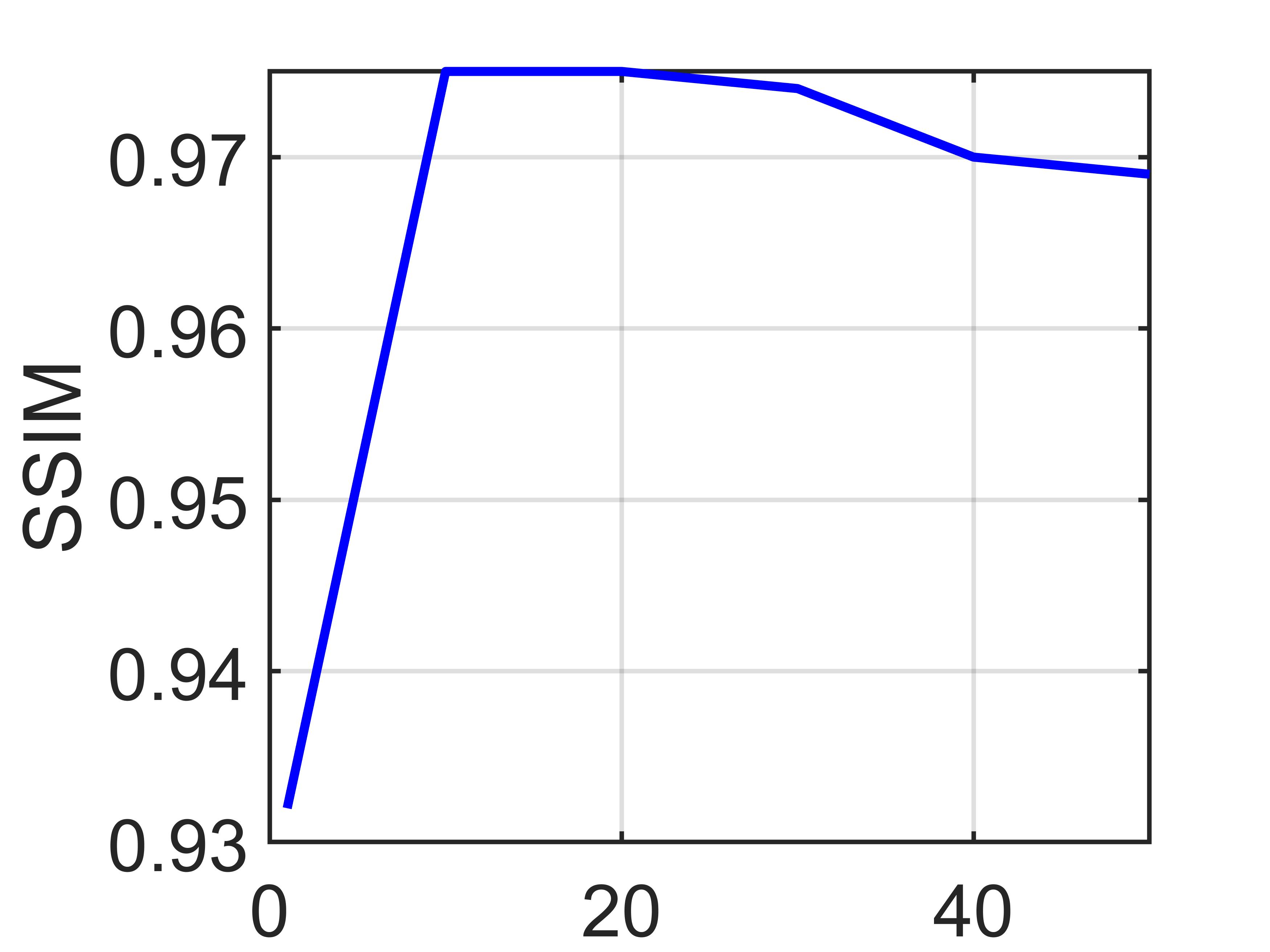}}%%
	\caption{Performance variation of the proposed method in terms of the
		NC-MCTF on different $C$ and different sampling rate.}
	\label{psnr_parameter}
\end{figure}

Since the proposed method consists of two balanced terms, i.e., $\tau_n \left\|\mathbf{X}_{n}\right\|_{\text{* or log}}
+\lambda_n \left\|\mathcal{G}_{n}\right\|_{\Lambda_n, \text{* or log}}$, that need parameters to tradeoff them, it is necessary to discuss the issue of setting the parameter appropriately. 
To reduce the workload of adjusting parameters, we fix one of $\tau_{n}$ and $\lambda_{n}$ (to enhance the generalization ability of the parameters, here, we set the same $\tau_{n}$ and $\lambda_{n}$ for different $n$), and then indirectly adjust the ratio of the two, i.e., $C = \frac{\tau_{n}}{\lambda_{n}}$.
In this subsection, we provided some experiments with real tensor data to analyze this problem. We set the sampling rate to 0.05, 0.2, and 0.3 respectively.
Under the above three different sampling rates, Fig. \ref{psnr_parameter} visually shows the performance of the proposed method under different settings of $C$.

\section{Conclusions}

In this paper, based on basic tensor decomposition theory, we first
defined a novel tensor decomposition to explore the low-rank structure
of underlying tensors. Then, we studied the structural characteristics
of the factors obtained by the decomposition, and proposed a novel
tensor low-rankness measure. Furthermore, we performed nonconvex
relaxation on the sub-measure, and obtained a better low-rankness
measure. Thanks to the ``good'' properties of TNN and Tucker
decomposition, our model and theoretical guarantee are both natural
extensions of TNN and Tucker.
We have also developed a effective method to calculate the optimization problem corresponding to the proposed model. Numerical experiments verify our theory, and the results of hyperspectral image, MRI and video prove that our algorithms can recover a variety of low-rank tensors with significantly fewer samples than the compared methods.

%\section*{Acknowledgment}
%The authors would like to express their thanks to Dr. Y. Xu, Dr. Z. Zhang, Dr. T. Ji and Dr. T. Jiang for sharing their codes for the test methods. This research is supported by the Fundamental Research Funds for the Central Universities under Grant No. 2452019073 and the National
%Natural Science Foundation of China under Grant No. 61876153.

%
\appendices
\section{Details of the Optimization}

\subsection{Update $\mathbf{X}_n$ with fixing others}

The $\mathbf{X}_n$-subproblem in (\ref{equation:XAY}) can be written as follows:
\begin{equation}
\label{equation_X}
\begin{aligned}
\mathbf{X}_n^{k+1}=\argmin_{\mathbf{X}_n} \sum_{n=1}^{3} \frac{\alpha_{n}}{2}\left\|\mathcal{Y}-\mathcal{G}_{n} \times_{n} \mathbf{X}_{n}\right\|_{\text{F}}^{2}+\tau_n \left\|\mathbf{X}_{n}\right\|_{\text{* or log}}\\
+\frac{\rho_n}{2}\left\|\mathbf{X}_n-\mathbf{X}_n^{k}\right\|_{\text{F}}^{2}.
\end{aligned}
\end{equation}
To efficiently solve it,
we first introduce one auxiliary variable $\mathbf{Z}_n$,
then (\ref{equation_X}) can be rewritten as
\begin{equation}
\label{equation_X_aux}
\begin{aligned}
\argmin_{\mathbf{X}_n, \mathbf{Z}_n}& \sum_{n=1}^{3} (\frac{\alpha_{n}}{2}\left\|\mathcal{Y}-\mathcal{G}_{n} \times_{n} \mathbf{X}_{n}\right\|_{\text{F}}^{2}+\tau_n \left\|\mathbf{Z}_{n}\right\|_{\text{*  or log}}\\
&+\frac{\rho_n}{2}\left\|\mathbf{X}_n-\mathbf{X}_n^{k}\right\|_{\text{F}}^{2}), s.t., \mathbf{X}_{n}=\mathbf{Z}_{n}.
\end{aligned}
\end{equation}
Based on the augmented Lagrange multiplier (ALM) method, the above minimization problem (\ref{equation_X_aux}) can be transformed into
\begin{equation}
\label{equation:X_alm}
\begin{aligned}
&\argmin_{\mathbf{X}_n, \mathbf{Z}_n} \sum_{n=1}^{3} (\frac{\alpha_{n}}{2}\left\|\mathcal{Y}-\mathcal{G}_{n} \times_{n} \mathbf{X}_{n}\right\|_{\text{F}}^{2}+\tau_n \left\|\mathbf{Z}_{n}\right\|_{\text{*  or log}}\\
&+\frac{\rho_n}{2}\left\|\mathbf{X}_n-\mathbf{X}_n^{k}\right\|_{\text{F}}^{2}+\left\langle\Gamma_{n}^\mathbf{X}, \mathbf{X}_n-\mathbf{Z}_n\right\rangle+\frac{\rho_{n}}{2}\left\|\mathbf{X}_{n}-\mathbf{Z}_{n}\right\|_{\text{F}}^{2}),
\end{aligned}
\end{equation}
where $\Gamma_{n}^\mathbf{X}$ is a Lagrange multiplier.

%*** You should define teh notations better: Is  $I_1I_2
%\times I_3 =  (I_1I_2) \times I_3$ or $I_1I_2
%\times I_3 =  I_1(I_2 \times I_3)$?****
%
%***what does $.^*$ mean? The proper matrix decomposition of the SVD is
%$U S V ^t$ or $US V^{\dag}$ for complex-valued data.****
%
%****???? obey means  ``follows orders from''***

With other variables fixed, the minimization subproblem for $\mathbf{Z}_n$ can be deduced from (\ref{equation:X_alm}) as follows:
\begin{equation}
\displaystyle \mathbf{Z}_n^{k+1}= \argmin_{\mathbf{Z}_n} \tau_n \left\|\mathbf{Z}_{n}\right\|_{\text{* or log}}+\frac{\rho_{n}}{2}\left\|\mathbf{X}_{n}^{k}-\mathbf{Z}_{n}+\Gamma_{n}^{\mathbf{X}}/\rho_n\right\|_{\text{F}}^{2}.
\end{equation}
For MCTF based on $\left\|.\right\|_{\text{*}}$, by using the SVT operator (\ref{SVT}), it is easy to get
\begin{equation}
\begin{aligned}
\mathbf{Z}_n^{k+1}=\operatorname{D}_{\frac{\tau_{n}}{\rho_n}}(\mathbf{X}_{n}^{k}+\Gamma_{n}^{\mathbf{X}}/\rho_n), n=1,2, \cdots, N;
\end{aligned}
\end{equation}
while for the $\mathbf{Z}_n$ related subproblem of nonconvex low-rankness measure (NC-MCTF):
\begin{equation}
\displaystyle \mathbf{Z}_n^{k+1}= \argmin_{\mathbf{Z}_n} \tau_n \left\|\mathbf{Z}_{n}\right\|_{\text{log}}+\frac{\rho_{n}}{2}\left\|\mathbf{X}_{n}^{k}-\mathbf{Z}_{n}+\Gamma_{n}^{\mathbf{X}}/\rho_n\right\|_{\text{F}}^{2},
\end{equation}
it can be solved by $\mathrm{WNNM}$ operator, i.e.,
\begin{equation}
\mathbf{Z}_n^{k+1} = W_{\frac{\tau_{n}}{\rho_n}, \epsilon}\left(\mathbf{X}_{n}^{k}+\Gamma_{n}^{\mathbf{X}}/\rho_n\right), n=1,2, \cdots, N.
\end{equation}

With other variables fixed, the minimization subproblem for $\mathbf{X}_n (n\not=3)$ can be deduced from (\ref{equation:X_alm}) as follows:
\begin{equation}
\label{equationforX}
\begin{aligned}
\mathbf{X}_n^{k+1}&= \argmin_{\mathbf{X}_n}\frac{\alpha_{n}}{2}\left\|\mathcal{Y}-\mathcal{G}_{n}^k \times_{n} \mathbf{X}_{n}\right\|_{\text{F}}^{2}\\
&+\frac{\rho_{n}}{2}\left\|\mathbf{X}_{n}-\frac{\mathbf{Z}_{n}^{k+1}-\Gamma_{n}^k/\mu_n+\mathbf{X}_n^{k}}{2}\right\|_{\text{F}}^{2}.
\end{aligned}
\end{equation}
They are convex and have the following closed-form solutions
\begin{equation}
\begin{aligned}
\mathbf{X}_n^{k+1}&=(\alpha_{n}\mathbf{G}_n^T\mathbf{G}_n+2\rho \mathbf{I}_n)^{-1}[\alpha_{n}\mathbf{G}_n^T\mathbf{Y}_{(n)}\\
&+\mu_n (\frac{\mathbf{Z}_{n}^{k+1}-\Gamma_{n}^k/\mu_n+\mathbf{X}_n^{k}}{2}).
\end{aligned}
\end{equation}
Based on the ALM method, the multipliers are updated by the following equations:
\begin{equation}
\Gamma_{n}^{\mathbf{X}} = \Gamma_{n}^{\mathbf{X}} + \mathbf{X}_n-\mathbf{Z}_n.
\end{equation}

\subsection{Update $\mathcal{G}_n$ with fixing others}

The $\mathcal{G}_n$-subproblem in (\ref{equation:XAY}) can be written as follows:
\begin{equation}
\label{equation_A_aux}
\begin{aligned}
\mathcal{G}^{k+1}=\argmin_{\mathcal{G}} \sum_{n=1}^{3}& (\frac{\alpha_{n}}{2}\left\|\mathcal{Y}-\mathcal{G}_{n} \times_{n} \mathbf{X}_{n}\right\|_{\text{F}}^{2}+\lambda_n \left\|\mathcal{G}_n\right\|_{\Lambda_n, \text{*}}\\
&+\frac{\rho_n}{2}\left\|\mathcal{G}_n-\mathcal{G}_n^{k}\right\|_{\text{F}}^{2}).
\end{aligned}
\end{equation}
By introducing an auxiliary variable, (\ref{equation_A_aux}) can be rewritten as
\begin{equation}
\begin{aligned}
\argmin_{\mathcal{G}_n} \sum_{n=1}^{3}& (\frac{\alpha_{n}}{2}\left\|\mathcal{Y}-\mathcal{G}_{n} \times_{n} \mathbf{X}_{n}\right\|_{\text{F}}^{2}+\lambda_n \left\|\mathcal{J}_{n}\right\|_{\Lambda_n,\text{*}}\\
&+\frac{\rho_n}{2}\left\|\mathcal{G}_n-\mathcal{G}_n^{k}\right\|_{\text{F}}^{2}),
s.t., \mathcal{G}_{n}=\mathcal{J}_{n}.
\end{aligned}
\end{equation}
Then, (\ref{equation_A_alm}) can also be reformulated as
\begin{equation}
\label{equation:A_alm}
\begin{aligned}
&\argmin_{\mathbf{A}_n, \mathbf{J}_n}  \sum_{n=1}^{3} (\frac{\alpha_{n}}{2}\left\|\mathcal{Y}-\mathcal{G}_{n} \times_{n} \mathbf{X}_{n}\right\|_{\text{F}}^{2}+\lambda_n \left\|\mathcal{J}_{n}\right\|_{\Lambda_n, \text{*}}\\
&+\frac{\rho_n}{2}\left\|\mathcal{G}_n-\mathcal{G}_n^{k}\right\|_{\text{F}}^{2}
+\left\langle\Gamma_{n}^\mathcal{G}, \mathcal{G}_n-\mathcal{J}_n\right\rangle+\frac{\rho_{n}}{2}\left\|\mathcal{G}_{n}-\mathcal{J}_{n}\right\|_{\text{F}}^{2}),
\end{aligned}
\end{equation}
where $\Gamma_{n}^\mathcal{G}$ is the Lagrangian multiplier.

Firstly, with other variables fixed, the minimization subproblem for $\mathcal{J}_n$ can be deduced from (\ref{equation:A_alm}) as follows:
\begin{equation}
\label{equation_Jn}
\displaystyle \mathcal{J}_n^{k+1}= \argmin_{\mathcal{J}_n} \lambda_n \left\|\mathcal{J}_{n}\right\|_{\Lambda_n, \text{*}}+\frac{\rho_{n}}{2}\left\|\mathcal{G}_{n}^{k}-\mathcal{J}_{n}+\Gamma_{n}^{\mathcal{G}}/\rho_n\right\|_{\text{F}}^{2}.
\end{equation}
%Its solution can also be obtained by SVT operator (\ref{SVT})
%\begin{equation}
%\label{equation_solution_Jn}
%\begin{aligned}
%\mathcal{J}_n^{k+1}=\operatorname{SH}_{\frac{\lambda_n}{\rho_{n}}}(\mathcal{G}_{n}^{k}+\Gamma_{n}^{\mathcal{G}}/\rho_n ), n=1,2, \cdots, N.
%\end{aligned}
%\end{equation}

The updating of $\mathcal{J}_n$ in (\ref{equation_Jn}) has a closed-form solution.
For the sake of simplicity, we denote the iteration of $\mathcal{J}_n$ as
\begin{equation}
\label{eq:Lmodel-simpl}
\mathcal{J}_n^{k+1}=\argmin_{ \mathcal{J}_n } \left\|\mathcal{J}_n\right\|_{\Lambda_n, *}+\times\frac{\rho_n}{2}\left\|\mathcal{J}_n-\mathcal{U}_n\right\|_{\mathrm{F}}^{2},
\end{equation}
where $\mathcal{U}_n=\mathcal{G}_{n}^{k}+\Gamma_{n}^{\mathcal{G}}/\rho_n$.

Solving the optimization subproblem (\ref{eq:Lmodel-simpl}) is equivalent to solving the following tensor recovery problem in the frequency domain:
\begin{equation}
\label{eq:Lmodel-fft}
_n\widehat{\mathcal{J}}_n^{k+1}=\argmin_{_n \widehat{\mathcal{J}}_n } \frac{1}{p}\sum_{q=1}^{p}\left\|_n\widehat{J}^{(q)}_n\right\|_{*}+\times\frac{\rho_n}{2}\left\|_n\widehat{\mathcal{J}}_n-_n\widehat{\mathcal{U}}_n\right\|_{\mathrm{F}}^{2},
\end{equation}
where $_n\widehat{\mathcal{J}}_n = \operatorname{fft}\left(\mathcal{J}_n,[~], n\right)$,
$_n\widehat{\mathcal{U}}_n = \operatorname{fft}\left(\mathcal{U}_n,[~], n\right)$,
and $_n\widehat{J}^{(q)}_n$ denotes the $q$-th frontal slice of $_n\widehat{\mathcal{J}}_n$.
For the optimization subproblem in the frequency domain defined in (\ref{eq:Lmodel-fft}), we can divide it into $p$ independent minimization subproblems:
\begin{equation}
\label{eq:Lmodel-fft-slice}
_n\widehat{{J}}_n^{k+1,(q)}=\argmin_{_n\widehat{{J}}_n^{(q)}} \left\|_n\widehat{J}_n^{(q)}\right\|_{*}+
\times\frac{\rho_n}{2}\left\|_n\widehat{J}_n^{(q)}-_n\widehat{U}_n^{(q)}\right\|_{\mathrm{F}}^{2}.
\end{equation}
By using the SVT operator (\ref{SVT}), it is easy to obtain the solutions of the minimization subproblems (\ref{eq:Lmodel-fft-slice})
\begin{equation}
_n\widehat{J}_n^{k+1,(q)} = D_{\frac{1}{\rho_n}}\left(_n\widehat{\mathcal{U}}_n^{(q)}\right), ~q=1,2,\cdots,p.
\end{equation}
Then, the $(k+1)$-th updating of $\mathcal{J}^{k+1}_n$ can be obtained via inverse Fourier transform
\begin{equation}
\mathcal{J}_n^{k+1} =\operatorname{ifft}\left(_n\widehat{\mathcal{J}}_n^{k+1},[~], n\right).
\end{equation}

Similarly, the $\mathcal{J}_n$ related subproblem of nonconvex low-rankness measure
\begin{equation}
\label{eq:Lmodel-logfft-slice}
_n\widehat{{J}}_n^{k+1,(q)}=\argmin_{_n\widehat{{J}}_n^{(q)}} \left\|_n\widehat{J}_n^{(q)}\right\|_{\text{log}}+
\times\frac{\rho_n}{2}\left\|_n\widehat{J}_n^{(q)}-_n\widehat{U}_n^{(q)}\right\|_{\mathrm{F}}^{2}.
\end{equation}
can be solved by the $\mathrm{WNNM}$ operator (\ref{equation:WNNM_operator}), i.e.,
\begin{equation}
_n\widehat{J}_n^{k+1,(q)} = W_{\frac{1}{\rho_n}, \epsilon}\left(_n\widehat{\mathcal{U}}_n^{(q)}\right), ~q=1,2,\cdots,p.
\end{equation}
Then, the $(k+1)$-th updating of $\mathcal{J}^{k+1}_n$ can be obtained via inverse Fourier transform
\begin{equation}
\mathcal{J}_n^{k+1} =\operatorname{ifft}\left(_n\widehat{\mathcal{J}}_n^{k+1},[~], n\right).
\end{equation}

Secondly, with other variables fixed, the minimization subproblem for $\mathcal{G}_n$ can be deduced from (\ref{equation:A_alm}) as follows:
\begin{equation}
\label{equationforA}
\begin{aligned}
\mathcal{G}_n^{k+1}= \argmin_{\mathcal{G}_n} \sum_{n=1}^{3} ( \frac{\alpha_{n}}{2}\left\|\mathcal{Y}-\mathcal{G}_{n} \times_{n} \mathbf{X}_{n}\right\|_{\text{F}}^{2}+\\
\rho_{n}\left\|\mathcal{G}_{n}-\frac{\mathcal{J}_{n}^{k+1}-\Gamma_{n}^\mathcal{G}/\rho_n+\mathcal{G}_n^{k}}{2}\right\|_{\text{F}}^{2}).
\end{aligned}
\end{equation}
It is also convex and has the following closed-form solution
\begin{equation}
\begin{array}{r}
\mathcal{G}_n^{k+1}=\operatorname{fold}\left(\left(\mathbf{Y}_{(n)}^{k}\left(\mathbf{X}_{n}^{k+1}\right)^{T}+2\rho_n (\frac{\mathbf{J}_{n}^{k+1}-\Gamma_{n}^\mathcal{G}/\rho_n+\mathbf{G}_n^{k}}{2})\right)\right.\\
\left. \left(\mathbf{X}_{n}^{k+1}\left(\mathbf{X}_{n}^{k+1}\right)^{T}+2\rho_{n} \mathbf{I}_{n}\right)^{\dagger}\right), \\
n=1,2,\cdots, N.
\end{array}
\end{equation}

Finally, the Lagrangian multiplier can be updated by the following equations
\begin{equation}
\Gamma_{n}^{\mathcal{G}} = \Gamma_{n}^{\mathcal{G}} + \mathcal{G}_n-\mathcal{J}_n.
\end{equation}

\subsection{Update $\mathcal{Y}$ with fixing others}

With other variables fixed, the minimization subproblem for $\mathcal{Y}_{n}$ in (\ref{equation:XAY}) can be written as
\begin{equation}
\begin{aligned}
\mathcal{Y}^{k+1} = \argmin_{\mathcal{Y}} & \sum_{n=1}^{3} ( \frac{\alpha_{n}}{2}\left\|\mathcal{Y}-\mathcal{G}_{n} \times_{n} \mathbf{X}_{n}\right\|_{\text{F}}^{2}+\frac{\rho}{2}\left\|\mathcal{Y}-\mathcal{Y}^{k}\right\|_{\text{F}}^{2} \\
& s.t., \mathcal{P}_{\Omega}(\mathcal{Y})=\mathcal{F}.
\end{aligned}
\end{equation}
Then, the update of $\mathcal{Y}_{k+1}$ can be written explicitly as
\begin{equation}
\begin{array}{l}
\displaystyle \mathcal{Y}^{k+1}=P_{{\Omega}^c}\left(\sum_{n=1}^{3} \alpha_{n} \text { fold }_{n}\left(\frac{\mathbf{G}_{n}^{k+1} \mathbf{X}_{n}^{k+1}+\rho_n \mathbf{Y}_{(n)}^{k}}{1+\rho_n}\right)\right)+\mathcal{F}.
\end{array}
\end{equation}

%%%%%%%%%
%%%%%%%%%
%%%%%%%%%Appendix one text goes here.
%%%%%%%%%
%%%%%%%%%% you can choose not to have a title for an appendix
%%%%%%%%%% if you want by leaving the argument blank
%%%%%%%%%\section{}
%%%%%%%%%Appendix two text goes here.
%%%%%%%%%
%%%%%%%%%
%%%%%%%%%% use section* for acknowledgment
%%%%%%%%%\section*{Acknowledgment}
%%%%%%%%%
%%%%%%%%%
%%%%%%%%%The authors would like to thank...

\ifCLASSOPTIONcaptionsoff
  \newpage
\fi

\bibliographystyle{IEEEtran}% our
\bibliography{IEEEabrv,mybibfile}% our

\end{document}